\documentclass{article}

\usepackage{microtype}
\usepackage{graphicx}
\usepackage{subfigure}
\usepackage{booktabs} 

\usepackage{chngcntr} 

\usepackage{hyperref}

\usepackage[accepted]{icml2024}

\usepackage{amsmath}
\usepackage{amssymb}
\usepackage{mathtools}
\usepackage{amsthm}

\usepackage[capitalize,noabbrev]{cleveref}

\usepackage[textsize=tiny]{todonotes}

\usepackage{wrapfig}
\usepackage[utf8]{inputenc}
\usepackage[english]{babel}
\usepackage{dsfont}
\usepackage{amsfonts}
\usepackage{amssymb}
\usepackage{graphicx}
\usepackage{tcolorbox}
\usepackage{enumitem}
\tcbuselibrary{listings,breakable} 

\usepackage{etoc}

\usepackage{enumitem}   
\usepackage[makeroom]{cancel}

\DeclareMathOperator{\atanh}{atanh}

\usepackage{mathtools}
\DeclarePairedDelimiter\ceil{\lceil}{\rceil}

\usepackage{natbib}

\DeclareMathOperator\erf{erf}

\newcommand{\ROutNode}[2]{O_{#1 #2}}
\newcommand{\OutNode}[2]{o_{#1 #2}}

\newcommand{\RVecHidNodeBAF}[1]{\boldsymbol{Z}^{(#1)}}
\newcommand{\RCompHidNodeBAF}[2]{Z^{(#1)}_{#2}}

\newcommand{\ArgHidNodeBAF}[0]{z}

\newcommand{\RVecHidNodeAAF}[1]{\boldsymbol{H}^{(#1)}}
\newcommand{\RCompHidNodeAAF}[2]{H^{(#1)}_{#2}}

\newcommand{\ArgHidNodeAAF}[0]{h}

\newcommand{\RVecHidNodeAMP}[2]{\boldsymbol{U}^{(#1 ; #2)}}
\newcommand{\RCompHidNodeAMP}[3]{U^{(#1 ; #2)}_{#3}}

\newcommand{\ArgHidNodeAMP}[0]{u}

\newcommand{\AFOperator}[1]{H \left({#1} \right)}
\newcommand{\MPOperator}[2]{U^{(#1)} \left({#2} \right)}

\newcommand{\RMean}[1]{\mu_{#1}}
\newcommand{\Mean}[1]{m_{#1}}

\newcommand{\RDiff}[1]{\Delta_{#1}}
\newcommand{\Diff}[1]{\delta_{#1}}

\newcommand{\Dataset}[0]{\chi}
\newcommand{\InputValue}[0]{\xi}
\newcommand{\LabelValue}[0]{y}
\newcommand{\Weights}[2]{w^{(#1)}_{#2}}
\newcommand{\RClassFraction}[1]{G_{#1}}
\newcommand{\ClassFraction}[1]{g_{#1}}
\newcommand{\RMultiClassFraction}[2]{G_{#1}^{ #2 }}
\newcommand{\RMultiClassRankedFraction}[2]{G_{\tilde #1}^{ #2 }}
\newcommand{\MultiClassFraction}[2]{g_{#1}^{ #2 }}

\newcommand{\CE}[2]{\mathbb{E}_{#1} \left( {#2}  \right)}
\newcommand{\Einput}[1]{\left\langle {#1}  \right\rangle}
\newcommand{\overbar}[1]{\mkern 1.5mu\overline{\mkern-1.5mu#1\mkern-1.5mu}\mkern 1.5mu}
\newcommand{\Eweights}[1]{\overbar{#1}}

\newcommand{\EinputComp}[1]{\mu_{#1}}

\newcommand{\BP}[0]{1.}

\newcommand{\CVar}[2]{\textrm{Var}_{#1} \left( {#2}  \right)}
\newcommand{\NumberClasses}[0]{N_{C}}
\newcommand{\LayerNumNodes}[1]{N_{#1}}
\newcommand{\WeightsVec}[2]{\boldsymbol{w}^{(#1)}_{#2}}
\newcommand{\WeightsMat}[1]{\boldsymbol{W}^{(#1)}}
\newcommand{\Heavyside}[1]{\Theta \left( #1 \right)}
\newcommand{\Dirac}[1]{\delta \left( #1 \right)}
\newcommand{\DatasetSize}[0]{D}
\newcommand{\WeightSet}[1]{\mathcal{W}^{#1}}
\newcommand{\Prob}[2]{\mathbb{P}^{#1} \left( #2 \right) }
\newcommand{\NormalDistr}[2]{\mathcal{N}\left( #1 , #2 \right) }
\newcommand{\NormalDens}[3]{\mathcal{N}\left( #1 ;  #2 , #3 \right) }

\newcommand{\Arch}[0]{\mathcal{A}}
\newcommand{\PreprData}[0]{\psi (\chi)}

\newcommand{\PreprDataOp}[1]{\psi \left( #1 \right)}

\newcommand{\VarRatio}[0]{\gamma (\Arch, \PreprData)}

\newcommand{\SMO}[2]{C_{\OutNode{#1}{#2}}}

\newcommand{\ProbRat}[0]{R_C}

\newcommand\Cdot{\boldsymbol{\cdot}}

\newcommand{\GB}[0]{\texttt{GB}}
\newcommand{\CIFAR}[0]{\texttt{C10}}
\newcommand{\CIFARHC}[0]{\texttt{C100}}
\newcommand{\MNIST}[0]{\texttt{E\&O}}
\newcommand{\CNNA}[0]{\texttt{CNN-A}}
\newcommand{\CNNB}[0]{\texttt{CNN-B}}
\newcommand{\MLPA}[0]{\texttt{SHLP}}
\newcommand{\MLPB}[0]{\texttt{MHLP}}
\newcommand{\ResNet}[0]{\texttt{ResNet}}
\newcommand{\ViT}[0]{\texttt{ViT}}
\newcommand{\MLPmix}[0]{\texttt{MLP-mix}}
\newcommand{\PTResNet}[0]{\texttt{PT-ResNet}}
\newcommand{\PTSwinTFC}[0]{\texttt{PT-SwinT-FC}}
\newcommand{\PTENFC}[0]{\texttt{PT-ENetV2-FC}}
\newcommand{\PTSwinTMLP}[0]{\texttt{PT-SwinT-MLP}}
\newcommand{\PTENMLP}[0]{\texttt{PT-ENetV2-MLP}}

\newcommand{\pdf}[3]{f_{#1}^{#2} \left( {#3}  \right)}

\newcommand{\cdf}[3]{F_{#1}^{#2} \left( {#3}  \right)}

\newcommand{\blocco}[1]{} 

\theoremstyle{plain}
\newtheorem{theorem}{Theorem}[section]
\newtheorem{proposition}[theorem]{Proposition}
\newtheorem{lemma}[theorem]{Lemma}

\theoremstyle{definition}
\newtheorem{definition}[theorem]{Definition}

\theoremstyle{remark}

\newtcolorbox{mybox}[2][]
{
  colframe = #2!25,
  colback  = #2!25!white!25,
  left=1mm,
  top=1mm,
  #1,
  breakable
}

\newenvironment{theorembox}
   {\begin{mybox}{gray}\begin{theorem}}
   {\end{theorem}\end{mybox}}

\newenvironment{defbox}
   {\begin{mybox}{gray}\begin{definition}}
   {\end{definition}\end{mybox}}

\icmltitlerunning{Initial Guessing Bias}

\begin{document}

\twocolumn[
\icmltitle{Initial Guessing Bias: \\
How Untrained Networks Favor Some Classes }



\icmlsetsymbol{equal}{*}

\begin{icmlauthorlist}
\icmlauthor{Emanuele Francazi}{EPFL,Eawag}
\icmlauthor{Aurelien Lucchi}{Basel}
\icmlauthor{Marco Baity-Jesi}{Eawag}

\end{icmlauthorlist}

\icmlaffiliation{Basel}{Department of Mathematics and Computer Science, University of Basel, Switzerland}
\icmlaffiliation{EPFL}{Physics Department, EPFL, Switzerland}
\icmlaffiliation{Eawag}{SIAM Department, Eawag, Switzerland}

\icmlcorrespondingauthor{Emanuele Francazi}{emanuele.francazi@epfl.ch}

\icmlkeywords{Machine Learning, ICML}

\vskip 0.3in
]



\printAffiliationsAndNotice{}  

\begin{abstract}
Understanding and controlling biasing effects in neural networks is crucial for ensuring accurate and fair model performance. In the context of classification problems, we derive
a theoretical analysis demonstrating that the structure of a deep neural network (DNN) can condition the model to assign all predictions to the same class, even before the beginning of training, and in the absence of explicit biases. We prove that, besides dataset properties, the presence of this phenomenon, which we call \textit{Initial Guessing Bias} (IGB), is influenced by model choices including dataset preprocessing methods, and architectural decisions, such as activation functions, max-pooling layers, and network depth.
Our analysis of IGB provides valuable information for selecting network architectures and initializing models. We also highlight theoretical consequences, such as a breakdown of node-permutation symmetry, a violation of self-averaging
and non-trivial effects that depth has on IGB.
\end{abstract}

\section{Introduction}
\label{sec:introduction}

\begin{figure*}
    \centering
    \includegraphics[width=0.7\textwidth]{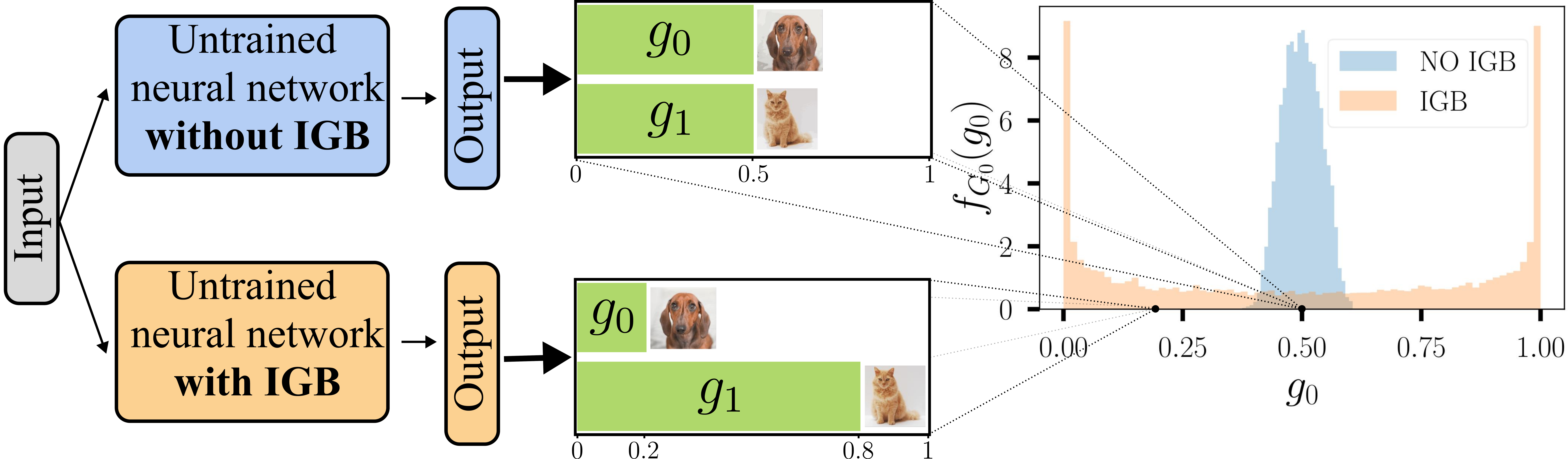}
    \caption{Initial Guessing Bias (IGB). Consider a task where we classify a binary dataset using an untrained network. Does it assign half of the examples to each class, or does it privilege one class?
    The answer depends on the model design. 
    In the top-left, we classify a binary dataset with an untrained network \textit{without IGB}. This model will generally assign half of the examples to each class (histogram on the top-center). 
    In the bottom-left, we classify the same dataset using an untrained network \textit{with IGB}. In this case, most of the guesses will usually go to one of the two classes (histogram on the bottom-center).
    As an example, we take the \texttt{dog/cat} classes (label $0$ / label $1$) from CIFAR10, and pass them through an untrained CNN with 2 layers, each followed by pooling. The non-IGB model uses $\tanh$ activations and average pooling, the IGB model uses ReLU and max pooling.
    We show in the center-right the distribution over different initializations, $\pdf{\RClassFraction{0}}{}{\ClassFraction{0}}$, of the fraction $\RClassFraction{0}$  of times that each model guessed \texttt{dog} (equivalently, $\RClassFraction{1}=1-\RClassFraction{0}$ indicates the fraction of images guessed as \texttt{cat}). While for the non-IGB models, $\RClassFraction{0}$ is most often 50\%, with IGB it most often is either 0\% or 100\%.
    }%
\label{IGB_diagram}
\end{figure*}

In the field of deep learning, the art of designing neural networks often traverses a terrain where empirical practices overshadow theoretical foundations. 
The design of DNNs involves a complex array of decisions, each of which can significantly influence the network's performance and learning dynamics \cite{liu2017survey, khanday2021taxonomy}. 
Choices such as data standardization, activation functions, and initial weight configurations, pivotal for network performance, are typically guided by heuristic methods due to limited theoretical insights.
A deeper theoretical understanding is crucial for developing more predictable and robust models.
Additionally, the ethical and fairness implications of biased model predictions have become crucial in responsible machine learning development \citep{fuchs2018dangers, parraga2022debiasing, siddique2023survey, louppe2017learning}.\newline
In this paper, we study how different choices in architecture design and data pre-processing influence the predictions of neural networks at initialization.
This leads us to the discovery of a previously unexplored phenomenon, which we illustrate in Fig.~\ref{IGB_diagram}, where \textbf{\textit{the initial predictions made by untrained neural networks are biased. We name this phenomenon \textit{Initial Guessing Bias} (IGB).}} IGB challenges naive assumptions about DNNs and informs crucial design decisions. Our study elucidates how, beyond effects that may be induced by the structure of the data itself, the design of the model plays a crucial role in shaping the initial predictive bias of the model. Specifically, our work makes the following contributions:
\begin{itemize}[leftmargin=*]
\setlength\itemsep{0.2em}
    \item \textbf{Identification and formalization of IGB:} We are the first to observe and formally articulate the concept of IGB. Its \underline{relevance} lies in:
    \begin{itemize}[leftmargin=*]
        \item[$\circ$] Showing that a model can be biased toward specific predictions, before it even saw the data it will be trained on.
        \item[$\circ$] Guiding critical design choices in terms of architecture, initialization, and data standardization.
        \item[$\circ$] Revealing a symmetry breaking and a violation of self-averaging, which are common working hypotheses.
        \item[$\circ$] Influencing the initial phase of learning dynamics, whose behaviour is affected by the level of IGB.
    \end{itemize}
    \item \textbf{Demonstration of IGB's robustness:} Through a comprehensive analysis, we establish the \underline{robustness} of IGB across various settings:
    \begin{itemize}[leftmargin=*]
        \item[$\circ$] We demonstrate the dependence of IGB on activation function choices and outline general rules for identifying IGB-inducing functions.
        \item[$\circ$] We uncover the role of max pooling in generating and intensifying IGB.
        \item[$\circ$] We show how data preprocessing procedures are crucial in activating and amplifying IGB.
        \item[$\circ$] We find that while network depth does not initiate IGB, it amplifies the bias when present.
        \item[$\circ$] We develop a theory that analytically describes IGB in MLPs with random data, encompassing the settings mentioned above.
        \item[$\circ$] We provide empirical evidence of the emergence of IGB in a broader range of practical scenarios, including real data, and a wide spectrum of architectures (e.g., CNNs, ResNets, Vision Transformers), demonstrating the prevalence of IGB.
        \item [$\circ$] Besides the theoretical setting employing untrained architectures, IGB is also present in the context of transfer learning, where only the head of the model is randomly initialized while the backbone is pre-trained.
    \end{itemize}
\end{itemize}

\section{Related work}
\subsection{Bias effects}
\label{sec:Rel_work_bias}

Bias in machine learning models is a multifaceted issue \citep{mehrabi2021survey}. While the term 'bias' often carries negative connotations, particularly when it leads to performance degradation \citep{francazi2023theoretical, engstrom2020identifying} or fairness concerns \citep{torralba2011unbiased}, it is important to recognize that not all forms of bias are inherently detrimental \citep{hagendorff2023we, pot2021not}. When controlled, biases can be essential to the learning process of a good model. The key lies in the ability to regulate these effects to avoid adverse outcomes.\newline
Most research in this domain has predominantly concentrated on biases stemming from dataset characteristics \citep{barocas2016big}, such as class imbalance \citep{francazi2023theoretical, ye2021procrustean, panigrahi2024comparing}, or arising due to specific algorithmic choices \citep{pessach2023algorithmic}. These forms of bias, while critical, represent only a part of the broader spectrum. The impact of biases resulting directly from model design decisions—such as network architecture, activation functions, and initialization strategies—remains comparatively unexplored.\newline
Our work contributes to this less-charted territory by examining how certain design choices in DNNs can introduce biases at the very onset of the training process. This perspective not only broadens our understanding of bias in machine learning but also emphasizes the need for a more comprehensive approach to model development that considers the potential effects of every design decision.\newline

\subsection{Properties of DNNs at initialization}
\label{sec:Rel_work_init}
Recently, the study of DNNs at the initialization stage has garnered increasing attention, given that the network's initial state can significantly influence the training process. For example, the initial distribution of the weights can determine an amplification/decay of the signal coming from the input, or even limit the depth to which signals can propagate through
random neural networks~\citep{schoenholz2016deep, hanin2018start, glorot2010understanding, saxe2013exact, orvieto2021vanishing, noci2022signal}. 
The initial state of a DNN also significantly impacts  generalization performance \citep{ramasinghe2023much}. 
While the study of the initial state of neural networks has received increasing attention over the past few years, our own study focuses on the initial bias and therefore differs from past work in some important key aspects which we detail in App.~\ref{sec:add_rel_wor}.

\section{Preliminaries} \label{sec:Imp_IGB}
\begin{figure*}[t]
    \centering    
    \includegraphics[width=.9\textwidth]{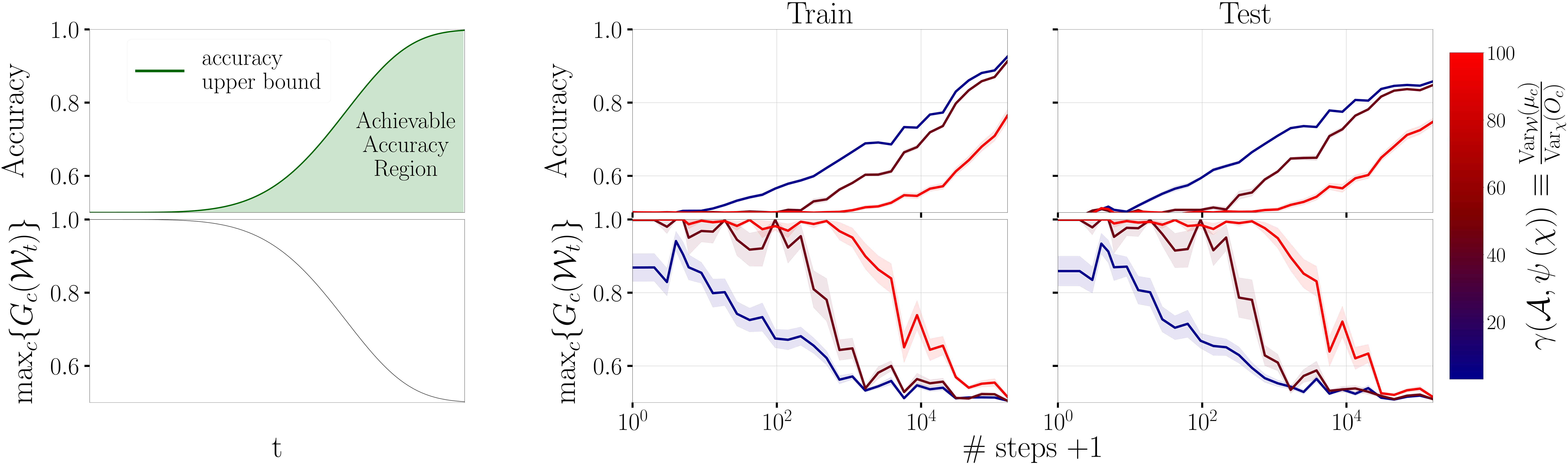}
    \caption{\textbf{Left}: IGB and Performance Bound: Diagram of the accessible performance range conditioned on the behavior of $\max_c \{ \RClassFraction{c} \left( \WeightSet{}_t \right) \}$ in a balanced binary dataset, in accordance with Eq.~\eqref{eq:acc_bound}. 
    \textbf{Right}: Comparison of the trend of $\max_c \{ \RClassFraction{c} \left( \WeightSet{}_t \right) \}$ with that of accuracy during the learning dynamics, varying with the level of guessing bias \underline{at initialization} (IGB). Particularly, $\VarRatio \in \mathbb{R}^+$ (colormap on the right) provides a measure of the level of IGB: the higher the value of $\VarRatio$, the higher the level of IGB (see Sec.~\ref{sec:BigPicture} for more details). The curves show a consistent pattern with the diagram on the left. They also demonstrate that the time for IGB absorption increases with the level of IGB itself. The simulations were conducted on an MLP-mixer using a binary dataset (dogs \textit{vs} cats from CIFAR) as input; more details on the setting and additional experiments with more architectures/datasets are provided in App.~\ref{app:dyn}.
    } \label{fig:Acc_bound}
\end{figure*}


\paragraph{Definition IGB}

We now articulate the concept of IGB and its practical implications. For clarity, we primarily focus on binary classification (so \scalebox{\BP}{$\NumberClasses =2$} classes). Our findings also apply to multiclass classification, as detailed in App.~\ref{sec:MC_extension}.

We start with an informal definition of the phenomenon of IGB. Consider a dataset $\Dataset$ and an architecture $\Arch$, parameterized by a set of weights. Specifically, we denote the configuration of the weight set at time $t$ as $\WeightSet{}_{t}$, with $\WeightSet{} \equiv \WeightSet{}_{0}$. An observable that characterizes the IGB phenomenon is the fraction of points classified, by the untrained model, as class $c$, and denoted by $\RClassFraction{c} \left( \WeightSet{} \right)$.

\begin{defbox}[IGB, informal]
 We say there is an Initial Guessing Bias if there is an imbalance in the initial fractions of points, captured by $\RClassFraction{c} \left( \WeightSet{} \right)$, assigned to a class.
\end{defbox}

\paragraph{Beyond initialization} As we will show in this paper, 
IGB is a pervasive phenomenon emerging in untrained models. Therefore, a natural question arises: what are the consequences of IGB in terms of training and generalization? Fully addressing this question and determining whether IGB has beneficial or detrimental effects on network training is complex as the answer might vary with the problem (data, architecture, optimizer, choices of hyperparameter, and so on might all have confounding effects). A comprehensive study of this kind is beyond the scope of this work. However, we can start with a simple observation; a discrepancy between the fractions of guesses, ${ \RClassFraction{c} \left( \WeightSet{}_t \right) }$, and the actual class proportions in the dataset, sets a limit on the achievable performance. For instance, considering a balanced dataset with $\NumberClasses$ classes, we can derive the following upper bound on the accuracy (Fig.~\ref{fig:Acc_bound} (left)):
\begin{equation}
\label{eq:acc_bound}
\text{Accuracy} (t) \leq 1- \left( \max_c \{ \RClassFraction{c} \left( \WeightSet{}_t \right) \} - \frac{1}{\NumberClasses} \right).
\end{equation}
Therefore, in the presence of strong IGB in balanced datasets, it is necessary for IGB to be absorbed during the learning process for the model to produce effective predictions. Note however, that while in this specific case IGB slows down training, it might instead be beneficial in other situations, \textit{e.g.} imbalanced training.\newline
Sec.~\ref{sec:QuantAn} will show how IGB can be activated and amplified in various ways, such as by acting on the architectural design or data pre-processing procedures.
In Fig.~\ref{fig:Acc_bound} (right), simulations on an MLP-mixer~\citep{tolstikhin2021mlp} for various levels of IGB are presented. Specifically, to maintain a constant architecture for comparison, we adjusted the IGB level by modifying the dataset standardization (for more details see App.~\ref{app:dyn}). The experiment reveals that the time required for IGB absorption (bottom plots), and consequently the improvement in performance (upper plots), increases with the level of IGB. This increase in absorption time correlated with the level of IGB is consistently observed in further experiments on different architectures, as detailed in App.~\ref{app:dyn}.
\paragraph{Methods for controlling IGB} These experiments highlight the importance of thoroughly understanding the phenomenon to formulate effective control strategies. The analysis we present offers several practical methods to regulate—either increase or decrease—IGB based on practitioners' needs. In particular, the main design choices outlined are:
\begin{itemize}
    \item \textbf{The choice of activation function}: Thm.~\ref{thm:act_rule_IGB}, and App.~\ref{sec:IGB_cond} analyse how the choice of the activation function determines the emergence of IGB. This offers practical guidelines on adapting activation functions to either mitigate or induce IGB. Detailed strategies are described in App.\ref{sec:subset_shift} and exemplified in Fig.\ref{fig:Comp_ReLU_SReLU}, showing that simply adding an offset to the ReLU function can significantly reduce IGB.
    \item \textbf{The choice of data standardization}: Thm.~\ref{thm:amp_IGB} describe how the choice of data pre-processing can be exploited to tune IGB. Beyond the theoretical derivation in App.~\ref{sec:Stand_effects}, we use these results in our experiments to adjust the level of IGB and compare different cases (see Fig.~\ref{fig:Acc_bound} Fig.~\ref{fig:dyn_exp_CD} and Fig.~\ref{fig:dyn_exp_Cifar}).
    \item \textbf{The choice of pooling layer}: Thm.~\ref{thm:amp_IGB}, and App.~\ref{sec:MP_effects} demonstrate how the max pool, depending on the chosen kernel size, can regulate the level of IGB, with larger kernel sizes amplifying the phenomenon.
    \item \textbf{The depth of the network}: Thm.~\ref{thm:amp_IGB}, and App.~\ref{sec:MLP} illustrate that while increasing the network's depth does not cause IGB to emerge, it can amplify IGB if it is already present.
\end{itemize}
Similarly, many other design choices can be analyzed using the framework presented to determine control strategies for the level of IGB. For example, the choice of the temperature of the SoftMax function (see App.~\ref{sec:sm_temp} for more details).
\section{Emergence of IGB: a first insight into the phenomenon}
\label{sec:BigPicture}

\paragraph{Setting and main notation}
Multi-Layer Perceptron (MLP) submodules are widely integrated into various architectures~\citep{he2016deep, vaswani2017attention,dosovitskiy2020image}. Moreover, recent advances, such as the MLP-mixer \citep{tolstikhin2021mlp}, highlight MLPs' capability to achieve competitive performance, suggesting their untapped potential, especially in large dataset applications. Our theoretical analysis centers on understanding the intricacies of MLP architectures. Formally, we consider an MLP as a set of $\DatasetSize$ inputs  $\{ \boldsymbol{\InputValue}^{(a)} \}_{a=1}^{\DatasetSize}$ that propagate through a set of $L$ hidden layers, until they reach the output layer, composed of a set of two nodes (one for each class), $\left\{ \ROutNode{c}{} \right\}_{c=0,1} \,$ (see App.~\ref{sec:Notation} for more details). Each input is classified by selecting the class $c$ with the largest output value.

Our main notation is described in Fig.~\ref{NN_scheme}--right, with data and weights modeled as follows:
\begin{itemize}[leftmargin=*]
    \item $\InputValue_b^{(a)} \sim \mathcal{N}(0,1)$ for each input component.
    \item $\Weights{l}{ij} \sim \mathcal{N} \left(0,\frac{\sigma_w^2}{\LayerNumNodes{l-1}}\right)$ for the weights, with all biases set to $0$ (see Eq.~\eqref{eq:h_prop_def}). $\LayerNumNodes{l}$ indicates the number of nodes of the $l^{\text{th}}$ hidden layer, while $\sigma_w^2$ is a constant that does not scale with the layer size (\textit{e.g.} the gain value).
\end{itemize}

The choice of random unstructured data as input is common in the literature~\cite{pennington2018spectrum, koehler2021uniform, loureiro2021learning, mignacco2020dynamical} as it simplifies theoretical analyses. In our case, this setting has the additional benefit of isolating the analysis from sources of IGB that are potentially embedded in the data structure. To support this assertion, we present empirical results using real data in App.~\ref{app:exp_data}. These results illustrate that correlations within the data can exacerbate the impact of IGB.
We follow the common practice of initializing bias parameters to zero, as noted in seminal works~\cite{glorot2010understanding, he2015delving}. By considering unstructured data \textit{i.i.d.} across different classes and DNNs with null biases, we create a highly symmetric setting, making the presence of predictive bias more counter-intuitive. Extending our analysis to include non-null bias parameters is straightforward and discussed in App.~\ref{App:Bias_ext}.
Similarly, App.~\ref{App:non_id_classes} discusses how the analysis can be extended to include datasets with classes that are not identically distributed.

We also note that the Gaussian initialization employed in our analysis is the standard Kaiming initialization \citep{he2015delving} but the analysis can be adapted to other initialization schemes. More specifically, our analysis applies to any set of independent weights $\{ \Weights{l}{ij} \}$ drawn from a centered distribution with variance $\mathcal{O}(1/\LayerNumNodes{l})$. 


\paragraph{Two kinds of averages}
In our analysis, the random variables (\textit{r.v.}s) we consider are essentially functions of two independent sources of randomness: the dataset, $\Dataset$, and the set of initialized weights, $\WeightSet{}$. We denote the cumulative distribution function (\textit{c.d.f.}) of a \textit{r.v.} $X$ as $\cdf{X}{}{x}$ and its probability density function (\textit{p.d.f.}) as $\pdf{X}{}{x}$. For variables dependent on both sources of randomness, we specify the active source in the notation. For instance, given a function of two independent sets of random variables, $X(\WeightSet{}, \Dataset)$:
\begin{align}
    \cdf{X}{(\Dataset)}{x} = \Prob{}{X<x | \WeightSet{}} \, , \, \, \pdf{X}{(\Dataset)}{x'} = \frac{d}{dx} \cdf{X}{(\Dataset)}{x}\Bigr|_{\substack{x=x'}},
\end{align}
represents the \textit{c.d.f.} and \textit{p.d.f.} of $X$ for a fixed set of weights $\WeightSet{}$. As we need to average over both the dataset and the weights, we use the concise notations
$$
\Einput{x} \equiv \CE{\Dataset}{x \mid \WeightSet{}} \quad \text{ and } \quad \Eweights{x} \equiv \CE{\WeightSet{}}{x}\,,
$$
to signify the expectations over these two distinct sources of randomness (refer to App.~\ref{sec:Notation} for more details).

\paragraph{Node Symmetry Breaking: the foundation of IGB}\label{sec:Int_IGB}
\begin{figure}
  \centering
    \includegraphics[width=0.4\textwidth]{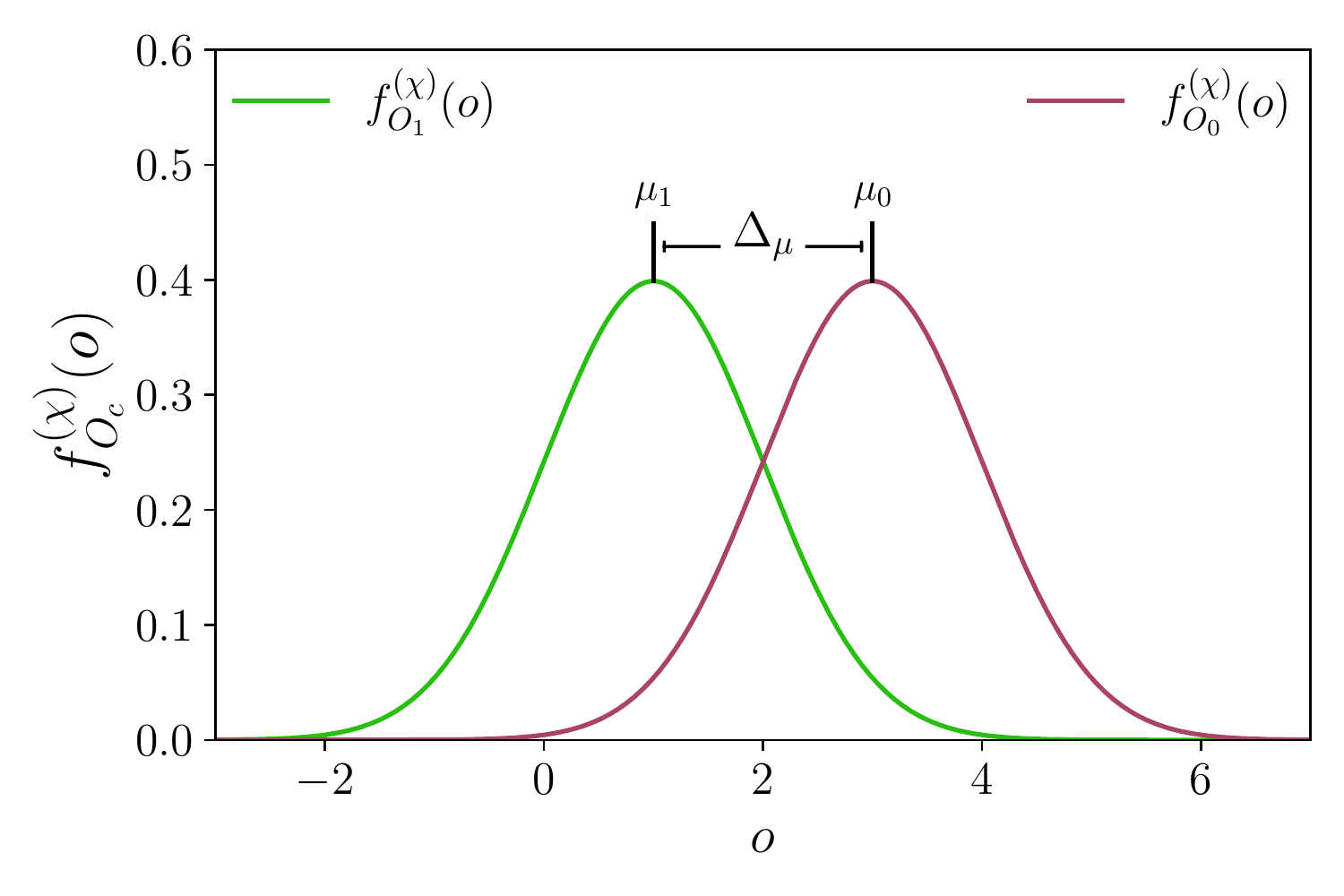}
    \vspace{-5mm}
 \caption{Illustration of the key quantities used in the analysis: 1) The green and purple curves represent the distributions of the two output nodes for a fixed set of network weights, $\WeightSet{}$, and 2) the mean of the distributions are denoted by $\RMean{c}$.}
\label{Gauss_Not}
\vspace{-1mm}
\end{figure}
We study how untrained neural networks assign data points to different classes before they start learning. 
In the absence of explicit bias weights, our intuition might suggest an even partitioning between classes. Instead, we demonstrate that the model design choices can induce an imbalance effect, resulting in a large fraction of data points being classified as a single class.
The key quantity in our analysis is the distribution (over  $\WeightSet{}$) of the fraction $\RClassFraction{c}\left( \WeightSet{} \right)$ of datapoints classified as class $c \in \{0,1 \}$. We have $\RClassFraction{c}=1/2$ in the absence of IGB, and a different value otherwise. 

Since our model's guess is assigned to the class with the largest output value, $\ROutNode{c}{}$,
in the limit of infinite datapoints,  the \emph{Law of large numbers} implies that
\begin{align}\label{eq:fi_def_LLN}
&\lim_{\DatasetSize \rightarrow \infty} \RClassFraction{c}\left( \WeightSet{} \right) = \Prob{}{\ROutNode{c}{}> \ROutNode{1-c}{} \mid \WeightSet{} } =\\
&\int_0^{\infty} \pdf{\RDiff{\ROutNode{c}{}}}{(\Dataset)}{ \Diff{\OutNode{}{}} } \; d\Diff{\OutNode{}{}}\, , \,\, \text{with} \, \, \RDiff{\ROutNode{c}{}} \equiv \ROutNode{c}{} - \ROutNode{1-c}{}. \notag
\end{align}
Without loss of generality, we will often use class 0 as a representative class, but the same discussion applies to any class. For class 0, Eq.~\eqref{eq:fi_def_LLN} simplifies to 
\begin{align}\label{eq:f0_def_LLN}
   \lim_{\DatasetSize \to \infty} \RClassFraction{0}\left( \WeightSet{} \right) = \Prob{}{ \RDiff{\ROutNode{}{}}>0 \mid \WeightSet{} } = \int_0^{\infty} \pdf{\RDiff{\ROutNode{}{}}}{(\Dataset)}{ x } \; dx\, ,  
\end{align}
with $\RDiff{\ROutNode{}{}} \equiv \ROutNode{0}{}-\ROutNode{1}{}$. For the sake of compactness, we define $\RMean{c} \equiv \Einput{\ROutNode{c}{}}$ and $\Delta_{\RMean{}} \equiv \RMean{0} - \RMean{1}$ as the difference between the distributions mean values. 
While $\RClassFraction{c}$ is a convenient and interpretable metric related to performance, our analysis can incorporate alternative measures that, for instance, allow for the analysis of the confidence level in the network's assignments, as shown in App.~\ref{app:conf_ass}.\newline
Eq.~\eqref{eq:f0_def_LLN} connects $\RClassFraction{0}$ (\textit{i.e.} the observable we are interested in) to the nodes variables $ \left\{ \ROutNode{c}{} \right\} $ (\textit{i.e.} the set of variables we analyze through our investigation).
The value of $\RClassFraction{0}$ depends on how often the output related to class $0$ has a higher value of the output than that of class $1$. More precisely, Eq.~\eqref{eq:f0_def_LLN} shows that $\RClassFraction{0}$ is determined by comparing the output distributions $\pdf{\ROutNode{c}{}}{(\Dataset)}{ \OutNode{}{}}$ related to each class $c$. We illustrate this in Fig.~\ref{Gauss_Not}. In the example of the figure, the output distribution related to class 0 is centered around higher values than that of class 1. Therefore, we will have $\RClassFraction{0}>\RClassFraction{1}$, indicating the presence of IGB.
In App.~\ref{app:proof_out_dist} we show that, for MLPs, \scalebox{\BP}{$\pdf{\ROutNode{c}{}}{(\Dataset)}{ \OutNode{}{}}$}, is asymptotically a Gaussian whose center, \scalebox{\BP}{$\RMean{c}$}, is itself a \textit{r.v.}, which is drawn from a Normal distribution $\pdf{\RMean{c}}{}{\Mean{}}$ that has a wide support:

\begin{theorembox}[Informal]
\label{thm:out_dist}
Consider a Gaussian distributed dataset processed through an MLP with $L$ hidden layers and weights initialized according to the Kaiming normal scheme (with null bias weights). In 
the limit of infinite width, the distribution of an output node $\ROutNode{c}{}$, at initialization, converges to:
\begin{align}
\label{eq:PO_asym}
& \pdf{\ROutNode{c}{}}{(\Dataset)}{ \OutNode{}{}} \xrightarrow{| \WeightSet{} | \rightarrow \infty} 
    \NormalDens{\OutNode{}{}}{\RMean{c}}{\CVar{\Dataset}{\ROutNode{c}{}}},
\end{align}
    where $|\WeightSet{}|$ indicates the cardinality of the set $\WeightSet{}$ and, for compactness, we denoted by $| \WeightSet{} | \rightarrow \infty$ the limit where the number of neurons, $\LayerNumNodes{l}$, of each hidden layer $l \in \{0, \dots, L \}$ tends to infinity. \newline
        Moreover, while the variance of the distribution, $\CVar{\Dataset}{\ROutNode{c}{}}$, converges with high probability (w.h.p.) to a deterministic value, the center of this distribution, $\RMean{c}$, is itself a \textit{r.v.}, varying from node to node and converging in distribution to:
\begin{align}
\label{eq:P_cen_asym}
& \pdf{\RMean{c}}{}{\Mean{}} \xrightarrow{| \WeightSet{} | \rightarrow \infty} \NormalDens{\Mean{}}{0}{\CVar{\WeightSet{}}{\RMean{c}}}.
\end{align}
\end{theorembox}

\begin{figure*}[]
    \centering    
    \includegraphics[width=.45\textwidth]{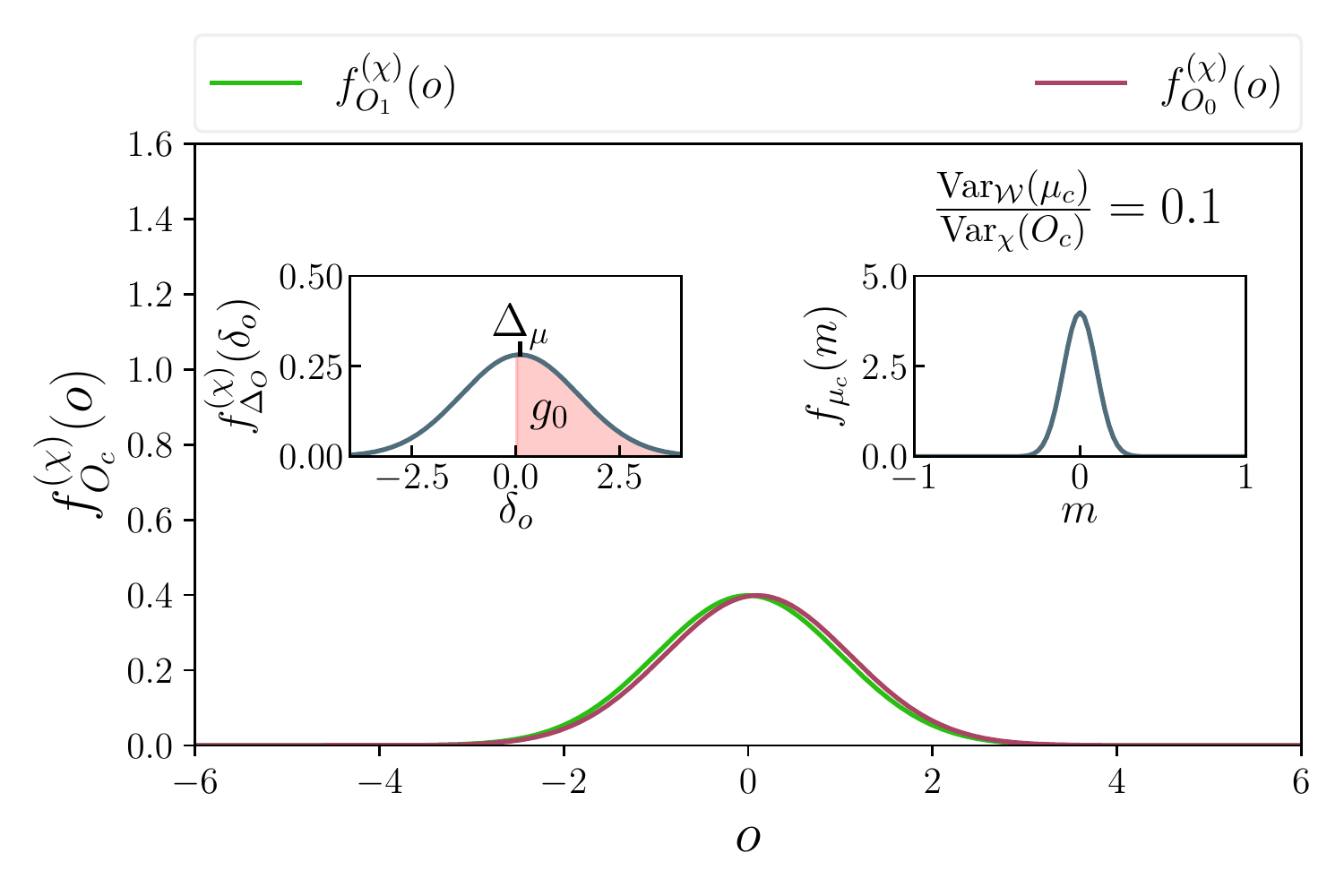}
    \includegraphics[width=.45\textwidth]{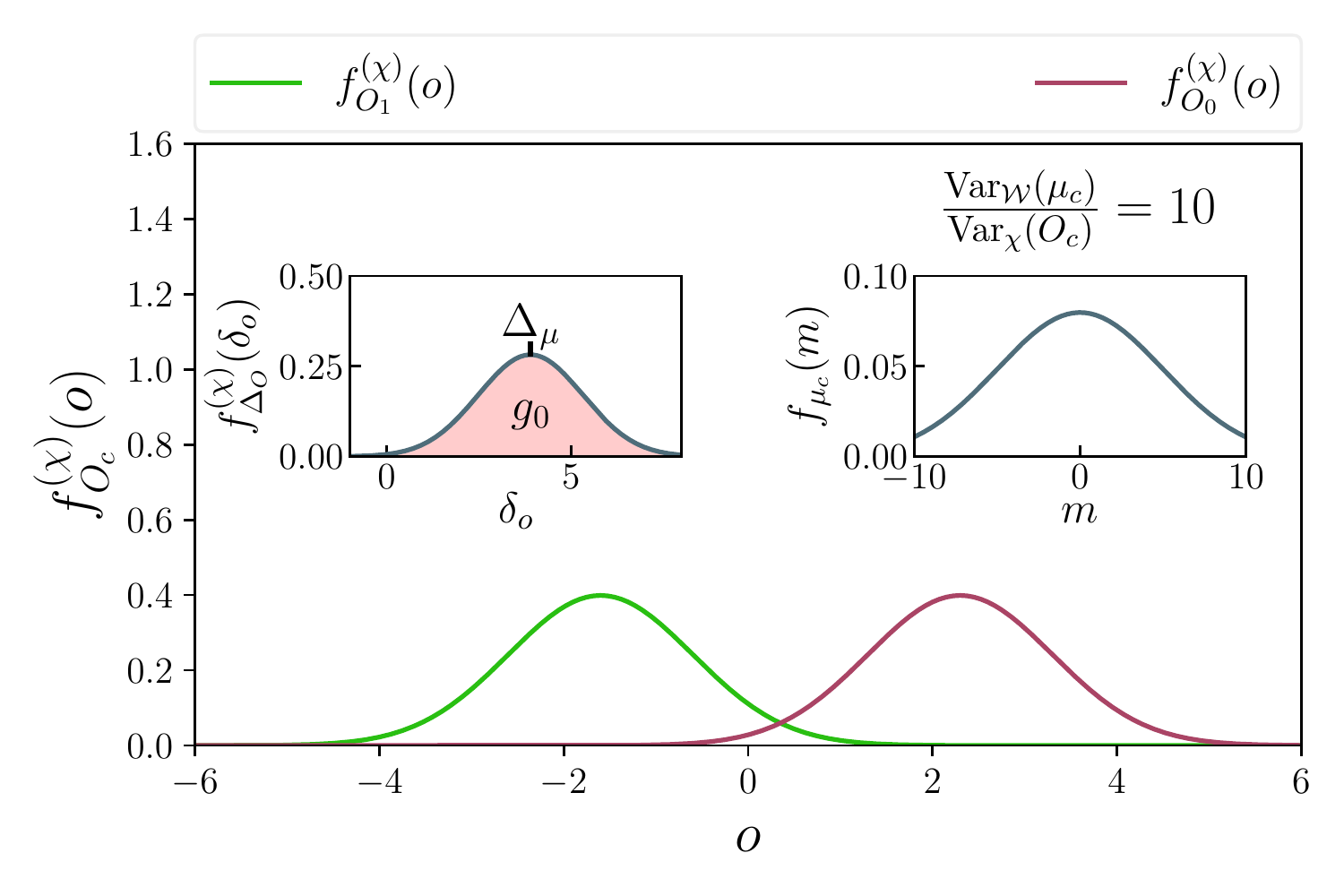}
    \vspace{-2mm}
    \caption{Comparison of two extreme scenarios: no IGB on the left, and strong IGB on the right. If the centers of the distributions, $\RMean{c}$,  have small fluctuations compared to the ones of the distributions \scalebox{\BP}{$\pdf{\ROutNode{c}{}}{(\Dataset)}{\OutNode{}{}} $}, the two distributions almost completely overlap, resulting in a similar probability that one output node exceeds the other (left). If, instead, the centers are typically much further apart than the fluctuations scale of the distributions \scalebox{\BP}{$\pdf{\ROutNode{c}{}}{(\Dataset)}{\OutNode{}{}}  $}, the values drawn from one distribution exceed the other one with high probability (right). Each plot contains two inset plots. The inset plot in the upper left represents the distribution of the difference of the \textit{r.v.}s shown in the main plot, (\scalebox{\BP}{$\RDiff{\ROutNode{}{}}$}). Note that, fixing the set $\WeightSet{}$ in a given experiment, and assuming a dataset big enough, \eqref{eq:f0_def_LLN} holds (the probability mass of the \textit{r.h.s.} is depicted with a red area bounded by the distribution and the integration extremes).
    The inset plot in the upper right shows instead \scalebox{\BP}{$\pdf{\RMean{c}}{}{\Mean{}}$} to give an idea of the fluctuations of $\RMean{c}$ for the two cases, measured also by the variances ratio reported above the inset plot.} \label{fig:Scheme_scenarios}
\end{figure*}

In other words, the outputs of different classes are distributed according to \textit{p.d.f.}s each centered on a different value:
as $\RMean{c}$ are \textit{r.v.}s, varying across output nodes, they are not all identically distributed. This difference results in a breakdown of permutation symmetry of nodes belonging to the same hidden layer (NSB). As we will explain now, this asymmetry is directly related to the emergence of IGB. \newline
In fact, the study of IGB can be conceptually summarized as a comparison between the fluctuations of \scalebox{\BP}{$\pdf{\RMean{c}}{}{\Mean{}}$}, which defines the distance between the Gaussians in Fig.~\ref{Gauss_Not}, and those of \scalebox{\BP}{$\pdf{\ROutNode{c}{}}{(\Dataset)}{ \OutNode{}{}} $}, which define how wide each of these Gaussians is. 
We now illustrate the two extreme cases to underline our point.
Starting from Eq.~\eqref{eq:fi_def_LLN}, we will discuss how the integral on the \textit{r.h.s.} varies in these two scenarios:
\begin{itemize}[leftmargin=*]
\item \textbf{Absence of IGB (Fig.~\ref{fig:Scheme_scenarios} (left))}: \newline
If the fluctuations of \scalebox{\BP}{$\pdf{\RMean{c}}{}{\Mean{}}$} are much smaller than the ones of \scalebox{\BP}{$\pdf{\ROutNode{c}{}}{(\Dataset)}{ \OutNode{}{}} $}, we will have two Gaussian \textit{r.v.}s centered almost on the same point, therefore, \scalebox{\BP}{$\Prob{}{ \ROutNode{0}{}> \ROutNode{1}{} \mid \WeightSet{} } = \Prob{}{ \RDiff{\ROutNode{}{}}>0 \mid \WeightSet{} } \simeq 1/2 $}.\newline
Indeed, the difference between two Gaussian \textit{r.v.}s is itself a Gaussian \textit{r.v.}, centered at the difference between the mean values of the original distributions, \scalebox{\BP}{$\RDiff{\RMean{}}$}.
If the fluctuations of $\RDiff{\RMean{}}$ are much smaller than those of $\RDiff{\ROutNode{}{}}$, we will typically have that \scalebox{\BP}{$ \pdf{\ROutNode{c}{}}{(\Dataset)}{ \OutNode{}{}} $} is a symmetric distribution centered very close to the origin. Therefore $\Prob{}{ \RDiff{\ROutNode{}{}}>0 \mid \WeightSet{} }  \simeq 1/2$
\textit{i.e.}, from Eq.~\eqref{eq:fi_def_LLN}, the fraction of points assigned to both classes is equal to 1/2.
\item \textbf{Strong IGB (Fig.~\ref{fig:Scheme_scenarios} (right))}: \newline
If, instead, the scale of $\pdf{\RMean{c}}{}{\Mean{}}$ fluctuations is much bigger than that of \scalebox{\BP}{$\pdf{\ROutNode{c}{}}{(\Dataset)}{\OutNode{}{}} $} we will typically fall in the opposing scenario where the two Gaussian distributions,  \scalebox{\BP}{$\pdf{\ROutNode{c}{}}{(\Dataset)}{\OutNode{}{}} $}, are well separated. We can assume, without loss of generality, that \scalebox{\BP}{$\Mean{0}>\Mean{1}$}. In this case we will have \scalebox{\BP}{$\Prob{}{\ROutNode{0}{} > \ROutNode{1}{} \mid \WeightSet{} } = \Prob{}{ \RDiff{\ROutNode{}{}}>0 \mid \WeightSet{} } \simeq 1 $}.
\end{itemize}

This difference between the last two scenarios suggests the following formal definition for IGB (we write it for a generic number $\NumberClasses$ of classes).

\begin{defbox}[IGB, formal]
\label{def:IGB}
Given an architecture $\Arch$ and a preprocessed dataset $\PreprData$, we have an absence of IGB if and only if, \textit{w.h.p.},
\begin{equation}
\lim_{\DatasetSize \rightarrow \infty} \RClassFraction{c} \left( \WeightSet{} \right) = \frac{1}{\NumberClasses}, \; \forall c \in \{0, \dots,   \NumberClasses-1\}.
\end{equation}

We instead have a presence of IGB if, in the limit of infinite data points ($\DatasetSize \rightarrow \infty$), we observe a disproportion between the values $\{ \RClassFraction{c} \}$ for different classes.
\end{defbox}
Note that, when $\DatasetSize$ is finite, there can be finite-size fluctuations which move $\RClassFraction{c} \left( \WeightSet{} \right)$ from its asymptotic value.

Consistently with the intuition presented in Sec.~\ref{sec:Int_IGB}, a possible measure of the level of IGB is given by the ratio of variances:
\begin{align}\label{eq:VarRatio_def}
   \VarRatio \equiv  \frac{\CVar{\WeightSet{}}{\RMean{c}}}{\CVar{\Dataset}{\ROutNode{c}{}}}\,.
\end{align}
In the absence of IGB, $\lim_{\DatasetSize \rightarrow \infty} \VarRatio  = 0$. The value of $\gamma$ increases with the level of IGB, providing a quantifiable metric to assess the extent of imbalance in initial guess fractions across classes.
In particular, we can use $\gamma$ to describe the limit of large IGB.

\begin{defbox}[Strong IGB]
Given a setting ($\Arch \,, \PreprData$), a model exhibits \textit{strong IGB} for such a setting if
\begin{align}
    \VarRatio = \infty \, .
\end{align}
\end{defbox}
 Note that, Eq.~\eqref{eq:f0_def_LLN} implies that, in the strong IGB limit,
\begin{align}\label{eq:delta_f0}
\VarRatio= \infty \Longrightarrow  \pdf{\RClassFraction{0}}{}{\ClassFraction{0}} = \tfrac{1}{2} \Dirac{\ClassFraction{0}} + \tfrac{1}{2} \Dirac{\ClassFraction{0}-1},
\end{align}
indicating that in each experiment, the dataset is completely classified as either belonging to class $0$ or class $1$.\newline


\section{IGB: emergence and amplification }
\label{sec:QuantAn}
By leveraging Theorem \ref{thm:out_dist}, we are able to theoretically estimate the \textit{p.d.f.} of $\RClassFraction{0}$, building upon the distributions defined in Eq.~\eqref{eq:PO_asym} and Eq.~\eqref{eq:P_cen_asym}. These estimations, described in App.~\ref{sec:SLP_analysis} and App.~\ref{sec:MLP}, enable us to systematically evaluate $\pdf{\RClassFraction{0}}{}{\ClassFraction{0}}$ across a spectrum of model designs. \newline
The key steps in the derivation are:
\begin{enumerate}
    \item[(A)] Deriving Eq.~\eqref{eq:PO_asym}, which leads to the \textit{p.d.f.} of $\RDiff{\ROutNode{}{}}$:
    \begin{align}\label{eq:P_DO}
        \pdf{\RDiff{\ROutNode{}{}}}{(\Dataset)}{\Diff{\OutNode{}{}}} = \NormalDens{\Diff{\OutNode{}{}}}{\RDiff{\RMean{}}}{2 \CVar{\Dataset}{\ROutNode{c}{}}} \, .
    \end{align}
    \item[(B)] Substituting Eq.~\eqref{eq:P_DO} into Eq.~\eqref{eq:f0_def_LLN} allows us to invert the equation linking $\RClassFraction{0}$ to $\RDiff{\RMean{}}$, thereby expressing $\RDiff{\RMean{}}$ as a function of $\RClassFraction{0}$.
    \item[(C)] Deriving Eq.~\eqref{eq:P_cen_asym}, which leads to the \textit{p.d.f.} of $\RDiff{\RMean{}}$:
    \begin{align}\label{eq:P_dmu}
        \pdf{\RDiff{\RMean{}}}{}{\Diff{\Mean{}}} = \NormalDens{\Diff{\Mean{}}}{0}{2 \CVar{\WeightSet{}}{\RMean{c}}}.
    \end{align}
    \item[(D)] Starting from Eq.~\eqref{eq:P_dmu} and the relationship $\RDiff{\RMean{}}(\RClassFraction{0})$ from point (B), we obtain $\pdf{\RClassFraction{0}}{}{\ClassFraction{0}}$ through a change of variables, that is, by applying the formula:
    \begin{align}\label{eq:f0_Dm_shift}
        \pdf{\RClassFraction{0}}{}{\ClassFraction{0}} \, d\ClassFraction{0} = \pdf{\RDiff{\RMean{}}}{}{\Diff{\Mean{}}(\ClassFraction{0})} \, d\Diff{\Mean{}} \, .
    \end{align}
\end{enumerate}

This analysis not only identifies the architectural elements critical to the emergence of IGB but also quantifies their influence.\footnote{We note that, to derive $\pdf{\RClassFraction{0}}{}{\ClassFraction{0}}$, we work in the infinite-width limit. This is however not a necessary condition for the appearance of IGB (see \textit{e.g.} experiments with finite-width models in App.~\ref{app:exp}), but rather an assumption that allows us to derive the theoretical curves.} 
%
\subsection{Single hidden layer}
\begin{figure}[t!]
  \begin{center}
    \includegraphics[width=0.45\textwidth]{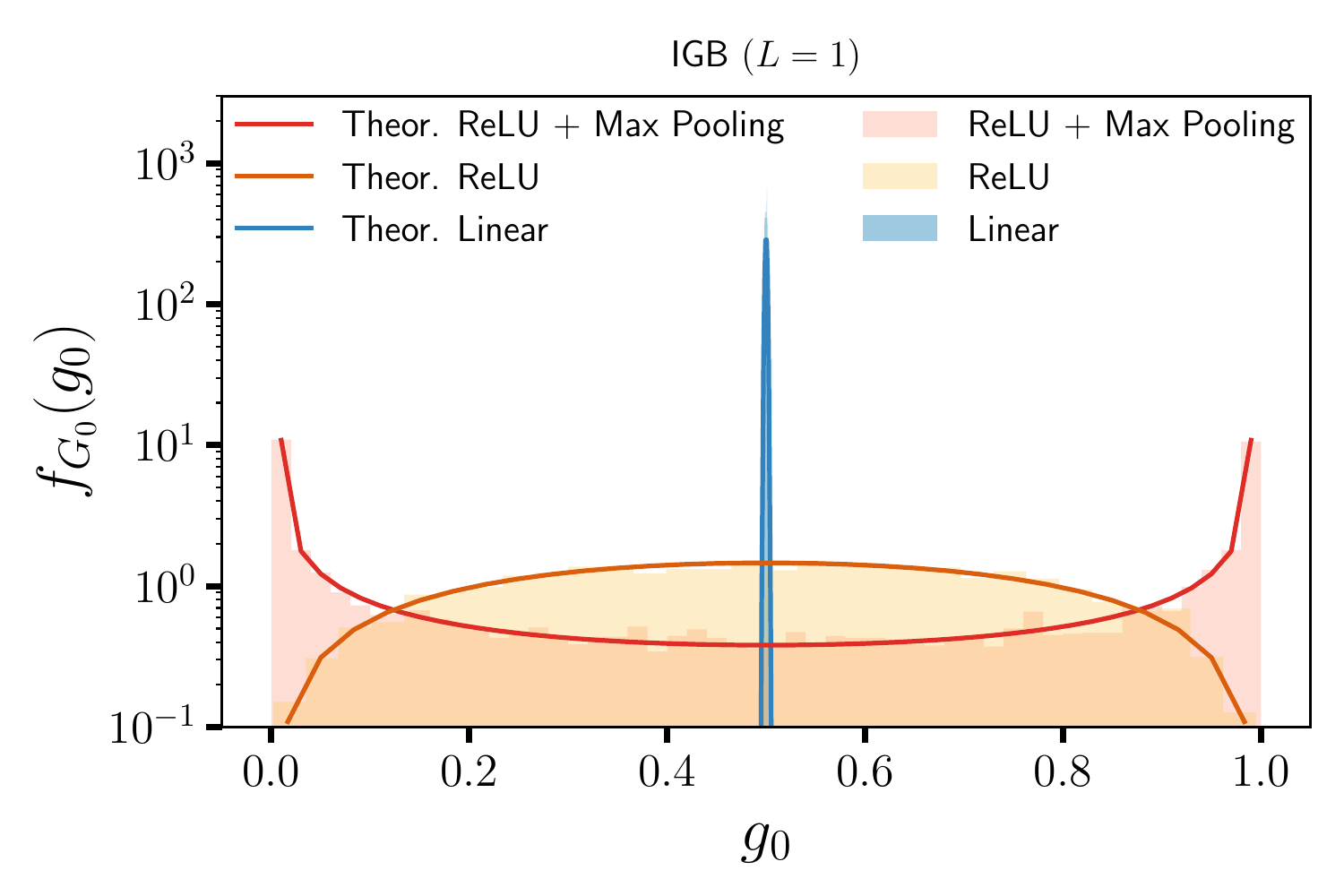}
  \end{center}
  \vspace{-6mm}
  \caption{The distribution $\pdf{\RClassFraction{0}}{}{\ClassFraction{0}}$ in a single-hidden-layer perceptron, for different choices of activation functions and with/without max pooling.}
  \vspace{-4mm}
\label{fig:Comp_SL_Pf0}
\end{figure}
Let us begin by considering the case with $L=1$; in this scenario, the element of non-linearity in the network (such as the activation function, pooling layer, etc.) can give rise to IGB.
In our analysis we consider three different setups to highlight the different behaviors induced by non-linearity elements. Fig.~\ref{fig:Comp_SL_Pf0} shows  $\pdf{\RClassFraction{0}}{}{\ClassFraction{0}}$ (both empirical histogram and theoretical curves) for the following cases: 
\begin{itemize}[leftmargin=*]
\setlength\itemsep{0.2em}
\item \textbf{Linear}: $\pdf{\RClassFraction{0}}{}{\ClassFraction{0}}$ asymptotically  converges to a delta distribution peaked on $\ClassFraction{0} = 1/2$.\footnote{By asymptotically, we mean in the ${\DatasetSize \rightarrow \infty}$ limit.} 
The theoretical curve shown in the plot takes into account the finite dataset size effects\footnote{To take finite dataset-size effects into account, we substitute the population data distribution, with the empirical one, defined over the finite set of $\DatasetSize$ points.} (since in real simulations $\DatasetSize < \infty$) as discussed in App.~\ref{sec:Dist_der}.
\item \textbf{ReLU}: In this case, $\pdf{\RClassFraction{0}}{}{\ClassFraction{0}}$ does not concentrate at $\ClassFraction{0} = 1/2$ and stays asymptotically wide (we detail this in App.~\ref{sec:Dist_der}, Fig.~\ref{D_Comparison}), so $\ClassFraction{0}$ will, \textit{w.h.p.}, be away from 1/2. 
The mode of the distribution remains, as for the linear case, at $\ClassFraction{0} = 1/2$. Yet, in this case, the fluctuations from the peak are not due to finite size effects and remain finite in the limit of infinite data points.
\item \textbf{ReLU + max pool}: With respect to the previous case we add a pooling layer defined as
\begin{align}
    \RCompHidNodeAMP{1}{k}{l} &= \MPOperator{k}{ \left\{ \AFOperator{ \RCompHidNodeBAF{1}{j}} \right\}_{j \in S_l^k}}  \,,
\end{align}
 where, in our notation, $\MPOperator{k}{\cdot}$ indicates a pooling function with kernel size equals to $k$, $S_l^k$ indicates the $l^\mathrm{th}$ subgroup of $k$ nodes. In particular, for the max pooling 
 \begin{align}
\RCompHidNodeAMP{1}{k}{l} \equiv \max_{j \in S_l^k} \left \{ \max \left(0, \RCompHidNodeAAF{1}{j}  \right) \right \}.
 \end{align}
As in the case of the ReLU, we have a wide distribution that does not concentrate in the limit of infinite data. The difference with the previous case is that now the distribution is peaked at the extremes of the support (we will elaborate more on this in Sec.~\ref{sec:PSB}); in this case, it is very likely that the untrained network will classify most of the dataset as belonging to only one of the two classes.
\end{itemize}
Note how, for the output layer, permutation symmetry in nodes corresponds to symmetry between classes. The latter is always preserved at the ensemble level,\footnote{Here, "ensemble" refers to all possible random initializations specified by the chosen scheme (e.g., Kaiming Normal).} in fact  in all three cases $\Eweights{\RClassFraction{0}\left( \WeightSet{} \right)} = 1/2$, where
$\Eweights{\RClassFraction{0}\left( \WeightSet{} \right)} \equiv \int \pdf{\RClassFraction{0}}{}{\ClassFraction{0}} \ClassFraction{0} \, d\ClassFraction{0}$ 
indicate the average of $\RClassFraction{0}\left( \WeightSet{} \right)$ over the ensemble of weight initializations.
However, while in the linear case, the symmetry between classes is also conserved on the single element of the ensemble,\footnote{In this case, we have $\lim_{\DatasetSize \rightarrow \infty} \RClassFraction{0}\left( \WeightSet{} \right) = \Eweights{\RClassFraction{0}\left( \WeightSet{} \right)} = 1/2$.} in the other two cases the single realization diverges, \textit{w.h.p.} from the mean over the ensemble. NSB thus results in a breaking of the symmetry between classes and a consequent self-averaging breakdown for the observable $\RClassFraction{0}$.
This difference between the two estimates is not due to finite network size effects; our predictions (theoretical curves in Fig.~\ref{fig:Comp_SL_Pf0}) are, in fact, formally exact in the limit of infinite size.
\vspace{-3mm}
\paragraph{Which activations cause IGB and which do not}
Our results can be extended to a generic activation function. In particular, we can clarify (App.~\ref{sec:IGB_cond}) what is the fundamental attribute of the activated nodes, \textit{i.e.} passed through the activation function and the pooling layer, $\{\RCompHidNodeAMP{1}{k}{i}\}_i$, that triggers NSB and consequently IGB:
\begin{theorembox}[Informal]
\label{thm:act_rule_IGB}
Consider a Gaussian distributed dataset processed through a single hidden layer perceptron, in the limit of infinite width hidden
 layer, and whose weights are initialized according to the Kaiming normal scheme (with null bias weights).         Then, we will have an absence of IGB in the network if, and only if, for the generic $i^{\text{th}}$ node of the hidden layer:
\begin{align}\label{eq:no_igbrho}
\Einput{\RCompHidNodeAMP{1}{k}{i}} = 0  \;, \;\; \forall i .
\end{align}

  Or conversely, IGB will emerge if, and only if
\begin{align}\label{eq:igbrho}
\Einput{\RCompHidNodeAMP{1}{k}{i}} \neq 0 \; , \;\;\forall i  .
\end{align}   
\end{theorembox}

Thus, the results we obtained on IGB apply to any kind of activation function: activations without IGB align with the description for linear activations, while those with IGB qualitatively resemble ReLU. 
\vspace{-3mm}
\paragraph{The effect of data pre-processing}
 Condition~\eqref{eq:igbrho} relies on data averages $\Einput{\ldots}$, which means IGB can be controlled by data standardization. Although our analysis until now centered on data around 0 (otherwise, we would not be able to attribute IGB to the mentioned architectural elements), other standardization methods (\textit{e.g.}, inputs in [0,1]) will induce IGB. In App.~\ref{sec:Stand_effects}, we focus on this aspect and demonstrate how the choice of input standardization can amplify IGB. Thus, in the experiments shown in Fig.~\ref{fig:Acc_bound} and App.~\ref{app:dyn}, we use standardization as way to tune IGB without changing the architecture design (which would instead result in incomparable training curves).
  
As a general guideline, antisymmetric activations exhibit no IGB when the data is centered around zero inputs. Condition~\eqref{eq:igbrho} also suggests that activations can be redefined to gain or lose their IGB property. For instance, in App.~\ref{sec:subset_shift}, we demonstrate that appropriately shifting ReLU functions can eliminate IGB.

\subsection{Conditions for strong IGB}\label{sec:PSB}
We are now interested in the conditions that lead an untrained model to assign \textit{all} the examples of a dataset to the same class (Strong IGB). There are in fact specific preprocessing and architecture choices that can amplify IGB (for example, in Fig.~\ref{fig:Comp_SL_Pf0}, IGB is exacerbated through max pooling). 
We can show that standardization, max pooling and network depth can all lead to \textit{strong IGB}.

\vspace{-1mm}
\begin{theorembox}[Conditions for strong IGB, informal]
\label{thm:amp_IGB}
Consider a centered Gaussian distributed dataset that undergoes preprocessing through the standardization 
\begin{equation*}
    \PreprDataOp{\boldsymbol{\InputValue^{(a)}}} = \boldsymbol{\InputValue^{(a)}} + \boldsymbol{K},
\end{equation*}
where $\boldsymbol{K}$ is an offset with the same dimensionality of the input.
The data is then processed through a Multi-Layer Perceptron (MLP) with \( L \) hidden layers. This MLP employs ReLU nonlinearities and max pooling layers with a kernel size of \( k \), and the weights are initialized using the Kaiming normal scheme (with zero biases).

In the limit of infinite width for hidden layers, the following scenarios are observed:
\begin{enumerate}[label=\Roman*.]
    \item  Shifting the center of the input distribution by a vector with the norm \( |\boldsymbol{K}| \), even when \( L = 1 \) and \( k = 1 \), results in: 
    \begin{equation}\label{eq:thm_inf_stand_ampl}
       \lim_{|\boldsymbol{K}| \rightarrow \infty}  \VarRatio = \infty.
    \end{equation}
    \item With an increase in the kernel size of the pooling layer, even for a single hidden layer (\( L = 1 \)), and in the absence of standardization offset (\( |\boldsymbol{K}| = 0 \)), one has:
    \begin{equation}
       \lim_{m \rightarrow \infty}  \VarRatio = \infty.
    \end{equation} 
    \item  
    In the infinite-depth limit, even in the absence of pooling layers (\( k = 1 \)) and standardization (\( |\boldsymbol{K}| = 0 \)):
    \begin{equation}
       \lim_{L \rightarrow \infty}  \VarRatio = \infty.
    \end{equation}

\end{enumerate}
\end{theorembox}
\vspace{-1mm}
The proofs for points I., II., and III. are detailed in Apps.~~\ref{sec:Stand_effects}, \ref{sec:MP_effects}, and~\ref{sec:deep_arc}, respectively. It is important to note a key distinction among the three sources of IGB amplification we analyzed. Our analysis reveals that while the choice of data standardization and pooling can both amplify and initiate IGB (for instance, leading to the emergence of IGB in models where it would not appear otherwise), increasing network depth can only amplify existing IGB in the model, but cannot initiate its onset.

\subsection{Experimental results}

\subsubsection{Robustness of IGB in real settings}
Beyond the theoretical analysis presented, in App.~\ref{app:exp} we present experiments demonstrating the presence of IGB in broader settings. In particular, we explore real data and various architectures including CNNs, ResNets, MLP-mixers, and Vision Transformers. These experiments demonstrate the presence of IGB in settings commonly used in practice. They also highlight that the use of structured data accentuates IGB, combining the influence of model design with that originating from dataset correlations. The experiments analyze binary as well as multi-class datasets, both of which are covered by our theory.\footnote{While we focus in our discussion on the binary case for the sake of clarity, refer to App.~\ref{sec:MC_extension}}

\subsubsection{Effects of IGB on Training Dynamics}
We also provide further experiments on the effect of IGB on learning dynamics with real data (both binary and multi-class) and different architectures (ResNet, MLP-mixer, and Vision Transformers), showing results qualitatively analogous to those shown in Fig.~\ref{fig:Acc_bound}. We consider the case of balanced datasets, where the presence of IGB translates into a discrepancy between the fractions of class labels and the corresponding fractions of model predictions. This mismatch poses a natural limit on the performance we can reach, as explained in Sec.~\ref{sec:Imp_IGB}. In the presence of IGB in these settings, it becomes crucial to reabsorb the predictive bias during the dynamics to achieve good performance. It then becomes important to determine the time required by the dynamics to reabsorb the bias. Our experiments show how this reabsorbing time is critically dependent on the level of IGB; with high levels of IGB, convergence can become dramatically slow. It can also be observed how the improvement in performance over the dynamics proceeds consistently with the absorption of Guessing Bias. Figs.~\ref{fig:Acc_bound}, \ref{fig:dyn_exp_CD}, and \ref{fig:dyn_exp_Cifar} show this by tracking both accuracy and the maximum among the class fraction of guesses, used as a measure of IGB. The comparison of these two trends shows that during the dynamics, the level of guessing bias reduces while the performance improves accordingly.

\subsubsection{Effects of IGB on Pre-Trained Models}
\begin{figure}
    \includegraphics[width=.48\textwidth]{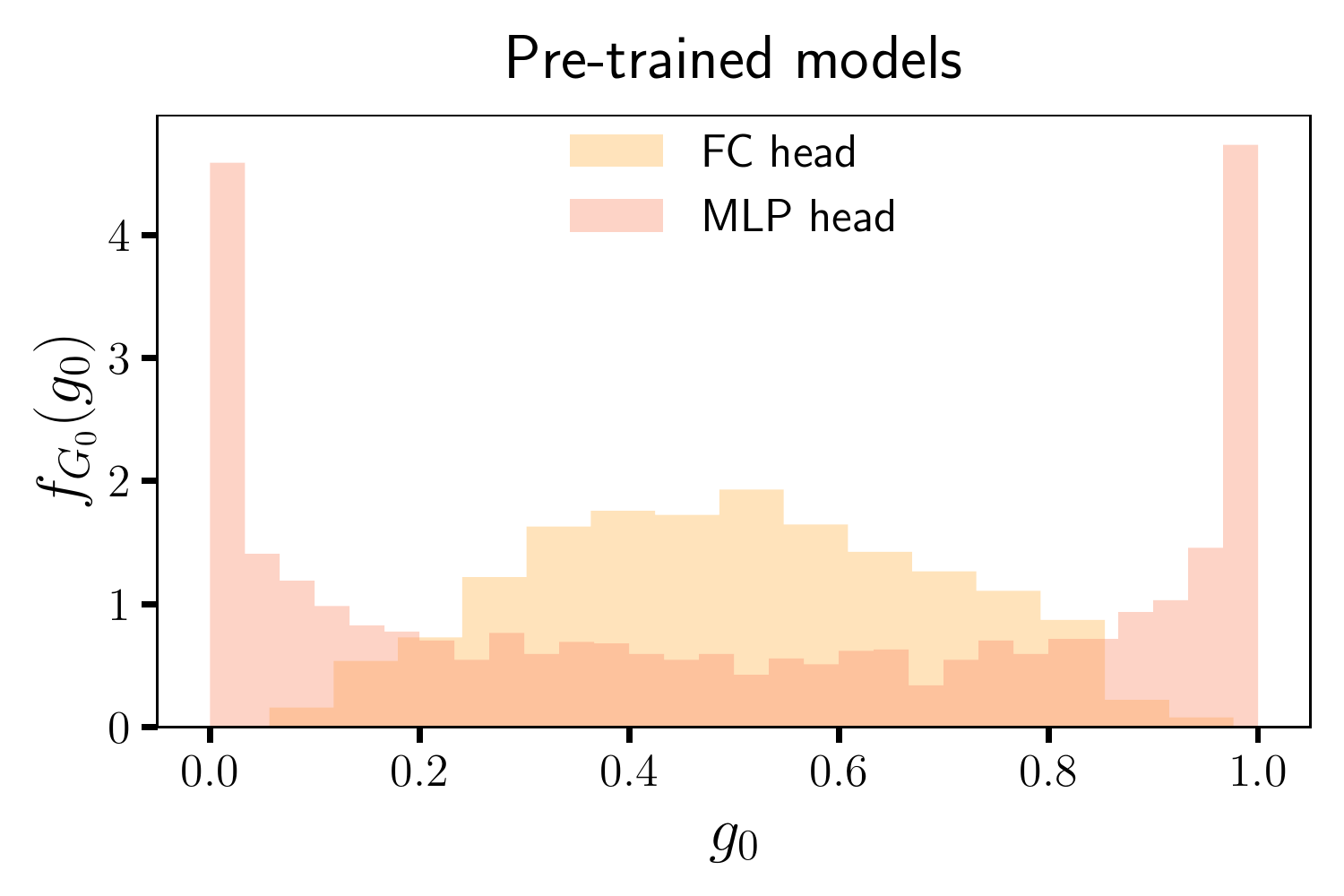}
    \vspace{-8mm}
    \caption{The distribution $\pdf{\RClassFraction{0}}{}{\ClassFraction{0}}$ on pre-trained EfficientNetV2 with the binary dataset (\CIFAR). The difference lies in the untrained head: a fully connected layer (\PTENFC) versus a MLP (\PTENMLP). More details in App.~\ref{sec:reprod}. Increasing head depth amplifies IGB, consistently with our analysis.
}
  \vspace{-2mm}
    \label{fig:ptm_comp}
\end{figure}

We show that in the context of transfer learning, IGB can persist in pre-trained models, where only the final layer(s) of the architecture are untrained. Experiments outlined in App.~\ref{app:pt-m} (see also Fig.~\ref{fig:ptm_comp}) show the presence of IGB in these models and its potential amplification, consistent with our analysis. Acknowledging the presence of IGB in this context is relevant, especially in few-shot learning \citep{wang2020generalizing}, where the fine-tuning dynamics do not involve a large number of steps/epochs.


\section{Discussion}\label{sec:discussion}

We examined the classification bias in untrained neural networks, uncovering a phenomenon named IGB that arises from architectural choices that break permutation symmetry among hidden and output nodes within the same layer. The conditions that trigger IGB are related to model design, including data preprocessing, network depth, the selection of activation function and  pooling. Factors such as max pooling, offsetted data standardization or network depth can exacerbate IGB to the extent that all dataset elements are assigned to a single class.\newline
While our analysis is based on random \textit{i.i.d.} data, we anticipate the presence of IGB in real datasets due to correlated data patches, increasing the likelihood of similar classifications. Hence, we expect stronger IGB in real datasets (as empirically confirmed in App.~\ref{app:exp_data}). Moreover, the hypothesis of unstructured data is instrumental in highlighting the effects induced solely by model design.

Due to its data and architecture-dependent nature, IGB remains a complex phenomenon that requires further investigation to fully understand its implications.   A comprehensive exploration of its effects across different scenarios remains an important area for future research.
Our experimental results in App.~\ref{app:dyn} demonstrate that IGB alters the training dynamics when using gradient-based methods. This alteration can sometimes be detrimental to training, but it can also be beneficial. For example, one could exploit IGB to combat data imbalance, or differences in the gradients related to different classes~\cite{francazi2023theoretical}. 
 Furthermore, many works set the dynamics in the small learning rate regime~\citep{francazi2023theoretical,sarao2020complex, tarmoun2021understanding}; also in this context, the presence of IGB could turn out to be relevant, since the dynamics are more bound to the initial state.


 \clearpage
\section*{Impact statement}
Our work identifies a source of prediction bias appearing at initialization in DNNs, and reveals that a model can be biased toward specific predictions, before it even saw the data it will be trained on. 
We also show that its presence has an effect on the early learning dynamics. This has, for example, an effect on hyperparameter tuning, which is part of model selection. Therefore, the choice of the final model could be influenced by biases.

By informing model selection, data preparation, and initial conditions, our results have the potential to enhance the training of machine learning models. 
Therefore our study not only advances theoretical knowledge but also promotes more considerate practices in the design of DNNs. It emphasizes the need to balance performance with fairness considerations.


\section*{Acknowledgements}
This work was supported by the Swiss National Foundation, SNF grant \# 196902.

\bibliographystyle{plainnat}
\bibliography{IGB_icml2024}

{

\newpage
\appendix

\counterwithin{figure}{section}
\onecolumn

\vbox{
  {\hrule height 2pt \vskip 0.15in \vskip -\parskip}
  \centering
  {\LARGE\bf Appendix\par}
  {\vskip 0.2in \vskip -\parskip \hrule height 0.5pt \vskip 0.09in}
}
\vspace{-5mm}

\newcommand\invisiblepart[1]{%
  \refstepcounter{part}%
  \addcontentsline{toc}{part}{\protect\numberline{\thepart}#1}%
}

\invisiblepart{Appendix}
\setcounter{tocdepth}{2}
\localtableofcontents


\section{Notation}\label{sec:Notation}
In this section, we meticulously present the setting and notation used in our analysis. Our theory examines Multi-Layer Perceptrons (MLPs), where the propagation of an input signal, denoted as $\boldsymbol{\InputValue}$, through the network is governed by the following pair of equations:
\begin{align}
\RCompHidNodeBAF{l+1}{i} &= \sum_{j} \Weights{l+1}{ij} \RCompHidNodeAMP{l}{k}{j} \, ,\label{eq:h_prop_def} \\
\RCompHidNodeAAF{l}{i} &= \AFOperator{\RCompHidNodeBAF{l}{i}} \, , \label{eq:g_prop_def} \\
\RCompHidNodeAMP{l}{k}{i} &= \MPOperator{k}{ \left\{ \AFOperator{  \RCompHidNodeBAF{l}{j}} \right\}_{j \in S_l^m}}, \label{eq:rho_prop_def}
\end{align}
where $l \in \{ 0, \dots, L+1 \}$, $ \AFOperator{\cdot} , : \mathbb{R} \rightarrow \mathbb{R}$ represents the activation function, $\MPOperator{k}{\cdot}, : \mathbb{R}^k \rightarrow \mathbb{R} $ denotes the pooling function with kernel size equals to $k$, $S_l^k$ indicate the $l^\mathrm{th}$ subgroup of $k$ nodes; $\RCompHidNodeAMP{0}{k}{i} \equiv \boldsymbol{\InputValue}$ and $\RCompHidNodeBAF{L+1}{c} \equiv \ROutNode{c}{}$. The principal notation is summarized in Fig.~\ref{NN_scheme}. Throughout this work, we adopt the following notation convention: upper-case letters (e.g., $\RCompHidNodeBAF{l+1}{i}$) refer to random variables, while lower-case letters (e.g., $\ArgHidNodeBAF{}{}$) denote their specific instances or realizations. For instance, $\ArgHidNodeBAF{}{} \sim \RCompHidNodeBAF{l+1}{i}$ indicates that $\ArgHidNodeBAF{}{}$ is a realization of the random variable $\RCompHidNodeBAF{l+1}{i}$.
\begin{figure}[h]
    \centering
    \includegraphics[width=0.95\columnwidth]{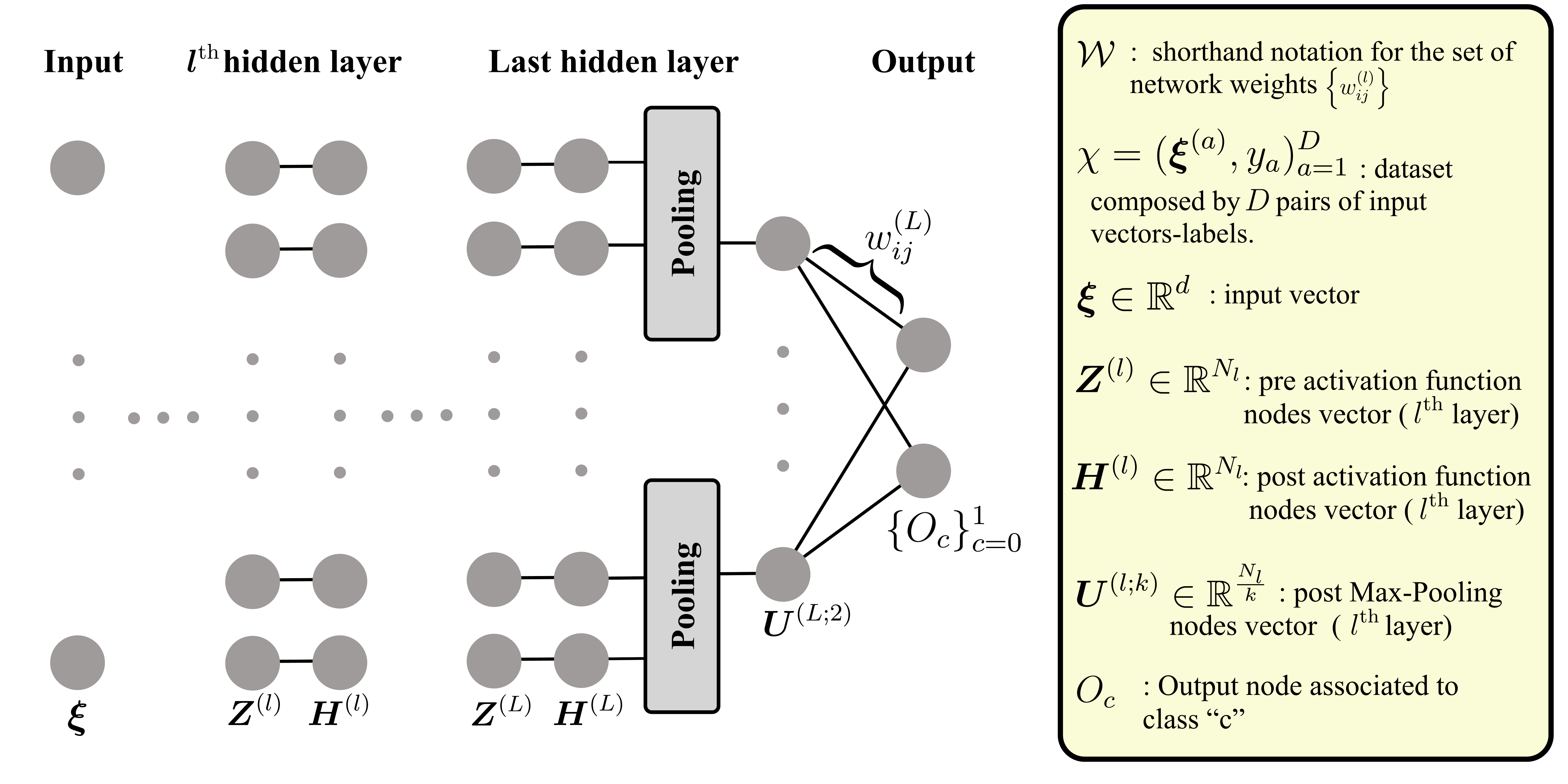}
    \caption{Illustration of a generic neural network for a binary classification problem (left) and main symbols (right).}
\label{NN_scheme}
\end{figure}
\newline
We also introduce the following notation:
\begin{itemize}
\item $\{ \cdot \}_{i=0}^{M-1}$: set of $M$ elements. If some of the indices of the variables are fixed (\emph{i.e.} equal for every element of the set), the set indices (indices that vary across different elements of the set) are reported explicitly on the right. If the index of the set elements is not explicitly reported, it means the absence of fixed indices for the set variables (\textit{i.e.} all indices are set indices).
\item $\CE{}{x}$: Indicate the expectation value of the argument, $x$. If the average involves only one source of randomness this is explicitly indicated, \emph{e.g.} $\CE{\Dataset}{x}$ indicates an average over the dataset distribution, while $\CE{\WeightSet{}}{x}$ an average over the distribution of network weights. For the sake of compactness, we will employ, where necessary the shorthand notation $\Einput{x} \equiv \CE{\Dataset}{x} $ and $\Eweights{x} \equiv \CE{\WeightSet{}}{x}$. 
\item $\erf\left( \cdot \right) $:  Error function. $\erf\left( x \right) \equiv \frac{2}{\sqrt{\pi}} \int_0^{x} e^{-t^2} dt$

\item $\cdf{X}{}{x}$: Given a random variable $X$, we denote its cumulative distribution function (\textit{c.d.f.}) as $\cdf{X}{}{x}$, where $\cdf{X}{}{x} = \Prob{}{X < x}$.

\item $\cdf{X}{(\Cdot)}{x}$: For random variables depending on multiple sources of randomness, we explicitly indicate the active source of randomness when one is fixed. For example, considering $X(\Dataset, \WeightSet{})$, a function of two independent sets of random variables $\Dataset$ and $\WeightSet{}$, $\cdf{X}{(\Dataset)}{x}$ denotes the \textit{c.d.f.} with respect to $\Dataset$, while holding $\WeightSet{}$ fixed. When $\Dataset$ has been marginalized, leaving only $\WeightSet{}$ as the active source, the subscript is omitted for conciseness, as there is no ambiguity.

\item $\pdf{X}{}{x'}$: Given a random variable $X$, $\pdf{X}{}{x'}$ denotes the probability density function (\textit{p.d.f.}) evaluated at $X = x'$. Formally, $\pdf{X}{}{x'} = \frac{d}{dx} \cdf{X}{}{x}\Bigr|_{\substack{x=x'}}$.

\item $\pdf{X}{(\Cdot)}{x'}$: For random variables with multiple sources of randomness, we specify the active sources of randomness as subscripts when some are fixed. For example, given $X(\Dataset, \WeightSet{})$, a function of two independent sets of random variables $\Dataset$ and $\WeightSet{}$, $\pdf{X}{(\Dataset)}{x'}$ represents the \textit{p.d.f.} of $X$ with respect to $\Dataset$, holding $\WeightSet{}$ fixed. When one of the sources of randomness, such as $\Dataset$, has been marginalized, there remains a single active source of randomness (e.g., $\WeightSet{}$). In such cases, the subscript can be omitted for simplicity and conciseness, as there is no ambiguity.

\item $\RMultiClassFraction{c}{}$: fraction of dataset elements classified as belonging to  class $c$. The argument $M$ indicates the total number of output nodes  for the variable definition, \emph{i.e.} the number of classes considered. For binary problems we omit this argument ($\RClassFraction{0}$) as there is only one non-trivial possibility, \emph{i.e.} $M=2$ .

\item $\RVecHidNodeAAF{l}$: vector of the $l^{\text{th}}$ hidden layer nodes, $\left( \RCompHidNodeAAF{l}{0} , \dots, \RCompHidNodeAAF{l}{\LayerNumNodes{l}-1}\right)$, after passing through the activation function; $\RCompHidNodeAAF{l}{i}$ indicate the component corresponding to node $i$.
\item $\RVecHidNodeBAF{l}$: vector of the $l^{\text{th}}$ hidden layer nodes, $\left( \RCompHidNodeBAF{l}{0} , \dots, \RCompHidNodeBAF{l}{\LayerNumNodes{l}-1}\right)$, before passing through the activation function; $\RCompHidNodeBAF{l}{i}$ indicate the component corresponding to node $i$.
\item $\NumberClasses \equiv \LayerNumNodes{L+1}$: number of output nodes, \emph{i.e.} the number of classes.
\item $\LayerNumNodes{l}$: number of nodes in the $l^{\text{th}}$ layer; $\LayerNumNodes{0}\equiv d$ indicates the dimension of the input data (number of input layer nodes) while $\LayerNumNodes{L+1}$ the number of classes (number of output layer nodes).

\item $\NormalDens{x}{\mu}{\sigma^2}$: Given a Gaussian \textit{r.v.} $X$, we indicate with $\NormalDens{x}{\mu}{\sigma^2}$ the \textit{p.d.f.} computed at $X=x$, \emph{i.e.}
$\NormalDens{x}{\mu}{\sigma^2} \equiv \pdf{X}{}{x} = \frac{e^{-\frac{1}{2 \sigma^2}(x-\mu)^2}}{\sqrt{2 \pi \sigma^2}}$

\item $\ROutNode{c,}{M}$: output layer node; $c$ is the node index; the index M, instead, indicate the total number of nodes considered. For binary problems we omit the subscript index to keep the notation lighter, \emph{i.e.} $\ROutNode{c}{}$ .
\item $\ROutNode{\tilde 0,}{M}$: the set $\left\{ \ROutNode{\tilde c,}{M} \right\}_{ c=0}^{M-1}$ contain the same elements of $\left\{ \ROutNode{\tilde c,}{M} \right\}_{\tilde c=0}^{M-1}$, but these are ranked by magnitude, such that  $\ROutNode{\tilde 0,}{M}$ is the greatest output value between the $M$, $\ROutNode{\tilde 1,}{M}$ the second one, and so on.
\item $\Prob{}{A}$: Denotes the probability associated with event $A$. $\Prob{}{A \mid B}$ indicates the probability of event $A$ given event $B$.

\item $\RVecHidNodeAMP{l}{k}$: vector of the $l^{\text{th}}$ hidden layer, $\left( \RCompHidNodeAMP{l}{k}{0} , \dots, \RCompHidNodeAMP{l}{k}{\ceil*{\frac{\LayerNumNodes{l}-1}{k}}}\right)$, after passing through the max pooling layer with kernel size $k$; $\RCompHidNodeAMP{l}{k}{i}$ indicate the component corresponding to node $i$.
\item $\Dataset = (\boldsymbol{\InputValue}^{(a)}, \LabelValue_a)_{a=1}^{\DatasetSize}$: dataset composed by $\DatasetSize$ pairs of input vectors-labels.
\item $\boldsymbol{\InputValue}^{(a)} \in \mathbb{R}^{d}$: $a^{\text{th}}$ input vector; when the index $a$ is omitted we mean a generic vector, $\boldsymbol{\InputValue}$, drawn from the population distribution.

\item $\CVar{}{\cdot}$: Indicate the variance of the argument. Since we have \textit{r.v.}s with multiple sources of randomness where necessary we will specify in the subscript the source of randomness used to compute the expectation. For example $\CVar{\Dataset}{\cdot} \equiv \Einput{\cdot - \Einput{\cdot}}^2$. For the sake of compactness, we will employ sometimes the shorthand notation $\CVar{\Dataset}{\cdot} = \sigma_{\cdot}^2$.

\item $\WeightSet{}_{t}$: shorthand notation for the set of network weights, $\{ \Weights{l}{ij} \}$ at time $t$; $\WeightSet{} \equiv \WeightSet{}_{0}$. We use, instead the notation $\WeightSet{l} \subseteq \WeightSet{}$ to indicate the subset of weights relative to a specific layer, \emph{i.e.} $\WeightSet{l} \equiv \{ \Weights{l}{ij} \}_{\substack{j \in [0, \dots, \LayerNumNodes{l}]\\i \in [0, \dots, \LayerNumNodes{l+1}]}}$. $\WeightSet{<l}, \; \WeightSet{>l}, \dots $ are defined analogously.

\item $\Weights{l}{ij}$: element $ij$ of the matrix $\WeightsMat{l}$, connecting two consecutive layers ($l \in [0, \dots, L]$). Given the matrix $\WeightsMat{l}$ we use a ‘placeholder’ index, $\Cdot$, to return column and row vectors from the weight matrices. In particular $\Weights{l}{j \Cdot}$ denotes row $j$ of the weight matrix $\WeightsMat{l}$; similarly, $\Weights{l}{\Cdot j}$ denotes column j.

\item $\RMultiClassRankedFraction{c}{}$: the set $\left\{ \RMultiClassFraction{c}{} \right\}_{c=0}^{M-1}$ contain the same elements of $\left\{ \RMultiClassRankedFraction{c}{} \right\}_{c=0}^{M-1}$, but these are ranked by magnitude, such that  $\RMultiClassRankedFraction{0}{}$ is the greatest output value between the $M$ elements of the set, $\RMultiClassRankedFraction{1}{}$ the second one and so on.

\item $\RMean{c} \equiv \Einput{\ROutNode{c}{}}$: This notation is a shorthand to represent the mean value of the output node $c$, computed with respect to the dataset, for a fixed set of initialized weights $\WeightSet{}$. In mathematical terms, it denotes the conditional expectation of the output node $c$, given a specific set of weights.

\blocco{\item $\WeightsVec{l}{i}$: row vector  $\left(\Weights{l}{i0}, \dots, \Weights{l}{i \LayerNumNodes{l}-1} \right)$ of the matrix $\WeightsMat{l}$, $\forall i \in [0, \dots,  \LayerNumNodes{l+1}-1]$.}

\item $\Heavyside{x}$: Heaviside step function. $\Heavyside{x}=1~\forall x\geq0$,~ $\Heavyside{x}=0~\forall x<0$. 
\item $\Dirac{x}$: Dirac delta function, defined by the relation $\int_{-\infty}^{\infty} f(x) \delta(x-x_0) dx = f(x_0)$.

\end{itemize}

\paragraph{Abbreviations}
\begin{itemize}
    \item \textit{c.d.f.}: cumulative distribution function
    \item CNN: Convolutional neural network
    \item DNN: Deep neural network
    \item IGB : Initial guessing bias
    \item MLP: Multi-layer perceptron
    \item NSB: Node symmetry breaking
     \item \textit{p.d.f.}: probability density function
     \item \textit{r.v.}: random variable
     \item \textit{w.h.p.}: with high probability
\end{itemize}

\section{Additional related work}\label{sec:add_rel_wor}
In the following, we discuss in detail some crucial aspects that distinguish our analysis from other investigations present in the literature.
\paragraph{Sources of randomness} 
A DNN at initialization can be interpreted as a random function parameterized by random weights, and whose inputs are sampled from an unknown data distribution.
There are thus two distinct and independent sources of randomness at initialization: weights and data. Previous theoretical studies of DNNs typically consider, for a given input, the whole ensemble of random initializations of the weights~\citep{poole2016exponential,matthews2018gaussian,novak2018bayesian}. In contrast, we fix the weight initialization and study the network's behavior by taking expectations over the data. Technically speaking, while previous work first averages over the weights and then over the data, our averages are first over the data and then over the weights. Note that this approach is closer to the natural order followed in practice, where data is classified by a single neural network (and thus for a single weight realization).


\paragraph{Breaking of self-averageness}
The mentioned inversion might seem a technicality but it actually constitutes a fundamental point in the study of IGB. One observable that characterizes the phenomenon is the fraction of points classified as class $c$, and denoted by $\ClassFraction{c} \left( \WeightSet{} \right)$,
(this quantity will be formally introduced shortly) 
in fact does not respect the self-averaging property~\citep{mezard1987spin,dotsenko1994introduction}.\footnote{In a system defined over an ensemble of realizations (in our case, each realization is a different weight initialization), a self-averaging quantity is one that can be equivalently calculated either by averaging over the whole ensemble or on a single, sufficiently large, realization; in such a situation, a single huge system is adequate to represent the entire ensemble.} 
In other words, even in the infinite dataset/network size limit, the value of $\RClassFraction{0} \left( \WeightSet{} \right)$ obtained on a given realization of the network initialization, $\WeightSet{}$, differs from that computed from the average over the ensemble of initializations. Self-averaging is often exploited (\textit{e.g.} in \textit{dynamical mean field theory}~\citep{sompolinsky1988chaos}, employed also in the context of mean-field theories of deep learning~\citep{schoenholz2016deep,poole2016exponential, xiao2018dynamical, yang2017mean, yang2019mean}), as it can lead to a simplification of the analysis. For the phenomena related to IGB  we cannot exploit self-averaging.
\paragraph{Nodes symmetry breaking} When we think of an untrained multi-layer perceptron (MLP), we are naturally inclined to assume a permutation symmetry among the various nodes of a given layer. Instead, we will show that this symmetry can be broken, and indeed, this symmetry breaking (\textit{i.e.} difference in the distribution of nodes in a given layer) constitutes the foundation of the IGB.
Nodes Permutation Symmetry Breaking (NSB) was already reported in previous work, \textit{e.g.} in specific shallow networks with a large number of examples~\citep{kang1993generalization}.\newline
We will see that the choice of architecture significantly influences the presence of IGB. Architecture design, particularly the selection of activation functions, has been extensively studied~\citep{dubey2022activation}, with a focus on ReLU versus differentiable activations such as sigmoid. For instance, sigmoid activations can achieve dynamical isometry, maximizing signal propagation depth~\citep{schoenholz2016deep}, unlike ReLUs~\citep{pennington2017resurrecting}. These activations are also compared in terms of generalization performance, revealing distinct behaviors for ReLUs and sigmoids~\citep{oostwal2021hidden}. Notably, these studies often consider averaging over weight initializations.

\section{Limit distributions}\label{sec:lim-dist}
This section reviews the convergence in distribution of a sequence of \textit{r.v.} to its  asymptotic distribution. We focus on the conditions that guarantee this convergence and the characterization of the resulting asymptotic distribution.

\subsection{Convergence in distribution}\label{sec:Dist_Conv}
Our discussion begins with the convergence of \textit{r.v.} combinations to their asymptotic distributions. We will examine the conditions under which this convergence occurs and how to define the resulting asymptotic distribution. The starting point is the \emph{Central Limit Theorem (CLT)} and its subsequent extensions.
\subsubsection{Central limit theorem}
The \emph{Central Limit Theorem} (CLT) plays a significant role in our analysis. Here, we provide a foundational discussion of the theorem, excluding proofs. For a more comprehensive exploration, refer to \citep{gnedenko1968limit, uchaikin2011chance, paul1999stochastic}. Consider a sequence of \emph{i.i.d.} \textit{r.v.}s $x_1, x_2, x_3, \dots$, each drawn from a population with a finite overall mean $\mu$ and variance $\sigma^{2}$. Denoting $\bar{x}_n$ as the sample mean of the first $n$ samples, the distribution's limiting form, $Z = \lim_{n \rightarrow \infty} \left( \frac{\bar{x}_n - \mu}{\sigma_{\bar{x}_n}} \right)$ where $\sigma_{\bar{x}_n} \equiv \frac{\sigma}{\sqrt{n}}$, converges to a standard normal distribution, \emph{i.e.}, $\NormalDens{z}{0}{1}$. For large but finite $n$, this represents the leading term of an expansion; corrections to the asymptotic Gaussian profile are $o\left( \frac{1}{\sqrt{n}} \right)$ as detailed in \citet{keller2001rate}.

\subsubsection{CLT extension}\label{sec:CLT_extention}
The CLT can be extended beyond the assumption of identically distributed variables. This extension is possible by applying conditions like the Lyapunov or Lindeberg conditions, which ensure the validity of the CLT under broader circumstances. For a sequence of independent \textit{r.v.}s $X_1, X_2, X_3, \dots$, each with mean $\mu_i = \CE{}{X_i}$, the Lindeberg condition is formulated as:
\begin{equation}
\lim_{n \rightarrow \infty} \sum_{i=1}^n \frac{1}{s_n^2} \int_{|x|\geq \epsilon s_n} (x-\mu_i)^2 dF_{X_i} (x) = 0\,,
\end{equation}
where $s_n^2 = \sum_{i=1}^n \CE{}{(X_i-\mu_i)^2} $ and $F_{X_i} (x)$ is the cumulative distribution function (c.d.f.) of $X_i$.

\begin{theorembox}[Lindeberg theorem]
For a set of independent \textit{r.v.}s $\{ X_i \}_{i=1}^n$, if Lindeberg’s condition holds for all positive  $\epsilon$, then, defining $S_n = X_1 + \dots + X_n$ and $Z_n = \frac{S_n - \sum_i \mu_i}{s_n}$, we have 
\begin{equation}
\pdf{Z_n}{}{z} \xrightarrow{n\rightarrow\infty} \NormalDens{z}{0}{1} \, .
\end{equation}

\label{th:Lindeberg}

\end{theorembox}

Our analysis will utilize an alternative set of necessary and sufficient conditions that guarantee the asymptotic convergence to a normal distribution. For further details, see \citet{petrov2012sums}.

Firstly, Thm.~\ref{thm:DistrConv} introduces necessary and sufficient conditions for the convergence of \textit{r.v.} sums to a Gaussian distribution. Subsequently, Thm.~\ref{thm:DistrConv_SC} outlines a set of sufficient conditions, which are particularly applicable to our cases of interest. We will demonstrate that \textit{r.v.}s satisfying these sufficient conditions also fulfill the requirements of Thm.~\ref{thm:DistrConv}.
\begin{theorembox}[Distribution convergence]

Let us consider a set of independent zero-mean \textit{r.v.}s $\{ X_i \}_{i=1}^n$.\footnote{In general if $\CE{}{X_i} \neq 0$ we can define a new set of variables $\{ \left(X_i-\CE{}{X_i} \right) \}$.}  If and only if, for every fixed $\epsilon>0$, the following conditions are satisfied:

\begin{align}
& \text{Concentration:} \notag \\
& \sum_{i=1}^{n} \Prob{}{  |X_i| \geq \epsilon } \xrightarrow{n\rightarrow\infty} 0 \;\;\;\; \forall \epsilon \in \mathbb{R}^+ \label{eq:concentration_cond}\,,
 \\ 
 & \text{Mean Normalization:} \notag \\
 &\sum_i
\left(  \int_{|x|<\epsilon} x \, \pdf{X_i}{}{x} dx \right)
 \xrightarrow{n\rightarrow\infty} 0 \label{eq:MeanNorm_cond}\,,\\
& \text{Variance Normalization:} \notag  \\
& s_n^2 = \sum_i
\left( \int_{|x|<\epsilon} x^2 \, \pdf{X_i}{}{x} dx - \left( \int_{|x|<\epsilon} x\, \pdf{X_i}{}{x} dx \right)^2 \right)
 \xrightarrow{n\rightarrow\infty} \sigma^2\,,  \label{eq:VarNorm_cond} 
\end{align}
the distributions of the sum $\sum_i X_i$ will converge to $\mathcal{N}\left( 0, \sigma^2  \right)$.

\label{thm:DistrConv}

\end{theorembox}
For the proof of Thm.~\ref{thm:DistrConv}, refer to Chapter 4 of \citet{petrov2012sums}. \newline

\begin{theorembox}[Sufficient condition for Thm.~\ref{thm:DistrConv}]

Let us consider a set $\{ X_i \}_{i=1}^n$ of independent, zero-mean,\footnote{In general if $\CE{}{X_i} \neq 0$ we can define a new set of variables $\{ \left(X_i-\CE{}{X_i} \right) \}$.} \textit{r.v.}s satisfying, for some constants $\tilde \sigma^2 \in \mathbb{R}^+$,  the following conditions
\begin{align}
& \text{Variance convergence:} \notag  \\
& \sum_{i=1}^n \CE{}{X_i^2 - \CE{}{X_i}^2} \xrightarrow{n\rightarrow\infty} n \tilde \sigma^2 \label{eq:Var_conv_cond} \,, \\
& \text{Fast decreasing tails: } \notag  \\ 
& \lim_{x \rightarrow \pm \infty}\, \pdf{X_i}{}{x} = \mathcal{O} \left( \frac{1}{x^4} \right) , \;\; \forall i \label{eq:fast_tail_dec_cond}\,.
\end{align}

Let us define the new set $\{ \tilde{X}_i \}_{i=1}^n$ such that 
\begin{equation}
\tilde{X}_i \equiv \frac{X_i}{c \sqrt{n}}\,,
\end{equation}

where $c>0$ is a constant.
We now proof that the set $\{ \tilde{X}_i \}_{i=1}^n$ satisfy \eqref{eq:concentration_cond}, \eqref{eq:MeanNorm_cond}, \eqref{eq:VarNorm_cond} leading to the convergence in distribution of $\tilde{S}_n = \sum_i \tilde{X}_i$ to $\mathcal{N}\left( 0, \sigma^2  \right)$.

\label{thm:DistrConv_SC}

\end{theorembox}

\begin{proof}
We will now prove that, starting from Eqs.~\eqref{eq:Var_conv_cond} and \eqref{eq:fast_tail_dec_cond}, the conditions of Thm.~\ref{thm:DistrConv} holds.\newline
We start showing the  Concentration condition [\eqref{eq:concentration_cond}]
\begin{align}
\begin{split}
&\sum_{i=1}^{n} \Prob{}{  |\tilde{X}_i| \geq \epsilon } = \sum_{i=1}^{n} \left( \int_{- \infty}^{-\epsilon} \pdf{\tilde{X}_i }{}{\tilde{x} } d\tilde{x} +\int_{\epsilon}^{\infty} \pdf{\tilde{X}_i }{}{\tilde{x} } d\tilde{x}  \right) = \\
& \sum_{i=1}^{n} \left( \int_{- \infty}^{- c \sqrt{n}\epsilon} \pdf{X_i}{}{x}  dx +\int_{c \sqrt{n}\epsilon}^{\infty} \pdf{X_i}{}{x}  dx  \right) \stackrel{n \gg 1}{=} \\
& \sum_{i=1}^{n} \left( \int_{- \infty}^{-c \sqrt{n}\epsilon} \mathcal{O} \left( \frac{1}{x^4} \right) dx +\int_{c \sqrt{n}\epsilon}^{\infty} \mathcal{O} \left( \frac{1}{x^4} \right) dx  \right) = \sum_{i=1}^{n} \left( \mathcal{O} \left( \int_{- \infty}^{-c \sqrt{n}\epsilon} \left| \frac{1}{x^4} \right| dx \right) + \mathcal{O} \left( \int_{c \sqrt{n}\epsilon}^{ \infty} \left| \frac{1}{x^4} \right| dx \right) \right) = \\
& \sum_{i=1}^{n} \mathcal{O} \left( \frac{1}{n^{\frac{3}{2}}}  \right) = \mathcal{O} \left( \frac{1}{n^{\frac{1}{2}}}  \right)
\xrightarrow{n\rightarrow\infty} 0
.
\end{split}
\end{align}
In the  third line we used the \emph{Fast decreasing tails} condition [Eq.~\eqref{eq:fast_tail_dec_cond}]. \newline
Now we show the validity of mean normalization condition [Eq.~\eqref{eq:MeanNorm_cond}]: \newline
\begin{align}
\begin{split}
&\sum_i^n
\left(  \int_{|\tilde{x}|<\epsilon} \tilde{x} \, \pdf{\tilde{X}_i }{}{\tilde{x} } d\tilde{x} \right) \stackrel{n \gg 1}{=} \sum_i^n \underbrace{\CE{}{\tilde{X}_i}}_{=0} - \sum_i^n \left( \mathcal{O} \left( \int_{- \infty}^{-c \sqrt{n}\epsilon} \frac{1}{c \sqrt{n}}\left| \frac{1}{x^3} \right| dx \right) + \mathcal{O} \left( \int_{c \sqrt{n}\epsilon}^{ \infty} \frac{1}{c \sqrt{n}} \left| \frac{1}{x^3} \right| dx \right) \right) = \\
& = \mathcal{O} \left( \frac{1}{n^{\frac{1}{2}}} \right) \xrightarrow{n\rightarrow\infty} 0\,.\label{eq:MeanConvergence}
\end{split}
\end{align}

Finally, for the Variance normalization condition [Eq.~\eqref{eq:VarNorm_cond}], we can now replace Eq.~\eqref{eq:MeanConvergence} into the definition of the variance. This gives
\begin{align}
& \sum_i \left( \int_{|\tilde x|<\epsilon} \tilde x^2 \,\pdf{\tilde{X}_i }{}{\tilde{x} } d\tilde x - \left( \int_{|\tilde x|<\epsilon} \tilde x \, \pdf{\tilde{X}_i }{}{\tilde{x} } d\tilde x \right)^2 \right) \stackrel{n \gg 1}{=} \\
& \frac{\tilde \sigma^2}{c^2} - \sum_i^n \left( \mathcal{O} \left( \int_{- \infty}^{-c \sqrt{n}\epsilon} \frac{1}{c^2 n} \left| \frac{1}{x^2} \right| dx \right) + \mathcal{O} \left( \int_{c \sqrt{n}\epsilon}^{ \infty} \frac{1}{c^2 n} \left| \frac{1}{x^2} \right| dx \right) \right) + \mathcal{O} \left( \frac{1}{n} \right) =\\ &\frac{\tilde \sigma^2}{c^2} +  \mathcal{O} \left( \frac{1}{n^{\frac{1}{2}}} \right) \xrightarrow{n\rightarrow\infty} \frac{\tilde \sigma^2}{c^2}.
\end{align}
\end{proof}

\subsection{Concentration of \textit{r.v.}s distribution} \label{sec:LayerConcExample}
When analyzing a set of $n$ \textit{r.v.}s whose variances scale with $n$, we encounter phenomena related to the concentration of their sum's distribution. In App.~\ref{sec:RescGaussVar}, we examine examples pertinent to our study, illustrating various scenarios based on the scaling differences. We will explore cases where the distribution remains asymptotically stable and others where it concentrates, \textit{i.e.}, narrows around a single value.

Then we apply these results to analyze the distribution $\pdf{  \RCompHidNodeBAF{l+1}{i}}{(\Dataset)}{\ArgHidNodeBAF{}{}}$; fixed $\WeightSet{}$, in fact, $\RCompHidNodeBAF{l+1}{i}$ can be expressed as a combination of \textit{r.v.}s. We will demonstrate how the nodes $\left\{ \RCompHidNodeBAF{l+1}{i}\right\}_{i}$ may not be identically distributed for $l\geq1$.

\subsubsection{Combination of random variables}\label{sec:RescGaussVar}
We approach the problem step by step, starting with a simplified case to demonstrate how the scaling of the variance of independent random variables is crucial in the distributional convergence of their sum. As we will see, this preliminary example contains key elements that we will extend to cases of interest for our analysis.\newline
Consider a set of independent \textit{r.v.}s $ \{ X_i \}_{i=1}^n$ where $X_i \sim \mathcal{N} \left( 0, \frac{\sigma^2}{n}  \right), \; \forall i$, and $\sigma^2$ does not scale with $n$, \emph{i.e.}, $\sigma^2 = \mathcal{O} (1)$. Our interest lies in analyzing the sum's distribution, particularly to ascertain if it exhibits asymptotic concentration phenomena. Define $S_x^{(n)} = \sum_{i=1}^n X_i$. To determine if the distribution of $S_x^{(n)}$ narrows as $n \rightarrow \infty$, we examine the standard deviation as an estimate for the fluctuation magnitude from the mean. From $X_i$'s definition and variance additivity, we derive that 
\begin{equation}
\sigma_{S_x^{(n)}} = \sqrt{\CE{}{ S_x^{(n)} - \CE{}{S_x^{(n)}}}^2 } = \mathcal{O} (1) .
\end{equation}
This implies that the sum's distribution remains asymptotically stable, meaning it does not narrow around the mean value.

Now, consider the \textit{r.v.} $S_{x^2}^{(n)} = \sum_{i=1}^n X_i^2$. Given that for a Gaussian variable $\sigma_{x^2}^2 = \mathcal{O} \left( \sigma_{x}^4 \right)$, and employing the variance's additivity, it follows that 
\begin{equation}
\sigma_{S_{x^2}^{(n)}} = \sqrt{\CE{}{ S_{x^2}^{(n)} - \CE{}{S_{x^2}^{(n)}}}^2 } = \mathcal{O} \left( \frac{1}{\sqrt{n}} \right).
\end{equation}
In this case, there is a concentration phenomenon; the fluctuations of $S_{x^2}^{(n)}$ asymptotically approach zero, \emph{i.e.}, the distribution's measure concentrates around the mean value.

We can proceed similarly in the presence of linear combinations of \textit{r.v.}s, where we again consider the sum of \textit{r.v.}s, this time each coupled with a fixed coefficient.\newline
To examine the changes in node distribution from one layer to the next, we analyze the distribution of $\RCompHidNodeBAF{l+1}{i} = \sum_j \Weights{l}{ij} \RCompHidNodeAAF{l}{j}$, with a fixed set $\WeightSet{}$. This leads to a linear combination of \textit{r.v.}s. It is straightforward to show that the hypotheses of Thm.~\ref{thm:DistrConv_SC} are satisfied in this context. We then have
\begin{equation}\label{eq:Ph^l_W}
\pdf{\RCompHidNodeBAF{l+1}{i}}{(\Dataset)}{\ArgHidNodeBAF{}{}} = \NormalDens{\ArgHidNodeBAF{}{}}{\sum_j \Weights{l}{ij} \Einput{\RCompHidNodeAAF{l}{j}}}{\sum_j \left( \Weights{l}{ij} \right)^2 \CVar{\Dataset}{\RCompHidNodeAAF{l}{j}}}.
\end{equation}
Our goal is to understand how $\pdf{\RCompHidNodeBAF{l+1}{i}}{(\Dataset)}{\ArgHidNodeBAF{}{}}$ changes across the set $\left\{ \RCompHidNodeBAF{l+1}{i}  \right\}_{i=1}^{\LayerNumNodes{l+1}}$. Since we are dealing with a Gaussian distribution, it suffices to examine how its mean and variance vary over different nodes $\left\{ \RCompHidNodeBAF{l+1}{i}  \right\}_{i=1}^{\LayerNumNodes{l+1}}$, \emph{i.e.}, using different sets of weights $\left\{ \Weights{l}{i \; \Cdot} \right\}_{i=1}^{\LayerNumNodes{l+1}}$, with $\Weights{l}{i \; \Cdot} \equiv \left\{ \Weights{l}{i j} \right\}_{j=1}^{\LayerNumNodes{l}}$.

Consider the mean $\Einput{\RCompHidNodeBAF{l+1}{i}} = \sum_j \Weights{l}{ij} \Einput{\RCompHidNodeAAF{l}{j}}$. Now, the set $\left\{ \Einput{\RCompHidNodeAAF{l}{j}} \right\}_{j=1}^{\LayerNumNodes{l}}$ acts as a fixed constant set,\footnote{This is the same set of elements $\forall i \in [1, \dots, \LayerNumNodes{l+1}]$.} while the set $\left\{ \Weights{l}{i  \;  \Cdot} \right\}_{j=1}^{\LayerNumNodes{l}}$ changes with the node index $i$. Again, Thm.~\ref{thm:DistrConv_SC} is easily seen to be applicable. We get 
\begin{align}
\begin{split}\label{eq:<h>^l+1}
    &\pdf{\Einput{\RCompHidNodeBAF{l+1}{i}}}{(\Weights{l}{i \Cdot})}{\ArgHidNodeBAF{}{}} = \NormalDens{\ArgHidNodeBAF{}{}}{\sum_j \Eweights{\Weights{l}{ij}} \Einput{\RCompHidNodeAAF{l}{j}}}{\sum_j \CVar{\WeightSet{}}{\Weights{l}{ij}} \Einput{\RCompHidNodeAAF{l}{j}}^2} =
    \NormalDens{\ArgHidNodeBAF{}{}}{0}{\sigma_w^2 \frac{1}{\LayerNumNodes{l}} \sum_{j=1}^{\LayerNumNodes{l}} \Einput{\RCompHidNodeAAF{l}{j}}^2}\,,
\end{split}
\end{align}
where the last step follows from the fact that $\Eweights{\Weights{l}{ij}} = 0$ and $\CVar{\WeightSet{}}{\Weights{l}{ij}} = \frac{\sigma_w^2}{\LayerNumNodes{l}}$. \newline Assuming $ \Einput{\RCompHidNodeAAF{l}{j}}^2$ follows the CLT,\footnote{The \emph{law of large numbers} would suffice here as we are not studying convergence to the mean value.} we have 
\begin{equation}\label{eq:Var_hl_cen_conv}
\frac{1}{\LayerNumNodes{l}} \sum_{j=1}^{\LayerNumNodes{l}}\Einput{\RCompHidNodeAAF{l}{j}}^2 \xrightarrow{\LayerNumNodes{l}\rightarrow\infty} \CE{\Weights{l-1}{j \Cdot}}{\Einput{\RCompHidNodeAAF{l}{j}}^2}\,,
\end{equation}
leading to 
\begin{equation}\label{eq:pdf<Z>^w}
\pdf{\Einput{\RCompHidNodeBAF{l+1}{i}}}{(\Weights{l}{i \Cdot})}{\ArgHidNodeBAF{}{}} \xrightarrow{\LayerNumNodes{l}\rightarrow\infty}  \NormalDens{\ArgHidNodeBAF{}{}}{0}{\sigma_w^2  \CE{\Weights{l-1}{j \Cdot}}{\Einput{\RCompHidNodeAAF{l}{j}}^2}} \,.
\end{equation}

Regarding the variance in Eq.~\eqref{eq:Ph^l_W}, it will also be a \textit{r.v.} dependent on the set $\{ \Weights{l}{ij} \}_{j=1}^{\LayerNumNodes{l}}$. Following a similar approach to the one used for the sum of Gaussian \textit{r.v.}s discussed before,  we can calculate its mean and variance to show that asymptotically $\sum_j \left( \Weights{l}{ij} \right)^2 \CVar{\Dataset}{\RCompHidNodeAAF{l}{j}}  $ converges to a deterministic value equal for all $\LayerNumNodes{l+1}$ nodes.

To make the discussion more concrete we will discuss now, more in detail the specific case of $l=3$
This will be used as an integral part of deriving $\pdf{\RClassFraction{0}}{}{\ClassFraction{0}}$ for deep architectures (see App.~\ref{sec:MLP}).
In App.~\ref{sec:MLP}, we demonstrate that
\begin{align}\label{eq:Ph3_W}
\begin{split}
    &\pdf{\RCompHidNodeBAF{3}{i}}{(\Dataset)}{\ArgHidNodeBAF{}{}} \xrightarrow{\LayerNumNodes{2} \rightarrow \infty} \NormalDens{\ArgHidNodeBAF{}{}}{\Einput{\RCompHidNodeBAF{3}{i}}}{\CVar{\Dataset}{\RCompHidNodeBAF{3}{i}}}
    \equiv \NormalDens{\ArgHidNodeBAF{}{}}{\sum_{j=1}^{\LayerNumNodes{2}} \Weights{2}{ij} \Einput{\RCompHidNodeAAF{2}{j}}}{\sum_{j=1}^{\LayerNumNodes{2}}\left(\Weights{2}{ij}\right)^2 \CVar{\Dataset}{\RCompHidNodeAAF{2}{j}}} \,.
\end{split}
\end{align}
Our aim is to apply the analysis described above to discern the differences between the set of distributions $\left\{\pdf{\RCompHidNodeBAF{3}{i}}{(\Dataset)}{\ArgHidNodeBAF{}{}} \right\}$. Variations in these distributions are induced by differences in the corresponding sets of weight vectors $\{\Weights{2}{i \; \Cdot}\}$. The sets $\left\{\Einput{\RCompHidNodeAAF{2}{j}}\right\}$ and $\left\{\CVar{\Dataset}{\RCompHidNodeAAF{2}{j}}\right\}$, however, remain consistent across each node $\RCompHidNodeBAF{3}{i}$. Therefore, we treat the latter sets as fixed random coefficients, while the variations are attributed to $\{\Weights{2}{i \; \Cdot}\}$. Let us focus on the variance:
\begin{equation}\label{eq:var_h_def}
\CVar{\Dataset}{\RCompHidNodeBAF{3}{i}} = \sum_{j=1}^{\LayerNumNodes{2}}\left(\Weights{2}{ij}\right)^2 \CVar{\Dataset}{\RCompHidNodeAAF{2}{j}}\,.
\end{equation}
To understand how much the random variable $\CVar{\Dataset}{\RCompHidNodeBAF{3}{i}}$ varies, we calculate its mean and variance:
\begin{align}\label{eq:var_Wi_h}
\CE{\Weights{2}{i \; \Cdot}}{\sum_{j=1}^{\LayerNumNodes{2}}\left(\Weights{2}{ij}\right)^2 \CVar{\Dataset}{\RCompHidNodeAAF{2}{j}}} = \sigma_w^2 \frac{1}{\LayerNumNodes{2}} \sum_{j=1}^{\LayerNumNodes{2}} \CVar{\Dataset}{\RCompHidNodeAAF{2}{j}}\, \xrightarrow{\LayerNumNodes{2} \rightarrow \infty} \sigma_w^2 \CE{\Weights{1}{j \; \Cdot}}{\CVar{\Dataset}{\RCompHidNodeAAF{2}{j}}} \,,
\end{align}
with
\begin{align}\label{E_sigma_g2}
\CE{\Weights{1}{j \; \Cdot}}{\CVar{\Dataset}{\RCompHidNodeAAF{2}{j}}}  &=  \int_{\mathbb{R}} \CVar{\Dataset}{\RCompHidNodeAAF{2}{j}} (\Mean{z})  \; \; \pdf{\Einput{\RCompHidNodeBAF{2}{j}}}{(\Weights{l}{i \Cdot})}{\Mean{z}}  d \Mean{z}\\ \notag
&=\int_{\mathbb{R}} \CVar{\Dataset}{\RCompHidNodeAAF{2}{j}} (\Mean{z})  \; \; \NormalDens{\Mean{z}}{0}{\sigma_w^2 \CE{\Weights{0}{j \Cdot}}{\Einput{\RCompHidNodeAAF{1}{}}^2}} d \Mean{z} \,,
\end{align}
where with $\CVar{\Dataset}{\RCompHidNodeAAF{2}{j}} (x) $ we underlined the dependence of $\CVar{\Dataset}{\RCompHidNodeAAF{2}{j}}$ from the \textit{r.v.} $\Einput{\RCompHidNodeBAF{2}{j}} $. In the last step we substituted the p.d.f. expression from Eq.~\eqref{eq:pdf<Z>^w}.
Note that in the integrand of Eq.~\eqref{E_sigma_g2} there is no scaling dependence with respect to $\LayerNumNodes{2}$; this means that
\begin{align}
\CE{\Weights{1}{j \; \Cdot}}{\CVar{\Dataset}{\RCompHidNodeAAF{2}{j}}}  = \mathcal{O} (1) \Longrightarrow \CE{\Weights{2}{i \; \Cdot}}{\CVar{\Dataset}{\RCompHidNodeBAF{3}{i}}} = \mathcal{O} (1).
\end{align}
To evaluate the order of magnitude of the fluctuations, we recall that, given a fixed coefficient $\CVar{\Dataset}{\RCompHidNodeAAF{2}{j}}$,
\begin{align}\label{eq:Var_const}
\CVar{\Weights{2}{i \; \Cdot}}{\left( \Weights{2}{ij} \right)^2  \CVar{\Dataset}{\RCompHidNodeAAF{2}{j}}} = \CVar{\Dataset}{\RCompHidNodeAAF{2}{j}}^2 \CVar{\Weights{2}{i \; \Cdot}}{\left( \Weights{2}{ij} \right)^2 }\,. 
\end{align}
Also, for gaussian variables $\sigma_{x^2}^2 = \mathcal{O} \left( \sigma_{x}^4 \right)$ so $\CVar{\Weights{2}{i \; \Cdot}}{\left( \Weights{2}{ij} \right)^2 } = \mathcal{O} \left( \frac{1}{\LayerNumNodes{2}^2} \right)$. Then from the extensitivity of the variance follows that 
\begin{equation}
\sqrt{\CE{\Weights{2}{i \; \Cdot}}{ \CVar{\Dataset}{\RCompHidNodeBAF{3}{i}}^2} - \CE{\Weights{2}{i \; \Cdot}}{\CVar{\Dataset}{\RCompHidNodeBAF{3}{i}}}^2 } = \mathcal{O} \left( \frac{1}{\sqrt{\LayerNumNodes{2}}} \right)\,.
\end{equation}

By this we conclude that, in the $\LayerNumNodes{2} \rightarrow \infty$ limit, the distribution of the \textit{r.v.} $\sum_{j=1}^{\LayerNumNodes{2}}  \left( \Weights{2}{ij} \right)^2  \CVar{\Dataset}{\RCompHidNodeAAF{2}{j}}$ narrows around the mean value $\sigma_w^2 \CE{\Weights{1}{j \; \Cdot}}{\CVar{\Dataset}{\RCompHidNodeAAF{2}{j}}} $.

We can proceed analogously to evaluate the mean value fluctuations; therefore we have to repeat the same analysis on the mean of $\pdf{\RCompHidNodeBAF{3}{i}}{(\Dataset)}{\ArgHidNodeBAF{}{}}$, \emph{i.e.} (from Eq.~\eqref{eq:Ph3_W}) $\sum_{j=1}^{\LayerNumNodes{2}} \Weights{2}{ij} \Einput{\RCompHidNodeAAF{2}{j}}$. 
In this case, we have 
\begin{align}
\CE{\Weights{2}{i \; \Cdot}}{\sum_{j=1}^{\LayerNumNodes{2}} \Weights{2}{ij} \Einput{\RCompHidNodeAAF{2}{j}}} = \sum_{j=1}^{\LayerNumNodes{2}} \underbrace{\CE{\Weights{2}{i \; \Cdot}}{\Weights{2}{ij}}}_{=0} \Einput{\RCompHidNodeAAF{2}{j}} =0\,.
\end{align}
Proceeding as in \eqref{eq:Var_const} we get
\begin{align}
\sqrt{\CE{\Weights{2}{i \; \Cdot}}{ \Einput{\RCompHidNodeBAF{3}{i}}^2} - \CE{\Weights{2}{i \; \Cdot}}{\Einput{\RCompHidNodeBAF{3}{i}}}^2 } = \mathcal{O} (1)\,.
\end{align}
This means that the center of the distribution $\pdf{\RCompHidNodeBAF{3}{i}}{(\Dataset)}{\ArgHidNodeBAF{}{}}$ keeps fluctuating from node to node even in the limit $\LayerNumNodes{2} \rightarrow \infty$.

More specifically
\begin{align}\label{eq:mu_var}
\left(\CE{\Weights{2}{i \; \Cdot}}{ \Einput{\RCompHidNodeBAF{3}{i}}^2} - \CE{\Weights{2}{i \; \Cdot}}{\Einput{\RCompHidNodeBAF{3}{i}}}^2  \right) = \sigma^2_w \frac{1}{\LayerNumNodes{2}} \sum_{j=1}^{\LayerNumNodes{2}} \Einput{\RCompHidNodeAAF{2}{j}}^2 \xrightarrow{\LayerNumNodes{2} \rightarrow \infty} \sigma_w^2  \CE{\Weights{1}{j \; \Cdot}}{\Einput{\RCompHidNodeAAF{2}{j}}^2}\,.
\end{align}

Also, it is easy to show that the set $\left\{ \Weights{2}{ij} \Einput{\RCompHidNodeAAF{2}{j}} \right\}_{j=1}^{\LayerNumNodes{2}}$ respect the conditions of Thm.~\ref{thm:DistrConv_SC}. Therefore, to summarize,
\begin{equation}
\pdf{\Einput{\RCompHidNodeBAF{3}{i}}}{}{\Mean{z}} \xrightarrow{\LayerNumNodes{2} \rightarrow \infty} \NormalDens{\Mean{z}}{0}{\sigma^2_w \CE{\Weights{1}{j \; \Cdot}}{\Einput{\RCompHidNodeAAF{2}{j}}^2}}\,,
\end{equation}
while 
\begin{equation}
\pdf{\CVar{\Dataset}{\RCompHidNodeBAF{3}{i}}}{}{v_z} \xrightarrow{\LayerNumNodes{2} \rightarrow \infty} \Dirac{ v_z-\sigma_w^2 \CE{\Weights{1}{j \; \Cdot}}{\CVar{\Dataset}{\RCompHidNodeAAF{2}{j}}}}\,.
\end{equation}

\section{Derivation of the distributions}\label{sec:Dist_der}

In this section, our focus is on the \textit{r.v.} $\ROutNode{c}{}$. The primary goal is to prove Thm.~\ref{thm:out_dist} by deriving expressions for $\pdf{\EinputComp{c}}{}{\Mean{}}$ and $\pdf{\ROutNode{c}{}}{(\Dataset)}{\OutNode{}{}}$. These expressions are crucial for determining $\pdf{\RClassFraction{0}}{}{\ClassFraction{0}}$, as explored in App.~\ref{sec:SLP_analysis}.

We begin by demonstrating Thm.~\ref{thm:out_dist} for Multi-Layer Perceptrons (MLPs) with a single hidden layer. Due to the modular structure of MLPs, this demonstration can be extended to deeper network architectures, as shown in App.~\ref{sec:deep_arc}. The sketch proof of Thm.~\ref{thm:out_dist} (found in App.~\ref{app:proof_out_dist}) will be followed by a detailed analysis focused on specific settings, serving as exemplary cases to highlight key qualitative differences. 

For this detailed analysis on concrete examples, we will proceed in steps:
\begin{itemize}
    \item In App.~\ref{sec:E_chi_O}, we will derive an expression for $\pdf{\EinputComp{c}}{}{\Mean{}}$ for various cases of interest.
    \item In App.~\ref{sec:Cond_PO}, we will derive $\pdf{\ROutNode{c}{}}{(\Dataset)}{\OutNode{}{}}
    $ for the same cases.
\end{itemize}

\paragraph{Remark 1.} \label{par:remark_1}
Consider the \textit{r.v.} $\RCompHidNodeBAF{1}{i}$ defined as:
\begin{equation}
\RCompHidNodeBAF{1}{i} = \sum_j \Weights{0}{ij} \InputValue_j.
\end{equation}
The independence of $\RCompHidNodeBAF{1}{i}$ is a direct consequence of the assumed independence of the input components $\InputValue_j$.

For the \textit{r.v.} $\ROutNode{c}{}$, defined by
\begin{equation}
    \ROutNode{c}{} \left( \boldsymbol{\InputValue} ; \WeightSet{} \right) = \sum_{m=1}^{\LayerNumNodes{1}} \Weights{1}{cm} \RCompHidNodeAAF{1}{m},
\end{equation}
we require independence in the set $\{ \RCompHidNodeAAF{1}{m} \}$ to apply the Central Limit Theorem (CLT). Since the activation function is an element-wise operation on $\{ \RCompHidNodeBAF{1}{m} \}$, the independence of $\RCompHidNodeBAF{1}{m}$ suffices.

Jointly normally distributed random variables are independent if and only if their covariance is zero. We calculate the covariance $\textrm{Cov} (\RCompHidNodeBAF{1}{i} , \RCompHidNodeBAF{1}{j} )$ and demonstrate that it converges to 0 \textit{w.h.p.} as the input dimensionality $d \rightarrow \infty$. Specifically, we have:

\begin{align}
\textrm{Cov} \left( \sum_{m=1}^{d} \Weights{1}{i m}  \InputValue_m, \sum_{n=1}^{d} \Weights{1}{j n} \InputValue_n \right)  \stackrel{\boldsymbol{a}}{=} \sum_{m=1}^{d} \sum_{n=1}^{d} \Weights{1}{i m} \Weights{1}{j n} \mathbb{E}(\InputValue_m \InputValue_n) 
\stackrel{\boldsymbol{b}}{=} \sum_{m=1}^{d} \Weights{1}{i m} \Weights{1}{j m} \mathbb{E}(\InputValue_m^2) \stackrel{\boldsymbol{c}}{=} \sum_{m=1}^{d} \Weights{1}{i m} \Weights{1}{j m}.
\end{align}

Here, step $\boldsymbol{a}$ utilizes linearity and the zero mean assumption of $\InputValue_m$. Steps $\boldsymbol{b}$ and $\boldsymbol{c}$ result from substituting the covariance matrix of $\boldsymbol{\InputValue}$, assuming $\boldsymbol{\InputValue} \sim \mathcal{N} (0, \mathbb{I})$. The final summation comprises products of pairs of Gaussian-distributed terms, each with zero mean and variance $\mathcal{O} \left( \frac{1}{{d}} \right)$. As such, it follows from the properties of Gaussian distributions that this summation converges to 0 \textit{w.h.p.} as ${d} \rightarrow \infty$.

\subsection{Proof of Thm. \ref{thm:out_dist}}\label{app:proof_out_dist}
We provide here a sketch proof of Thm.~\ref{thm:out_dist}, detailed as follows:
\begin{theorembox}\label{thm:out_dist_ext}
    Consider a Gaussian distributed dataset with \textit{i.i.d.} components, i.e., $\InputValue_b^{(a)} \sim \mathcal{N}(0,1)$ \footnote{$\InputValue_b^{(a)}$ denotes the $b$-th component of the $a$-th input vector}
    
    processed through an MLP (mapping the input to output according to Eq.~\eqref{eq:h_prop_def}, Eq.~\eqref{eq:g_prop_def}, Eq.~\eqref{eq:rho_prop_def}) with a single hidden layer (\textit{i.e.} $L=1$) and weights initialized according to the Kaiming normal scheme, i.e., $\Weights{l}{ij} \sim \mathcal{N} \left(0,\frac{\sigma_w^2}{\LayerNumNodes{l-1}}\right)$, with zero bias weights. 
Assuming that the variables $\{ \RCompHidNodeAMP{1}{k}{i}  \}_{i=1}^{\ceil{\frac{\LayerNumNodes{1}}{k}}}$ satisfy the conditions of Thm.~\ref{thm:DistrConv_SC}, in the limit of infinite input size and infinite width hidden layer, the distribution of an output node $\OutNode{c}{}$, at initialization, is given by:
\begin{align}
\label{eq:PO_asym_app_thm}
& \pdf{\ROutNode{c}{}}{(\Dataset)}{\OutNode{}{}} \xrightarrow{d,\LayerNumNodes{1} \rightarrow \infty} 
    \NormalDens{\OutNode{}{}}{\EinputComp{c}}{\CVar{\Dataset}{\ROutNode{c}{}}},
\end{align}
    
    Furthermore, while the variance of the distribution, $\CVar{\Dataset}{\ROutNode{c}{}}$, converges w.h.p. to a deterministic value, the center of this distribution, $\EinputComp{c}$, is itself a random variable, varying from node to node, following the distribution:
\begin{align}
\label{eq:P_cen_asym_app_thm}
& \pdf{\EinputComp{c}}{}{\Mean{}} = \NormalDens{\Mean{}}{0}{\CVar{\WeightSet{}}{\EinputComp{c}}}.
\end{align}
We can distinguish two cases:
\begin{itemize}
    \item \textbf{Absence of IGB}: In the absence of IGB, in the limit of infinite datasets, $\CVar{\WeightSet{}}{\EinputComp{c}}$ converges to 0, i.e.,
    \begin{align}\label{eq:NoIGB_Thm_out_cond}
        \lim_{\DatasetSize \rightarrow \infty} \CVar{\WeightSet{}}{\EinputComp{c}} = 0 \Rightarrow \lim_{\DatasetSize \rightarrow \infty} \pdf{\RMean{c}}{}{\Mean{}}  = \delta(\Mean{})
    \end{align}
    \item \textbf{Presence of IGB}: In the presence of IGB, the differences in node centers are not exclusively due to finite size effects, in other words:
    \begin{align}
        \lim_{\DatasetSize \rightarrow \infty} \CVar{\WeightSet{}}{\RMean{c}} \neq 0
    \end{align}
\end{itemize}
\end{theorembox}

\begin{proof}

Fixing the set $\WeightSet{}$, $\RCompHidNodeBAF{1}{i}$ is a linear combination of \textit{r.v.}s which meet the conditions of Thm.~\ref{thm:DistrConv_SC}. Thus, in the limit as $d \rightarrow \infty$, we have
\begin{align}\label{eq:h1_distr}
\pdf{\RCompHidNodeBAF{1}{i}}{(\Dataset)}{\ArgHidNodeBAF{}{}} \xrightarrow{d\rightarrow\infty} \NormalDens{\ArgHidNodeBAF{}{}}{\sum_j \Weights{0}{ij} \Einput{\InputValue_j}}{ \sum_j \left( \Weights{0}{ij} \right)^2 \sigma^2_{\InputValue_j}}  = \NormalDens{\ArgHidNodeBAF{}{}}{0}{\sum_j \left( \Weights{0}{ij} \right)^2},
\end{align}
where we have used the definition of $\boldsymbol{\InputValue}$ for our analysis to substitute $\Einput{\InputValue_j}=0$ and $\sigma^2_{\InputValue_j} = 1 \; \forall j$.
As shown in Sec.~\ref{sec:RescGaussVar}, we find
\begin{align}\label{eq:P_h1}
\sum_{j=1}^d \left( \Weights{0}{ij} \right)^2 = d \; \Eweights{\left( \Weights{0}{ij} \right)^2} \xrightarrow{d\rightarrow\infty} \sigma^2_w \Longrightarrow \pdf{\RCompHidNodeBAF{1}{i}}{(\Dataset)}{\ArgHidNodeBAF{}{}} \xrightarrow{d\rightarrow\infty}
\NormalDens{\ArgHidNodeBAF{}{}}{0}{\sigma^2_w} .
\end{align}
Now, passing through the activation function, and possibly through the pooling layer, we have a set of \textit{i.i.d.} variables, $\{ \RCompHidNodeAMP{1}{k}{l}  \}_{l=1}^{\ceil{\frac{\LayerNumNodes{1}}{k}}}$. Furthermore, by hypothesis, the set $\{ \RCompHidNodeAMP{1}{k}{l}  \}_{l=1}^{\ceil{\frac{\LayerNumNodes{1}}{k}}}$ satisfies the conditions of Thm.~\ref{thm:DistrConv_SC}. Therefore, the distribution of the \textit{r.v.}
\begin{equation}\label{eq:Oc_def}
\ROutNode{c}{} \left( \boldsymbol{\InputValue} ; \WeightSet{} \right) = \sum_{l=1}^{\LayerNumNodes{1}} \Weights{1}{cl} \RCompHidNodeAMP{1}{k}{l}.
\end{equation}
will converge, in the limit of $\LayerNumNodes{1} \rightarrow \infty$, to a Gaussian distribution:
\begin{align}
   \label{eq:Oc_conv_1L_proof}
   \lim_{\LayerNumNodes{1} \rightarrow \infty}  \pdf{\ROutNode{c}{}}{(\Dataset)}{\OutNode{}{}} =
    \NormalDens{\OutNode{}{}}{\EinputComp{c}}{\CVar{\Dataset}{\ROutNode{c}{}}}.
\end{align}

In particular, since the variables $\{ \RCompHidNodeAMP{1}{k}{l}  \}_{l=1}^{\ceil{\frac{\LayerNumNodes{1}}{k}}}$ are \textit{i.i.d.} and the weights $\{ \Weights{1}{cl} \}_{l=1}^{\ceil{\frac{\LayerNumNodes{1}}{k}}}$ are fixed constants from initialization, defining  
\begin{align}
   &\lim_{\DatasetSize \to  \infty} \Einput{\RCompHidNodeAMP{1}{k}{l}}= \Einput{\RCompHidNodeAMP{1}{k}{}}  \; \; \; \forall l,  \\
   &\lim_{\DatasetSize \to  \infty} \CVar{\Dataset}{\RCompHidNodeAMP{1}{k}{l}} = \CVar{\Dataset}{\RCompHidNodeAMP{1}{k}{}}  \; \; \; \forall l.
\end{align}

from Eq.~\eqref{eq:Oc_def} it follows: 
\begin{align}
\EinputComp{c} &= \Einput{\RCompHidNodeAMP{1}{k}{}} \sum_{l=1}^{\LayerNumNodes{1}} \Weights{1}{cl} \equiv \Einput{\RCompHidNodeAMP{1}{k}{}} S_{w_c}^{(\LayerNumNodes{1})} \label{eq:muc_1L_proof}\\
\CVar{\Dataset}{\ROutNode{c}{}} &= \CVar{\Dataset}{\RCompHidNodeAMP{1}{k}{}} \sum_{l=1}^{\LayerNumNodes{1}} \left( \Weights{1}{cl} \right)^2 \equiv \CVar{\Dataset}{\RCompHidNodeAMP{1}{k}{}} S_{w_c^2}^{(\LayerNumNodes{1})}.
\end{align}

From the analysis in App.~\ref{sec:RescGaussVar}, we know that while $S_{w_c}^{(\LayerNumNodes{1})}$ is a \textit{r.v.} with a non-degenerate distribution for all $\LayerNumNodes{1}$, the distribution of $S_{w_c^2}^{(\LayerNumNodes{1})}$ concentrates around its mean value in the limit $\LayerNumNodes{1} \rightarrow \infty$. In other words,
\begin{align}\label{eq:VarO_rho}
    \lim_{\LayerNumNodes{1} \rightarrow \infty} \CVar{\Dataset}{\ROutNode{c}{}} = \sigma_w^2 \CVar{\Dataset}{\RCompHidNodeAMP{1}{k}{}} \; \; \forall c.
\end{align}
We can thus distinguish two cases:
\begin{itemize}
    \item $\Einput{\RCompHidNodeAMP{1}{k}{}} = 0$: \newline In this case, the output nodes will be identically distributed, i.e., there will be an absence of IGB. Specifically, from Eq.s~\eqref{eq:Oc_conv_1L_proof} and ~\eqref{eq:muc_1L_proof} and , it follows:
    \begin{align}\label{eq:VarO_asymp_thm_out_dist}
        \lim_{\LayerNumNodes{1} \rightarrow \infty}  \pdf{\ROutNode{c}{}}{(\Dataset)}{\OutNode{}{}} =
    \NormalDens{\OutNode{}{}}{0}{\sigma_w^2 \CVar{\Dataset}{\RCompHidNodeAMP{1}{k}{}}}.
    \end{align}

    \item $\Einput{\RCompHidNodeAMP{1}{k}{}} \neq 0$: \newline In this case, the output nodes will be centered at different points, leading to the emergence of IGB. Specifically, from Eq.s~\eqref{eq:Oc_conv_1L_proof} and ~\eqref{eq:muc_1L_proof} , it follows:
    \begin{align} \label{eq:P_O_IGB}
        \lim_{\LayerNumNodes{1} \rightarrow \infty}  \pdf{\ROutNode{c}{}}{(\Dataset)}{\OutNode{}{}} =
    \NormalDens{\OutNode{}{}}{\Einput{\RCompHidNodeAMP{1}{k}{}} S_{w_c}^{(\LayerNumNodes{1})}}{\sigma_w^2 \CVar{\Dataset}{\RCompHidNodeAMP{1}{k}{}}}.
    \end{align}
   Since $S_{w_c}^{(\LayerNumNodes{1})}$ is a sum of $\LayerNumNodes{1}$ \textit{i.i.d.} Gaussians, $\Weights{1}{cl} \sim \NormalDistr{0}{\frac{\sigma_w^2}{\LayerNumNodes{1}}}$, it follows
   \begin{align}
       S_{w_c}^{(\LayerNumNodes{1})} \sim \NormalDistr{0}{\sigma_w^2}.
   \end{align}
   Consequently,
   \begin{align}\label{eq:P<O>_teo_IGB_cond_ext}
       \pdf{\EinputComp{c}}{}{\Mean{}} = \NormalDens{\Mean{}}{0}{\sigma_w^2 \Einput{\RCompHidNodeAMP{1}{m}{}}^2 }.
   \end{align}
\end{itemize}

\end{proof}

\subsection{Derivation of \texorpdfstring{$\boldsymbol{\pdf{\RMean{c}}{}{\Mean{}}}$}{pmu(x)} for some emblematic cases}\label{sec:E_chi_O}
Thm.~\ref{thm:out_dist_ext} demonstrates that the decisive element for deriving $\pdf{\RMean{c}}{}{\Mean{}}$ is $\Einput{\RCompHidNodeAMP{1}{k}{}}$. To make the discussion more concrete, we will consider some cases to illustrate their distinguishing features. In particular, we will use the same settings discussed at the beginning of Sec.~\ref{sec:QuantAn}. 

\paragraph{Linear}\label{sec:E_chi_LinActFun}
Consider the case where
\begin{align}\label{eq:lin_case_cond}
\RCompHidNodeAMP{1}{1}{i} = \MPOperator{1;1}{ \AFOperator{\RCompHidNodeBAF{1}{i}}} = \RCompHidNodeBAF{1}{i}.
\end{align}
From Eq.s~\eqref{eq:h1_distr} and ~\eqref{eq:lin_case_cond}, it follows that $\Einput{\RCompHidNodeAMP{1}{1}{}} = 0$. Therefore, in this setting, we have no presence of IGB, and according to Eq.~\ref{eq:NoIGB_Thm_out_cond}, we have:
\begin{align}\label{eq:lin_P<O>}
    \lim_{\DatasetSize \rightarrow \infty} \pdf{\RMean{c}}{}{\Mean{}} = \delta(\Mean{}).
\end{align}

Note that \eqref{eq:lin_P<O>} is formally true only in the limit of an infinite size dataset. For a finite size dataset, we must compute the average $\Einput{\RCompHidNodeAMP{1}{1}{}}$ using the empirical distribution, $\pdf{\RCompHidNodeBAF{1}{k}}{(\Dataset; E)}{\ArgHidNodeBAF{}{}} \equiv \sum_{a=1}^{\DatasetSize} \frac{1}{\DatasetSize} \delta \left( \ArgHidNodeBAF{}{} - \hat{h}_k^{(1)} \left( \boldsymbol{\InputValue}^{(a)} \right) \right)$, where the set $  \left\{ \hat{h}_k^{(1)} \left( \boldsymbol{\InputValue}^{(a)} \right) \right\}_{a=1}^{\DatasetSize}$ contains the values mapped from each element of the dataset to the $k^{\text{th}}$ node of the first hidden layer. We then have
\begin{align}\label{eq:FS_DatasetAverage}
\begin{split}
    &\Einput{\RCompHidNodeAMP{1}{1}{j} } = \int_{\mathbb{R}}  \ArgHidNodeBAF{}{} \,  \pdf{\RCompHidNodeBAF{1}{k}}{(\Dataset; E)}{\ArgHidNodeBAF{}{}} d \ArgHidNodeBAF{}{}
 =
  \frac{1}{\DatasetSize} \sum_{a=1}^{\DatasetSize}   \hat{h}_j^{(1)} \left( \boldsymbol{\InputValue}^{(a)} \right) \underset{\boldsymbol{A}}{\simeq}
  \frac{1}{\sqrt{\DatasetSize}} Z_j = \tilde Z_{j},
 \end{split}
\end{align}
where  $Z_j \sim \mathcal{N}(0, \sigma_{w}^2)$, and $\tilde Z_{j} \sim \mathcal{N}\left( 0, \frac{\sigma_{w}^2}{\DatasetSize} \right)$.

Therefore, considering finite size effects, $\Einput{\RCompHidNodeAMP{1}{1}{} } \neq \Einput{\RCompHidNodeAMP{1}{1}{j} }$. In this case, instead of starting from Eq.~\eqref{eq:muc_1L_proof}, we can proceed similarly to the derivation of Eq.~\eqref{eq:<h>^l+1}, obtaining
\begin{align}\label{eq:Linear_FSE_<O>}
    \pdf{\RMean{c}}{}{\Mean{}} = \NormalDens{\Mean{}}{0}{\sigma_w^2 \Eweights{\tilde Z_{j}^2} } = \NormalDens{x}{0}{ \frac{4}{ D} }.
\end{align}
Note that in step $\boldsymbol{A}$ of Eq.~\eqref{eq:FS_DatasetAverage}, we again used the CLT for the sum of \textit{r.v.}s. The key difference with the previous derivation is that we do not use the CLT for $\pdf{\RCompHidNodeBAF{1}{k}}{(\Dataset)}{\ArgHidNodeBAF{}{}}$; instead, we substitute the empirical distribution, $\pdf{\RCompHidNodeBAF{1}{k}}{(\Dataset; E)}{\ArgHidNodeBAF{}{}}$. This latter distribution, defined on the dataset elements, accounts for the finite size effects of the dataset itself (see dependence from $\DatasetSize$). Note that the distribution of the variables involved in the sum in the last part of Eq.~\eqref{eq:FS_DatasetAverage} ($\tilde Z_{j}$) narrows as $\DatasetSize \rightarrow \infty$; in this limit, the finite size effects disappear, and we converge to the result in Eq.~\eqref{eq:lin_P<O>}.
\paragraph{ReLU}\label{sec:relu_analysis}
We now repeat the computation using the ReLU activation function (introduced in \citet{hahnloser2000digital}), \textit{i.e.},
\begin{align}
  \RCompHidNodeAMP{1}{1}{i} = \MPOperator{1;1}{\RCompHidNodeAAF{1}{j}} =   \RCompHidNodeAAF{1}{j} = \AFOperator{ \RCompHidNodeBAF{1}{j} }\equiv \max \{0, \RCompHidNodeBAF{1}{j} \}.
\end{align}

$\RCompHidNodeAAF{1}{j}$ will follow the same distribution as $\RCompHidNodeBAF{1}{j}$ on the positive support, since $\RCompHidNodeAAF{1}{j} = \RCompHidNodeBAF{1}{j}$ for $\RCompHidNodeBAF{1}{j} >0$. The probability density on the negative support of $ \pdf{\RCompHidNodeBAF{1}{j}}{(\Dataset)}{\ArgHidNodeBAF{}{}} $ will collapse to $0$ since $\RCompHidNodeAAF{1}{j} = 0$ for $\RCompHidNodeBAF{1}{j} <0$. Substituting $ \pdf{\RCompHidNodeBAF{1}{j}}{(\Dataset)}{\ArgHidNodeBAF{}{}} $ with the Gaussian distribution from Eq.~\eqref{eq:P_h1}, we get:
\begin{align}\label{Pg_Relu}
\begin{split}
& \pdf{\RCompHidNodeAAF{1}{j}}{(\Dataset)}{\ArgHidNodeAAF{}{}} = \left( \int_{\mathbb{R}^{-}}  \pdf{\RCompHidNodeBAF{1}{j}}{(\Dataset)}{\ArgHidNodeBAF{}{}}  d \ArgHidNodeBAF{}{} \right) \Dirac{\ArgHidNodeAAF{}{}} + \Heavyside{\ArgHidNodeAAF{}{}}  \pdf{\RCompHidNodeBAF{1}{j}}{(\Dataset)}{\ArgHidNodeBAF{}{}}  = \frac{1}{2} \Dirac{\ArgHidNodeBAF{}{}} + \Heavyside{\ArgHidNodeBAF{}{}} \NormalDens{\ArgHidNodeBAF{}{}}{0}{\sigma_{w}^2},
\end{split}
\end{align}
where $\Dirac{x}$ represents the Dirac delta distribution, and $\Heavyside{x}$ represents the Heaviside step function.

Integrating over this distribution, we proceed as in the previous case:
\begin{align}\label{eq:input_average_relu}
\begin{split}
& \Einput{\RCompHidNodeAMP{1}{1}{} }= \Einput{\RCompHidNodeAMP{1}{1}{j} } = \int_{\mathbb{R}}  \ArgHidNodeAAF{}{} \, \pdf{\RCompHidNodeAAF{1}{j}}{(\Dataset)}{\ArgHidNodeAAF{}{}} d \ArgHidNodeAAF{}{} =  \int_{\mathbb{R}} \ArgHidNodeAAF{}{} \left( \frac{1}{2} \delta(\ArgHidNodeAAF{}{}) + \NormalDens{\ArgHidNodeAAF{}{}}{0}{\sigma_{w}^2} \right) d\ArgHidNodeAAF{}{} =\\
&   \int_{0}^{+ \infty} \ArgHidNodeAAF{}{} \NormalDens{\ArgHidNodeAAF{}{}}{0}{\sigma_{w}^2} d\ArgHidNodeAAF{}{} =  \sigma_{w}^2 \int_{0}^{+ \infty} \NormalDens{\ArgHidNodeAAF{}{}}{0}{\sigma_{w}^2} d\frac{\ArgHidNodeAAF{}{}^2}{2 \sigma_{w}^2} = \frac{\sigma_{w}}{\sqrt{2 \pi }}.
\end{split}
\end{align}

The first notable difference in the case of ReLU activation is that $\Einput{\RCompHidNodeAMP{1}{1}{} } \neq 0$, thus indicating the emergence of IGB. Specifically, from Eq.~\eqref{eq:P<O>_teo_IGB_cond_ext}, it follows:
\begin{align}\label{eq:ReLU_<O>}
    \pdf{\RMean{c}}{}{\Mean{}} = \NormalDens{\Mean{}}{0}{\sigma_w^2 \frac{\sigma_w^2}{2 \pi} } = \NormalDens{\Mean{}}{0}{ \frac{2}{ \pi} },
\end{align}
where in the last step we have substituted $\sigma_w$ with the typical gain value used in the presence of ReLU, \textit{i.e.}, $\sigma_w = \sqrt{2}$ \citep{he2015delving}.

In this scenario, unlike the linear activation function case discussed at the end of App.~\ref{sec:E_chi_LinActFun}, we do not need to consider the finite size effects of the dataset. While for a linear activation function $\lim_{\DatasetSize \rightarrow \infty} \Einput{\RCompHidNodeAMP{1}{1}{}} = 0$, for ReLU $\lim_{\DatasetSize \rightarrow \infty} \Einput{\RCompHidNodeAMP{1}{1}{}} \neq 0$ (as shown in \eqref{eq:input_average_relu}). Therefore, the finite size corrections of $\mathcal{O} \left( \frac{1}{\DatasetSize } \right)$ are negligible when $\DatasetSize \Einput{\RCompHidNodeAMP{1}{1}{}} \gg 1$.

\begin{figure}
    \centering
    \includegraphics[width=.7\columnwidth]{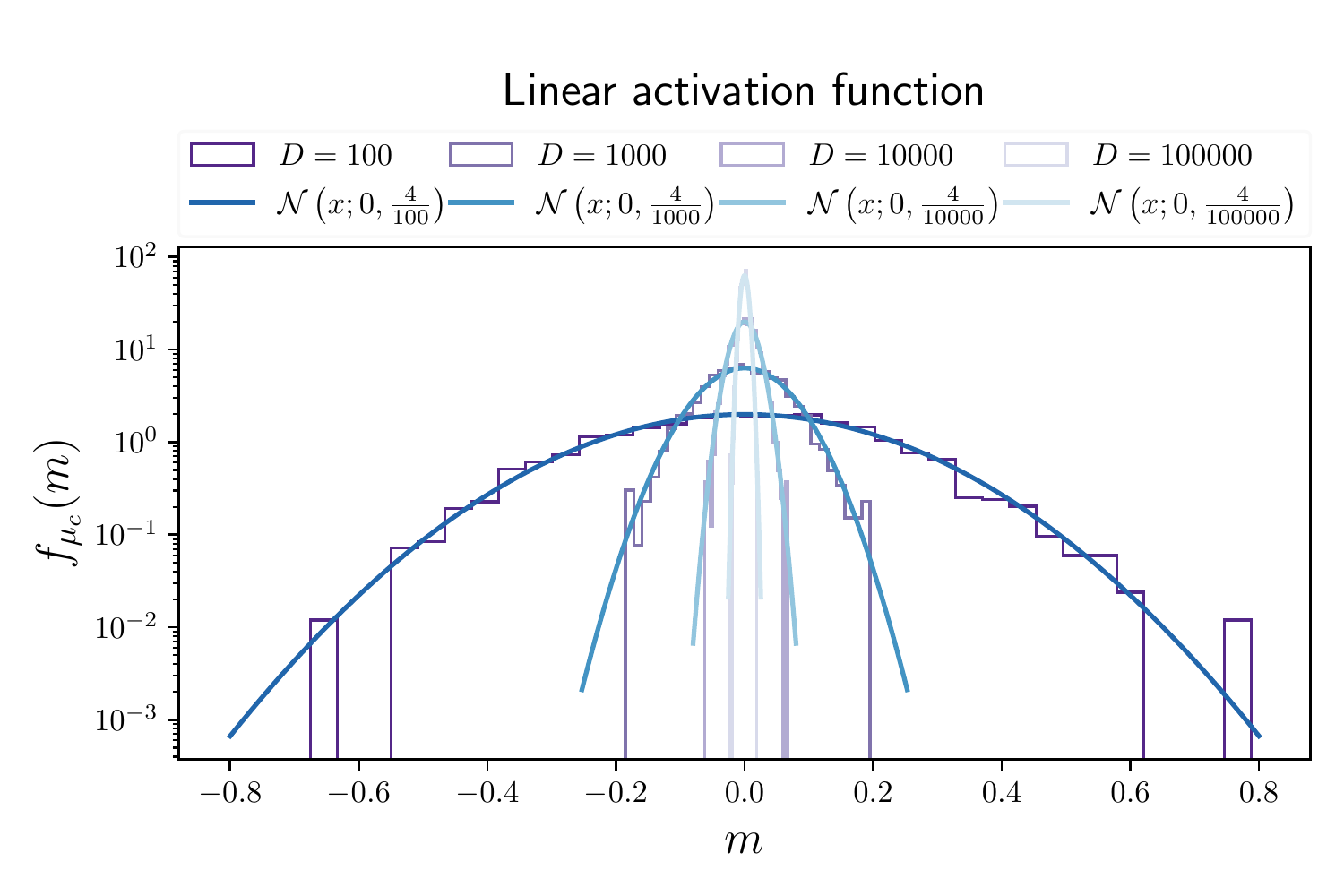}
    \includegraphics[width=.7\columnwidth]{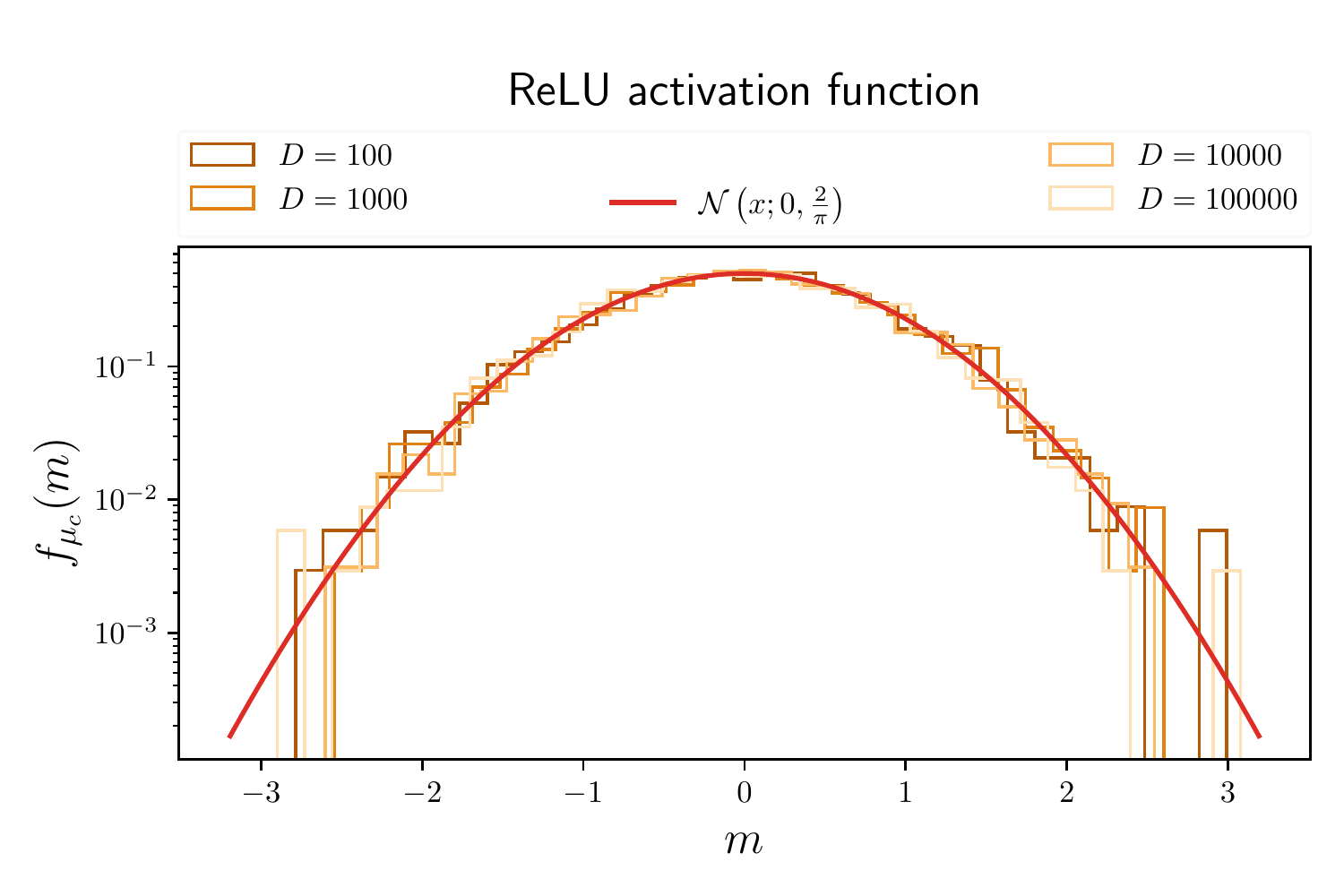}
    \caption{Comparison of the distribution $\pdf{\EinputComp{c}}{}{x}$ between the linear activation function (upper plot) and ReLU (bottom plot) across different dataset sizes ($\DatasetSize$). Both plots also include the theoretical distributions derived in App.~\ref{sec:E_chi_O}. Observe the distinct behaviors in the two cases: for the linear activation function, the distribution narrows as $\DatasetSize$ increases, while for the ReLU case, it remains stable. This stability in the ReLU case aligns with expectations from Eq.~\eqref{eq:input_average_relu} and Eq.~\eqref{eq:FS_DatasetAverage}. These simulations employed the \GB~ dataset and the \MLPA~ model. For more details, refer to App.~\ref{sec:reprod}.
}
\label{D_Comparison}
\end{figure}
In Fig.~\ref{D_Comparison}, the distributions $\pdf{\RMean{c}}{}{\Mean{}}$ along with their respective theoretical curves, \textit{i.e.}, Eq.~\eqref{eq:Linear_FSE_<O>} and Eq.~\eqref{eq:ReLU_<O>}, are compared for the two types of activation functions. Consistent with Eq.~\eqref{eq:Linear_FSE_<O>} and Eq.~\eqref{eq:ReLU_<O>}, we observe that in the linear case, the distribution narrows around $0$ as $\DatasetSize$ increases, whereas in the ReLU case, it remains stable.

\paragraph{ReLU + max pool} \label{sec:MaxPool_effect}
With max pooling, we group the set $\{ \RCompHidNodeAAF{1}{k} \}$ into subgroups and select the maximum value from each, thereby reducing the dimensionality of the layer. Incorporating this into the analysis in App.~\ref{sec:relu_analysis}, we have:
\begin{align}
\RCompHidNodeAMP{1}{k}{l} = \MPOperator{k}{\left\{  \AFOperator{\RCompHidNodeBAF{1}{j}} \right\}_{j \in S_l^k}} =\max_{j \in S_l^k} \{ \max \{0, \RCompHidNodeAAF{1}{j}  \} \},
\end{align}
where $S_l^k$ indicates the $l$-th subgroup of $k$ nodes.

To proceed with the computation from the previous section, we need to derive $\pdf{\RCompHidNodeAMP{1}{k}{l}}{(\Dataset)}{\ArgHidNodeAMP{}{} }$. Generally, if $Y \equiv \max \{ X_1, \dots , X_m \}$ with $\{X_i \}_{i=1}^m$ being i.i.d., we have:
\begin{align}\label{cdf_max}
F_Y(y) \equiv \Prob{}{Y\leq y} = \Prob{}{\max \{ X_1, \dots , X_m \} \leq y} = \prod_{i=1}^m \Prob{}{X_i \leq y} \equiv \prod_{i=1}^m F_X(y) = F_X(y)^m.
\end{align}
Differentiating, we find:
\begin{equation}\label{P_Y_relu}
\pdf{Y}{}{y }=m \, \pdf{X}{}{ y } F_X(y)^{m-1}.
\end{equation}
For our case, we substitute $\pdf{X}{}{x}$ with the distribution used in App.~\ref{sec:relu_analysis}, \textit{i.e.},
\begin{equation}\label{PX_relu}
\pdf{X}{}{x}\equiv \pdf{\RCompHidNodeAAF{1}{j}}{(\Dataset)}{x} =   \frac{1}{2} \Dirac{x} + \Heavyside{x} \NormalDens{x}{0}{\sigma_{w}^2}.
\end{equation}

From Eq.~\eqref{PX_relu}, it follows:
\begin{align}
\begin{split}
& F_X(y) \equiv \Prob{}{\RCompHidNodeAAF{1}{j} \leq y \mid  \WeightSet{} } = \int_{-\infty}^y \frac{1}{2} \Dirac{x} + \Heavyside{x} \NormalDens{x}{0}{\sigma_{w}^2} dx = \frac{1}{2} + \frac{1}{2}\erf\left( \frac{y}{\sqrt{2 \sigma_{w}^2}}\right)\Theta(y).
\end{split}
\end{align}
Applying the Fisher–Tippett–Gnedenko theorem, we can find an asymptotic result for $\pdf{ \RCompHidNodeAMP{1}{k}{l} }{(\Dataset)}{\ArgHidNodeAMP{}{}}$ as $k \rightarrow \infty$. For general $k$ values, combining all components yields:
\begin{align}\label{dataset_average_maxPool}
\begin{split}
    &\Einput{\RCompHidNodeAMP{1}{k}{}}=\Einput{\RCompHidNodeAMP{1}{k}{l}} = \int_{\mathbb{R}}  \ArgHidNodeAMP{}{} \, \pdf{ \RCompHidNodeAMP{1}{k}{l} }{(\Dataset)}{\ArgHidNodeAMP{}{}} d \ArgHidNodeAMP{}{} =
  \int_0^{\infty} m \, \ArgHidNodeAMP{}{} \, \NormalDens{\ArgHidNodeAMP{}{}}{0}{\sigma_{w}^2} \left( \frac{1}{2} + \frac{1}{2}\erf\left( \frac{\ArgHidNodeAMP{}{}}{\sqrt{2 \sigma_{w}^2}} \right) \right)^{m-1} d\ArgHidNodeAMP{}{}.
\end{split}
\end{align}
This can be approximated by an asymptotic expansion or evaluated numerically. In the case of linear activation with max pooling, a simple Gaussian distribution is substituted for $\pdf{X}{}{x}$.

Defining a notation to simplify expressions:
\begin{equation}
c_k(\sigma_w) \equiv \int_0^{\infty} k \, \ArgHidNodeAMP{}{} \, \NormalDens{\ArgHidNodeAMP{}{}}{0}{\sigma_{w}^2} \left( \frac{1}{2} + \frac{1}{2}\erf\left( \frac{x}{\sqrt{2 \sigma_w^2}} \right) \right)^{k-1} d\ArgHidNodeAMP{}{}.
\end{equation}

In this case too, IGB emerges; specifically, from Eq.~\eqref{eq:P<O>_teo_IGB_cond_ext}, it follows:
\begin{align}\label{eq:MP_<O>}
   \pdf{\EinputComp{c}}{}{\Mean{}} = \NormalDens{\Mean{}}{0}{\sigma_w^2 c_k(\sigma_w)^2 }.
\end{align}

\subsection{Derivation of \texorpdfstring{\scalebox{\BP}{$\boldsymbol{\pdf{\ROutNode{c}{}}{(\Dataset)}{\OutNode{}{}} }$}}{P(O(x)|W)} for some emblematic cases}\label{sec:Cond_PO}
As discussed, $\EinputComp{c}$ is a \textit{r.v.} that varies with the set of weights $\WeightSet{}$. In App.~\ref{sec:E_chi_O}, we derived its distribution. In this section, we focus on the distribution of the output, given a specific configuration of the weight set, \scalebox{\BP}{$\pdf{\ROutNode{c}{}}{(\Dataset)}{ \OutNode{}{}} $}, for the same cases analyzed in App.~\ref{sec:E_chi_O}.

\begin{itemize}
\item \textbf{Linear}: Starting from Eq.~\eqref{eq:lin_case_cond} and combining Eq.\eqref{eq:P_h1} with Eq.~\ref{eq:VarO_asymp_thm_out_dist}, we get
\begin{equation}
\lim_{\LayerNumNodes{1} \rightarrow \infty} \pdf{\ROutNode{c}{}}{(\Dataset)}{\OutNode{}{}} =
    \NormalDens{\OutNode{}{}}{0}{\sigma_w^4 }.
\end{equation}

\item \textbf{ReLU}: 
From Eq.~\eqref{Pg_Relu}, we have
\begin{align}
\Einput{\RCompHidNodeAAF{1}{i}} &= \frac{\sigma_w}{\sqrt{2 \pi}},\\
\begin{split}
\CVar{\Dataset}{\RCompHidNodeAAF{1}{i}} &= \int_{\mathbb{R}^+} \ArgHidNodeAAF{}{}^2 \pdf{\RCompHidNodeAAF{1}{i}}{(\Dataset)}{ \ArgHidNodeAAF{}{}} d  \ArgHidNodeAAF{}{} - \Einput{\RCompHidNodeAAF{1}{i}}^2=\frac{\sigma_w^2 (\pi -1)}{2 \pi} \label{eq:Var_ReLU_1L},
\end{split}
\end{align}
where the symmetry of $\pdf{\RCompHidNodeBAF{1}{k}}{(\Dataset)}{\ArgHidNodeBAF{}{}}$ was used. Substituting into Eq.~\eqref{eq:P_O_IGB} and using Eq.~\eqref{eq:Var_ReLU_1L}, we obtain
\begin{align}\label{Cond_Distr_ReLU}
\begin{split}
    &\lim_{\LayerNumNodes{1} \rightarrow \infty} \pdf{\ROutNode{c}{}}{(\Dataset)}{\OutNode{}{}} = \NormalDens{\OutNode{}{}}{\EinputComp{c} }{\sigma_w^2 \frac{\sigma_w^2 ( \pi -1)}{2 \pi}} \, ,
\end{split}
\end{align}
where $\EinputComp{c}$ is a \textit{r.v.} distributed according to Eq.~\eqref{eq:ReLU_<O>}.

\item \textbf{ReLU + max pool}:
This case is conceptually similar to the ReLU case; the main difference is that we do not have an analytical expression for $\Einput{\RCompHidNodeAMP{1}{k}{i}}$ and $\CVar{\Dataset}{\RCompHidNodeAMP{1}{k}{i}}$. We can numerically compute these cumulants for a fixed $k$, with their expressions given by
\begin{align}
\begin{split}
& \Einput{\RCompHidNodeAMP{1}{k}{}} = \int_0^{\infty} k \ArgHidNodeAMP{}{} \NormalDens{\ArgHidNodeAMP{}{}}{0}{\sigma_{w}^2} \left( \frac{1}{2} + \frac{1}{2}\erf\left( \frac{\ArgHidNodeAMP{}{}}{\sqrt{2 \sigma_w^2}} \right) \right)^{k-1} d\ArgHidNodeAMP{}{},
\end{split}
\end{align}
\begin{align}
\begin{split}
& \CVar{\Dataset}{\RCompHidNodeAMP{1}{k}{}} = \left( \int_0^{\infty} k \ArgHidNodeAMP{}{}^2 \NormalDens{\ArgHidNodeAMP{}{}}{0}{\sigma_{w}^2} \left( \frac{1}{2} + \frac{1}{2}\erf\left( \frac{\ArgHidNodeAMP{}{}}{\sqrt{2 \sigma_w^2}} \right) \right)^{k-1} dx \right) - \Einput{\RCompHidNodeAMP{1}{k}{}}^2.
\end{split}
\end{align}
Thus, analogous to Eq.~\eqref{Cond_Distr_ReLU}, we obtain
\begin{align}
\begin{split}
    & \lim_{\LayerNumNodes{1} \rightarrow \infty}\pdf{\ROutNode{c}{}}{(\Dataset)}{\OutNode{}{}} = \NormalDens{\OutNode{}{}}{\EinputComp{c}}{\sigma_w^2 \CVar{\Dataset}{\RCompHidNodeAMP{1}{k}{}}},
\end{split}
\end{align}
where $\EinputComp{c}$ is a \textit{r.v.} distributed according to Eq.~\eqref{eq:MP_<O>}.
\end{itemize}

\subsection{Non identically distributed classes}\label{App:non_id_classes}
For our analysis we consider unstructured data identically distributed among classes. This choice places us in a symmetric setting where the presence of a predictive bias cannot be linked to differences between classes. This setting allows us to isolate the effect of the architecture sidestepping potential additional effects coming from dataset attributes. However, extending the analysis to other settings is possible; to illustrate this we now show that the analysis can be extended to include classes that are not identically distributed. Specifically, we will extend the current setting by introducing class-specific differences in the variance of the Gaussian blobs.
We begin by demonstrating that, given an initialization, both classes exhibit a predictive bias towards the same class. Let's define a generic value for the class $c$ input variance, $\sigma_{\xi^{(c)}}^2$. We can then repeat the computation for each class subgroup as done in App.~\ref{app:proof_out_dist}, specifically for an MLP with ReLU activation. We divide the dataset into subgroups based on class membership (each now following different statistics) and repeat the computation for each. In particular, consider Eq.~\eqref{eq:muc_1L_proof} and assume (without loss of generality) that for a given class $c$, we have a prediction bias towards class 0, i.e.

\begin{equation} 
\mu^{(c)}_0- \mu^{(c)}_1 = \frac{\sigma_w \sigma_{\xi^{(c)}}}{\sqrt{2 \pi}} (S_{w_0}-S_{w_1}) >0.
\end{equation}

This expression comes from substituting Eq.~\eqref{eq:input_average_relu} with the class-dependent variance. Then, for a generic class $c'$, we will have:

$$\mu^{(c')}_0- \mu^{(c')}_1 = \frac{\sigma_w \sigma_{\xi^{(c')}}}{\sqrt{2 \pi}} (S_{w_0}-S_{w_1}) = \frac{\sigma_{\xi^{(c')}}}{\sigma_{\xi^{(c)}}} \mu^{(c)}_0- \mu^{(c)}_1>0$$

indicating that the predictions for elements of class $c'$ will also be biased towards class 0.

Thus, for any given initialization, the bias is directed towards the same class for each subgroup. Furthermore, we can show that each subgroup is characterized by the same level of bias, irrespective of $\sigma_{\xi^{(c)}}^2$.The ratio between the two variances, $\textrm{Var}_{\mathcal{W}}(\mu_{c',c})$ and $\textrm{Var}_{\chi}(O^{(c',c)})$, quantifies the level of IGB for the subgroup of datapoints belonging to class $c$. Repeating the computation from App.~\ref{app:proof_out_dist} (particularly from Eq.~\eqref{eq:h1_distr}, Eq.~\eqref{eq:P<O>_teo_IGB_cond_ext}, and Eq.~\eqref{eq:input_average_relu}), we get:

$$ \textrm{Var}_{\mathcal{W}}(\mu_{c',c}) = \frac{\sigma_w^4 \sigma_{\xi^{(c)}}^2}{2 \pi}$$

Similarly, for the variance of the output nodes (from Eq.~\eqref{eq:P_O_IGB}, Eq.~\eqref{eq:Var_ReLU_1L}, and Eq.~\eqref{Cond_Distr_ReLU}), we get:

$$ \textrm{Var}_{\chi}(O^{(c',c)}) = \frac{\sigma_w^4 \sigma_{\xi^{(c)}}^2 (\pi -1)}{2 \pi}$$

The expressions above confirm that the ratio between the two variances is independent of $\sigma_{\xi^{(c)}}^2$, for each subgroup $c$.

\section{Single hidden layer perceptron}\label{sec:1lp}
In this section, we utilize the results derived in App.~\ref{sec:Dist_der} to deduce our primary goal, $\pdf{\RClassFraction{0}}{}{\ClassFraction{0}}$. We specifically focus on a single hidden layer perceptron; in App.~\ref{sec:deep_arc} we will extend the results to deep architectures.

\subsection{Derivation of \texorpdfstring{$\boldsymbol{\pdf{\RClassFraction{0}}{}{\ClassFraction{0}}}$}{Pw(f0)}}\label{sec:SLP_analysis}

Consider a binary classification problem using a single hidden layer perceptron. We will use the notation outlined in Fig.~\ref{NN_scheme}. The nodes of the hidden layer, after processing through the activation function and the pooling layer, are connected to two output nodes \scalebox{\BP}{$\left\{ \ROutNode{c}{} \right\}_{c=0}^{1}$} via two random vectors \scalebox{\BP}{$\left\{ \Weights{1}{c \; \Cdot} \right\}_{c=0}^1$}. Each output node \scalebox{\BP}{$ \ROutNode{c}{} \left( \boldsymbol{\InputValue}; \WeightSet{0}, \Weights{1}{c \; \Cdot} \right)$} follows the distributions derived in App.~\ref{sec:Cond_PO}.

Note that, given a set $\WeightSet{}$, different output nodes depend on distinct subsets of $\WeightSet{}$, as the final matrix of weights is not shared among all nodes. The notation \scalebox{\BP}{$ \ROutNode{c}{} \left( \boldsymbol{\InputValue}; \WeightSet{0}, \Weights{1}{c \; \Cdot} \right)$} highlights this aspect, underscoring why the output nodes have different distributions. However, for brevity, we will denote a general dependency on the entire set $\WeightSet{}$, \emph{i.e.}, we will use the notation \scalebox{\BP}{$ \ROutNode{c}{} \left( \boldsymbol{\InputValue}; \WeightSet{} \right)$}.

To derive the distribution \scalebox{\BP}{$\pdf{\RClassFraction{0}}{}{\ClassFraction{0}}$}, we proceed as follows: \newline
In a single experiment, the set \scalebox{\BP}{$\left\{ \Weights{1}{c \; \Cdot} \right\}_{c=0}^1$} is fixed, allowing us to compute \scalebox{\BP}{$\pdf{\ROutNode{c}{}}{(\Dataset)}{\OutNode{}{}} $}. Recalling that $\RClassFraction{0}$ represents the fraction of the data-points guessed as members of class $0$, according to the \emph{Law of Large Numbers}, this fraction will converge to the probability
\begin{equation}
\lim_{\DatasetSize \rightarrow \infty} \RClassFraction{0} \left( \WeightSet{} \right) = \Prob{}{ \ROutNode{0}{} >  \ROutNode{1}{} \mid \WeightSet{} } = \Prob{}{  \ROutNode{0}{} - \ROutNode{1}{} > 0  \mid \WeightSet{} } =  \int_0^{\infty} \pdf{\RDiff{\ROutNode{}{}}}{(\Dataset)}{\Diff{\OutNode{}{}}} \; d \Diff{\OutNode{}{}},
\end{equation}
where $\RClassFraction{0} \left( \WeightSet{} \right)$ indicates the fraction 
$\RClassFraction{0}$ obtained for the specific fixed configuration $\WeightSet{}$.

$\RClassFraction{0}$ is a quantity clearly dependent on the set of random variables $\WeightSet{}$\blocco{\scalebox{\BP}{$\{ \Weights{1}{c \; \Cdot} \}_{c=0}^1$}}. Consequently, it is a random variable itself, varying with the network's set of weights. Defining \scalebox{\BP}{$\RDiff{\ROutNode{}{}} = \ROutNode{0}{} - \ROutNode{1}{} $}, we can express:
\begin{equation}\label{Pf0_expression}
\pdf{\RClassFraction{0}}{}{\ClassFraction{0}} = \int_{\Omega}  \pdf{\WeightSet{}}{}{ \tilde{\WeightSet{}} } \delta \left(  \Prob{}{\Delta_{O} >0 \mid \tilde{\WeightSet{}}  } - \ClassFraction{0} \right) d\tilde{\WeightSet{}} 
\end{equation}
where $\Omega$ represents the space of all possible initial configurations.

Eq.~\eqref{Pf0_expression} provides an expression for $\pdf{\RClassFraction{0}}{}{\ClassFraction{0}}$. However, computing the integral on the right-hand side (r.h.s.) is not straightforward. To circumvent this, we introduce a second approach that is asymptotically exact in the limit of $\LayerNumNodes{1} \rightarrow \infty$. 

In App.~\ref{sec:Cond_PO}, we derived expressions for \scalebox{\BP}{$\pdf{\ROutNode{c}{}}{(\Dataset)}{\OutNode{}{}}$} for different cases. These distributions are generally Gaussian, with cumulants dependent on \scalebox{\BP}{$S_{\boldsymbol{w}_c} \equiv \sum_j^{\LayerNumNodes{1}} \Weights{1}{cj}$} and \scalebox{\BP}{$S_{\boldsymbol{w}_c^2} \equiv \sum_j^{\LayerNumNodes{1}} \left( \Weights{1}{cj} \right)^2$}. The standard deviations can be viewed as estimates for the magnitude of fluctuations from the mean value. 

In App.~\ref{sec:RescGaussVar}, we show that:

\begin{align}
& \sqrt{\Eweights{ S_{\boldsymbol{w}_c} - \Eweights{S_{\boldsymbol{w}_c}}}^2 } = \mathcal{O} (1), \\
& \sqrt{\Eweights{S_{\boldsymbol{w}_c^2} - \Eweights{S_{\boldsymbol{w}_c^2}}}^2 } = \mathcal{O} \left( \frac{1}{\sqrt{\LayerNumNodes{1}}} \right). \label{Fluct_Sa^2}
\end{align}

In other words, the distribution of $S_{\boldsymbol{w}_c}$ remains asymptotically stable, whereas the distribution of $S_{\boldsymbol{w}_c^2}$ concentrates around its mean value as we approach the asymptotic limit. This distinction in the scaling of fluctuations between the two cases will be significant in our subsequent discussion.

To establish a unified notation, we can generalize from App.~\ref{sec:Cond_PO} that:
\begin{equation}
\pdf{\ROutNode{c}{}}{(\Dataset)}{\OutNode{}{}} = \NormalDens{\OutNode{}{}}{\mu \left( S_{\boldsymbol{w}_c} \right)}{\sigma^2 \left( S_{\boldsymbol{w}_c^2} \right)},
\end{equation}
where \scalebox{\BP}{$\mu \left( S_{\boldsymbol{w}_c} \right)$} and \scalebox{\BP}{$\sigma \left( S_{\boldsymbol{w}_c^2} \right)$} are functions of the set of \textit{r.v.}s  \scalebox{\BP}{$\{\Weights{1}{cj} \}_{j=1}^{\LayerNumNodes{1}}$}, contingent on the specific setting (e.g., the choice of activation function). App.~\ref{sec:E_chi_O} provides the distribution for \scalebox{\BP}{$\mu \left( S_{\boldsymbol{w}_c} \right)$} in various scenarios. Conversely, from Eq.~\eqref{Fluct_Sa^2}, we know that the distribution of \scalebox{\BP}{$\sigma \left( S_{\boldsymbol{w}_c^2} \right)$} becomes increasingly narrow around its mean values as $\LayerNumNodes{1} \rightarrow \infty$, thus converging to a deterministic quantity.

Since the distribution of \scalebox{\BP}{$\mu \left( S_{\boldsymbol{w}_c} \right)$} does not narrow asymptotically around a deterministic value, \scalebox{\BP}{$\pdf{\ROutNode{c}{}}{(\Dataset)}{\OutNode{}{}}$} will differ for various nodes, \scalebox{\BP}{$\{ \ROutNode{c}{} \}$}, even in the limit of infinite size. We refer to the emergence of this discrepancy in distributions among nodes of the same layer as \emph{Node Symmetry Breaking} (NSB). Section~\ref{sec:BigPicture} elucidates how NSB and IGB are interrelated phenomena, with IGB being a direct consequence of this symmetry breaking. In Sec.~\ref{sec:PSB}, we will delve deeper into NSB, illustrating how this phenomenon manifests not only in output layers but also in intermediate hidden layers of deep architectures.
\newline

Returning to the initial problem, let's consider a given initialization fixing the configuration $\WeightSet{}$, and in particular, the two random vectors \scalebox{\BP}{$\Weights{1}{0 \; \Cdot}$} and \scalebox{\BP}{$\Weights{1}{1 \; \Cdot}$}. This leads to two corresponding distributions for \scalebox{\BP}{$\pdf{\ROutNode{0}{}}{(\Dataset)}{\OutNode{}{}}$} and \scalebox{\BP}{$\pdf{\ROutNode{1}{}}{(\Dataset)}{\OutNode{}{}}$}. To calculate the probability that \scalebox{\BP}{$\ROutNode{0}{} > \ROutNode{1}{}$}, we first transition to the distribution \scalebox{\BP}{$\pdf{\RDiff{\ROutNode{}{}}}{(\Dataset)}{\Diff{\OutNode{}{}}}$}. Since $\RDiff{\ROutNode{}{}}$ is the difference between two Gaussian random variables, it follows that:
\begin{equation}
\pdf{\RDiff{\ROutNode{}{}}}{(\Dataset)}{\Diff{\OutNode{}{}}} = \NormalDens{\Diff{\OutNode{}{}}}{\mu \left( S_{\boldsymbol{w}_0} \right)  - \mu \left( S_{\boldsymbol{w}_1} \right)}{\sigma^2 \left( S_{\boldsymbol{w}_0^2} \right)   + \sigma^2 \left( S_{\boldsymbol{w}_1^2} \right)}.
\end{equation}
From Eq.~\eqref{Fluct_Sa^2}, we understand that the distribution of $S_{\boldsymbol{w}_c^2}$ narrows around its mean in the limit $\LayerNumNodes{1} \rightarrow \infty$, allowing us to disregard the fluctuations of the random variable \scalebox{\BP}{$\sigma^2 \left( S_{\boldsymbol{w}_c^2} \right)$}. Substituting \scalebox{\BP}{$\sigma^2 \left( S_{\boldsymbol{w}_c^2} \right) = \sigma^2 \left( \Eweights{S_{\boldsymbol{w}_c^2}} \right) \equiv \sigma^2_{\infty} $}, we obtain:

\begin{equation}
\pdf{\RDiff{\ROutNode{}{}}}{(\Dataset)}{\Diff{\OutNode{}{}}}  \xrightarrow{\LayerNumNodes{1} \rightarrow\infty}  \NormalDens{\Diff{\OutNode{}{}}}{\mu \left( S_{\boldsymbol{w}_0} \right)  - \mu \left( S_{\boldsymbol{w}_1} \right)}{2 \sigma^2_{\infty}} .
\end{equation}
This convergence is pivotal to eliminate the explicit dependence on the random vectors \scalebox{\BP}{$\{ \Weights{1}{c \; \Cdot} \}$} and to avoid integration over all possible configurations. Once we have an explicit form for the distribution of \scalebox{\BP}{$\mu \left( S_{\boldsymbol{w}_c} \right)$} (see App.~\ref{sec:E_chi_O}), and defining
\begin{equation}
\Delta_{\mu} = \mu \left( S_{\boldsymbol{w}_0} \right)  - \mu \left( S_{\boldsymbol{w}_1} \right),
\end{equation}
we can find an implicit expression for $\Delta_{\mu}$, \textit{i.e.},
\begin{equation}\label{D_mu_implicit}
 \RClassFraction{0} \left( \WeightSet{} \right) =
 \int_{0}^{\infty} \NormalDens{\Diff{\OutNode{}{}}}{\Delta_{\mu}(\RClassFraction{0})}{2 \sigma^2_{\infty}} d\Diff{\OutNode{}{}}.
\end{equation}
Given the centers of the two Gaussian distributions \scalebox{\BP}{$\pdf{\ROutNode{c}{}}{(\Dataset)}{\OutNode{}{}}$}, specifically their difference $\Delta_{\mu}$, Eq.~\eqref{D_mu_implicit} provides the corresponding value of $\RClassFraction{0}$. Inverting Eq.~\eqref{D_mu_implicit} numerically yields $\Delta_{\mu}(\RClassFraction{0})$ associated with a given value of $\RClassFraction{0}$. 

Notably, \scalebox{\BP}{$\mu \left( S_{\boldsymbol{w}_c} \right) \sim \pdf{\RMean{c}}{}{\Mean{}}$} is a \textit{r.v.}; consequently, $\Delta_{\mu}(\RClassFraction{0})$ is a \textit{r.v.} as well. From the monotonicity of the relationship $\Delta_{\mu}(\RClassFraction{0})$, it follows that

\begin{align}
    \pdf{\RClassFraction{0}}{}{\ClassFraction{0}} \, d\ClassFraction{0} = \pdf{\Delta_{\mu} (\RClassFraction{0})}{}{\Diff{\Mean{}}(\ClassFraction{0})} \, d\Diff{\Mean{}}
\end{align}

Since \scalebox{\BP}{$\mu \left( S_{\boldsymbol{w}_0} \right) $} and \scalebox{\BP}{$\mu \left( S_{\boldsymbol{w}_1} \right)$} are i.i.d. random variables, we can deduce:
\begin{align}\label{eq:P_f0_MF}
\begin{split}
 \pdf{\Delta_{\mu} (\RClassFraction{0})}{}{\Diff{\Mean{}}(\ClassFraction{0})} = \int_{-\infty}^{\infty} \pdf{\EinputComp{1}}{}{\Mean{}} \pdf{\EinputComp{0}}{}{\Mean{} + \Diff{\Mean{}}(\ClassFraction{0})} d \Mean{}
 = \int_{-\infty}^{\infty} \pdf{\EinputComp{c}}{}{\Mean{}} \pdf{\EinputComp{c}}{}{\Mean{} + \Diff{\Mean{}}(\ClassFraction{0})} d \Mean{},
\end{split}
\end{align}
where the integral arises from the fact that we only enforce a condition on the difference between the nodes' mean values, which is invariant under a translation of both values. The final equality leverages the identical distribution of \scalebox{\BP}{$\EinputComp{0}$} and \scalebox{\BP}{$\EinputComp{1}$}.

Given that \scalebox{\BP}{$\pdf{\EinputComp{c}}{}{\Mean{}} \xrightarrow{\LayerNumNodes{1} \rightarrow\infty} \NormalDens{\Mean{}}{0}{\hat{\sigma}^2_{\infty}} $},\footnote{The quantity $\hat{\sigma}^2_{\infty}$ is model-specific; for more details, see App.~\ref{sec:E_chi_O}.} Eq.~\eqref{eq:P_f0_MF} represents an integral of a product of two independent Gaussian distributions.\footnote{The product of Gaussian distributions itself is proportional to a Gaussian distribution.} Solving this integral gives us \blocco{(refer to \cite{duda2018gaussian} for details)}
\begin{equation}\label{P_f0}
\pdf{\Delta_{\mu} (\RClassFraction{0})}{}{\Diff{\Mean{}}(\ClassFraction{0})}= \NormalDens{\Diff{\Mean{}}(\ClassFraction{0})}{0}{2 \hat{\sigma}^2_{\infty}}.
\end{equation}

\paragraph{Remark 2.}
While Eq.~\eqref{P_f0} illustrates a Gaussian distribution in terms of the variable $\Delta_{\mu}(\RClassFraction{0})$, it is important to note that $\pdf{\RClassFraction{0}}{}{\ClassFraction{0}}$ will not generally be Gaussian since $\Delta_{\mu}\left( \RClassFraction{0} \right)$ is a non-linear function.

\paragraph{{max pool peaks: A practical example for computing \texorpdfstring{$\boldsymbol{\pdf{\RClassFraction{0}}{}{\ClassFraction{0}}}$}{Pw(f0)}}}
Utilizing the results from App.~\ref{sec:SLP_analysis}, we can calculate the theoretical prediction for $\pdf{\RClassFraction{0}}{}{\ClassFraction{0}}$, particularly focusing on the probability density at extreme peaks empirically observed with max pooling.

Suppose we have a dataset of $\DatasetSize$ elements, and we aim to calculate the probability density $\pdf{\RClassFraction{0}}{}{0}$. The steps involved are:

\begin{itemize}
\item \textbf{Computing $\Delta^{(T)}$:} 
First, we need to invert Eq.~\eqref{D_mu_implicit}:
\begin{equation}
\ClassFraction{0} = 1 - \left(\frac{1}{2} + \frac{1}{2} \erf\left( \frac{\Delta_{\mu}(\ClassFraction{0})}{2 \sigma_{\infty}} \right) \right)  
\Rightarrow 
\Delta_{\mu}(\ClassFraction{0}) = 2 \sigma_{\infty} \erf^{-1} \left(2\ClassFraction{0} -1 \right).
\end{equation}
$\Delta^{(T)}$ is the threshold value; for $\Delta_{\mu}(\ClassFraction{0}) < \Delta^{(T)}$, Eq.~\eqref{D_mu_implicit} yields a value below $\frac{1}{\DatasetSize}$, corresponding to $\ClassFraction{0}=0$ due to the dataset size $\DatasetSize$ resolution.

\item \textbf{Integrating Eq.~\eqref{P_f0}:} 
Next, we integrate over all $\Delta_{\mu}(\ClassFraction{0}) < \Delta^{(T)}$:
\begin{equation}
\pdf{\RClassFraction{0}}{}{0} = \int_{-\infty}^{\Delta^{(T)}} \NormalDens{\Diff{\Mean{}}}{0}{2 \hat{\sigma}^2_{\infty}}  d \Diff{\Mean{}} =  \frac{1}{2} + \frac{1}{2} \erf \left( \frac{\Delta^{(T)}}{2 \hat{\sigma}_{\infty}}  \right).
\end{equation}
\end{itemize}

\textbf{Generalizing for arbitrary probability mass:} For the probability mass between $\ClassFraction{0}^{(min)}$ and $\ClassFraction{0}^{(max)}$, the computation is:
\begin{align}
\begin{split}
&\int_{\ClassFraction{0}^{(min)}}^{\ClassFraction{0}^{(max)}}  \pdf{\RClassFraction{0}}{}{\ClassFraction{0}}  d\ClassFraction{0} = \int_{\Delta_{\mu} \left(\ClassFraction{0}^{(min)} \right)}^{\Delta_{\mu} \left(\ClassFraction{0}^{(max)} \right)} \NormalDens{\Diff{\Mean{}}}{0}{2 \hat{\sigma}^2_{\infty}} d\Diff{\Mean{}} =\frac{1}{2} \erf \left( \frac{\Delta_{\mu} \left(\ClassFraction{0}^{(max)} \right)}{2 \hat{\sigma}_{\infty}}  \right) -\frac{1}{2} \erf \left( \frac{\Delta_{\mu} \left(\ClassFraction{0}^{(min)} \right)}{2 \hat{\sigma}_{\infty}}  \right).
\end{split}
\end{align}

These computations facilitate a precise understanding of the impact of max pooling on $\RClassFraction{0}$, especially in terms of evaluating probability densities at the extreme peaks.

\subsection{Beyond \texorpdfstring{$\boldsymbol{\RClassFraction{0}}$}{f0}: relating the confidence in model assignments to IGB}\label{app:conf_ass}

While our discussion is based on an intuitive and clearly interpretable measure, namely $\RClassFraction{0}$ and its distribution, this measure has a significant limitation. $\RClassFraction{0}$ indicates the fraction of the dataset assigned to class 0 by the network at initialization, but it doesn't provide information about the average confidence with which this assignment is made for a given dataset element. 

It is important to note that, although $\RClassFraction{0}$ is a central focus of our narrative for its intuitive clarity, it is not the main focus of our analysis. Our analyses are instead centered around the output nodes, from which we derive distributions, and only subsequently, through these and Eq.~\eqref{eq:fi_def_LLN}, we transition to the variable $\RClassFraction{0}$. \newline
Therefore, we can leverage our knowledge of the output nodes' statistics to investigate different measures. Typically, to convert the outputs into probabilities or, to better say, into confidence levels, a softmax transformation is applied, defined as:
\begin{align}\label{eq:sm_out}
    \ROutNode{c}{} \rightarrow \SMO{c}{} \equiv \frac{e^{ \ROutNode{c}{}}}{\sum_i e^{ \OutNode{i}{}}}\,.
\end{align}

We note that this metric provides additional insights, leading to more stringent conditions than just the absence of IGB. Specifically, when every element of the dataset fulfills the condition
\begin{align}\label{eq:sa_IGB}
    \SMO{c}{} = \frac{1}{\NumberClasses} \; \forall c \,,
\end{align}
it implies not only the absence of IGB (wherein the DNN's classifications distribute the dataset evenly across classes), but also that every assignement has an equal probability for each class.
This leads us to the concept of Strong Absence of IGB, which we define as follows:
\begin{defbox}[Strong Absence of IGB]
Given a dataset $\Dataset$ (or its pre-processed version $\psi(\Dataset)$) and an architecture $\Arch$, there is a strong absence of Initial Guessing Bias (IGB) if, for every element in the dataset and for every initialization of weights \textit{w.h.p.}, the model predicts each class with the same level of confidence, i.e., Eq.~\eqref{eq:sa_IGB} is satisfied.
\end{defbox}

With binary classification, we can, in general, consider a measure to evaluate the gap between assignment probabilities, defined as:
\begin{align}\label{eq:RP}
    \ProbRat \equiv \frac{\SMO{0}{}}{\SMO{1}{}} = e^{\Delta_O}\,,
\end{align}
where we  remind the reader that $\Delta_O \equiv \ROutNode{0}{}-\ROutNode{1}{}$.

Since we have shown that the distribution of a generic output, $\pdf{\ROutNode{c}{}}{(\Dataset)}{\OutNode{}{}}$, is Gaussian (see Eq.~\eqref{eq:PO_asym_app_thm}), it follows that
\begin{align}\label{eq:PdeltaO}
  \pdf{\Delta_O}{(\Dataset)}{\Diff{\OutNode{}{}}}  = \NormalDens{\Diff{\OutNode{}{}}}{\Delta_{\mu}}{2\CVar{\Dataset}{\ROutNode{c}{}}}\,,
\end{align}
where $\Delta_{\mu} \equiv \EinputComp{0} - \EinputComp{1}$.
With a given initialization $\WeightSet{}$, the quantity $\Delta_{\mu}$ is deterministically fixed. Combining Eqs.~\eqref{eq:RP} and \eqref{eq:PdeltaO}, we can explicitly write $\ProbRat$ as a log-normal distribution:
\begin{align}\label{eq:log-norm_dist}
    \pdf{\ProbRat}{(\Dataset)}{r} = \frac{1}{\Delta_{\mu} \sqrt{4 \pi \CVar{\Dataset}{\ROutNode{c}{}}}} e^{- \frac{(\ln (r) - \Delta_{\mu})^2 }{4 \CVar{\Dataset}{\ROutNode{c}{}} }} .
\end{align}
The first two cumulants (over the distribution of the data) of $\pdf{\ProbRat}{(\Dataset)}{r}$ are:
\begin{align}
    \Einput{\ProbRat} &= e^{\Delta_{\mu} + \CVar{\Dataset}{\ROutNode{c}{}}}\,, \label{eq:mean_Rp} \\
    \CVar{\Dataset}{\ProbRat} &= e^{2\Delta_{\mu}} \left( e^{2 \CVar{\Dataset}{\ROutNode{c}{}}} -1\right) e^{2 \CVar{\Dataset}{\ROutNode{c}{}}} \,.
    \label{eq:var_Rp}
\end{align}

Equations \eqref{eq:mean_Rp} and \eqref{eq:var_Rp} demonstrate that:
\begin{align}
    \lim_{\Delta_{\mu}\to0}~ \lim_{\CVar{\Dataset}{\ROutNode{c}{}} \to 0} \Einput{\ProbRat} &= 1\,,\\[1ex]
    \lim_{\CVar{\Dataset}{\ROutNode{c}{}} \to 0} \CVar{\Dataset}{\ProbRat} &= 0\,.
\end{align}
In this limit, not only do we have an absence of IGB, but the neural network assigns each input to either of the two classes with the same probability (since the distribution concentrates around a single value), \textit{i.e.} we have strong absence of IGB. This constitutes the most extreme limit case: in the limit $\DatasetSize \rightarrow \infty$, half of the dataset will be assigned to each class, and each element will be assigned to one of the two classes with the same probability ($\ProbRat = 1$). 
\newline

We may encounter a scenario where $\Delta_{\mu} = 0$ and $\CVar{\Dataset}{\ROutNode{c}{}} \neq 0$. Even though this indicates an absence of IGB, the distribution for $\ProbRat$ is non-degenerate. Like the previous case, the dataset will be evenly divided in assignments between the different classes. However, the typical confidence is different for each class (it is described by Eq.~\eqref{eq:log-norm_dist}).

In the presence of IGB, $\Delta_{\mu}$ will have a non-degenerate distribution (see Eq.~\eqref{eq:P_cen_asym_app_thm}). As the level of IGB increases (by increasing $\CVar{\WeightSet{}}{\EinputComp{c}}$), there will also be an increase in $\Eweights{\Einput{\ProbRat}}$, meaning the typical gap between the two assignment probabilities for a generic element will widen.\newline
It is important to note that, given a specific setting $(\Arch, \PreprData)$, we can explicitly calculate Eq.~\eqref{eq:mean_Rp} and Eq.~\eqref{eq:var_Rp} based on the output statistics and evaluate the network's confidence in its assignments.

In conclusion, the approach shown in this section paves the way for more accurate analyses. For example, knowing the $\{\SMO{c}{}\}$, we could calculate the entropy $H\left(\{\SMO{c}{}\} \mid \WeightSet{}, \Dataset \right)$ and reformulate the absence of IGB as a principle of maximum entropy. However, it is important to note that the presence of the normalization factor complicates the derivation of the distribution of $\SMO{c}{}$, which can be avoided by using $\ProbRat$. This also provides a natural variable to encapsulate the gap between the assignment variables. Finally, although we focused on the binary problem for simplicity, extensions to the multi-class case are possible using approaches similar to those described in App.~\ref{sec:MC_extension}, for example by considering the ratio of the two classes with the highest confidence.

\subsubsection{Effect of the Softmax temperature}\label{sec:sm_temp}
Eq.~\eqref{eq:sm_out} is the standard softmaxed expression of the output with the temperature parameter $T=1$.
Acting on the temperature can effectively reduce IGB. By applying a high-temperature parameter, we observe a narrowing distribution of $\ProbRat$ around $\ProbRat=1$, indicating a reduction in IGB as $T \to \infty$. However, it is crucial to acknowledge the complex effects of temperature on training dynamics; specifically, a small beta value (high temperature) may affect learning stability \citep{agarwala2020temperature}. While further investigation is needed to fully understand the impact of this mitigation strategy on the training process.

\section{Conditions for the emergence/absence of IGB}\label{sec:IGB_cond}

In our exploration of how activation functions can induce IGB, we have focused on specific, emblematic examples, such as the linear activation function and the ReLU. These were chosen for their analytical tractability. However, the underlying principles extend beyond these cases. In this section, we'll broaden the discussion, classifying activation functions into distinct categories based on their influence on IGB.

The crucial distinction between the activation functions we've discussed can be summarized as follows: 

1. With a linear activation function, the output nodes are asymptotically identically distributed. This uniformity leads to the absence of IGB. 
2. In contrast, the ReLU activation function introduces a symmetry break in the output nodes. While these nodes remain asymptotically Gaussian, they are centered at different points, leading to the emergence of IGB.

This symmetry breaking is intricately linked to how the ReLU, applied at the first hidden layer, transforms null-averaged inputs into random variables, $\RCompHidNodeAAF{1}{i}$, with a non-zero mean. 

To generalize our findings, we'll now examine a typical activation function from each category and demonstrate how the presence or absence of IGB manifests. Our analysis will maintain the same foundational assumptions as before: Gaussian-distributed \emph{i.i.d.} inputs and Kaiming initialization for the weights. As established in App.~\ref{sec:E_chi_O}, regardless of the activation function, we observe that:
\begin{align}
    \pdf{\RCompHidNodeBAF{1}{i}}{(\Dataset)}{\ArgHidNodeBAF{}{}} \xrightarrow{d\rightarrow\infty}
 \NormalDens{\ArgHidNodeBAF{}{}}{0}{\sigma^2_w} .
\end{align}

The forthcoming analysis will elucidate the conditions under which different types of activation functions either foster or mitigate the occurrence of IGB in neural network models.

\subsection{Non-null mean activation function}
Let us consider a generic activation function $\AFOperator{\cdot}$, such that
\begin{align}\label{eq:N-N_subset_cond}
 \Einput{\RCompHidNodeAAF{1}{i}} \equiv \Einput{\AFOperator{\RCompHidNodeBAF{1}{i}}} \stackrel{}{=} \Einput{\RCompHidNodeAAF{1}{}} \neq 0  .    
\end{align}
where the second step follow from the fact that  $\{\RCompHidNodeBAF{1}{i}\}$ are identically distributed. 
Starting from this hypothesis on the activation function (Eq.~\eqref{eq:N-N_subset_cond}) we can follow exactly the same analysis presented in App.~\ref{sec:MLP}. In particular:
\begin{align}\label{eq:h2_dist_nn_af}
\begin{split}
    &\pdf{\RCompHidNodeBAF{2}{i}}{(\Dataset)}{\ArgHidNodeBAF{}{}} \xrightarrow{\LayerNumNodes{1} \rightarrow \infty} \NormalDens{\ArgHidNodeBAF{}{}}{\Einput{\RCompHidNodeAAF{1}{}} S_{\boldsymbol{w}_i}}{\CVar{\Dataset}{\RCompHidNodeAAF{1}{}} S_{\boldsymbol{w}_i^2}}  =\NormalDens{\ArgHidNodeBAF{}{}}{\Einput{\RCompHidNodeAAF{1}{}} S_{\boldsymbol{w}_i}}{\CVar{\Dataset}{\RCompHidNodeAAF{1}{}} \Eweights{S_{\boldsymbol{w}_i^2}}}\, ,
    \end{split}
\end{align}
From \eqref{eq:h2_dist_nn_af} we can already see the nodes symmetry breaking; the centers of the distributions is a \textit{r.v.} varying from node to node. Note that for a single hidden layer architecture 
\begin{align}
\pdf{\ROutNode{c}{}}{(\Dataset)}{\OutNode{}{}} = \pdf{\RCompHidNodeBAF{2}{c}}{(\Dataset)}{\OutNode{}{}}\,.
\end{align}
For a generic deep architecture we can keep following the analysis of App.~\ref{sec:MLP}, substituiting the right expression to $\pdf{\RCompHidNodeAAF{2}{i}}{(\Dataset)}{\ArgHidNodeAAF{}{}}$ dependent to the specific used activation function.
Proceeding in this way we will find, for each layer, a symmetry breaking in the nodes distribution, similarly to what we observed in the second layer (Eq.~\eqref{eq:h2_dist_nn_af}). 
\subsection{Null mean activation function}
We now explore the converse scenario, namely an activation function $\AFOperator{\cdot}$ satisfying:
\begin{align}\label{eq:null_mean_AF_cond}
 \Einput{\RCompHidNodeAAF{1}{i}} \equiv \Einput{\AFOperator{\RCompHidNodeBAF{1}{i}}} \stackrel{}{=} \Einput{\RCompHidNodeAAF{1}{}} = 0.
\end{align}
This equality arises because the set $\{\RCompHidNodeBAF{1}{i}\}$ is identically distributed. Unlike the non-null mean case (\eqref{eq:h2_dist_nn_af}), this scenario leads to:
\begin{align}\label{eq:h2_dist_n_af}
    \pdf{\RCompHidNodeBAF{2}{i}}{(\Dataset)}{\ArgHidNodeBAF{}{}} \xrightarrow{\LayerNumNodes{1} \rightarrow \infty} \NormalDens{\ArgHidNodeBAF{}{}}{0}{\CVar{\Dataset}{\RCompHidNodeAAF{1}{}} S_{\boldsymbol{w}_i^2}}  = \NormalDens{\ArgHidNodeBAF{}{}}{0}{\CVar{\Dataset}{\RCompHidNodeAAF{1}{}} \Eweights{S_{\boldsymbol{w}_i^2}}} .
\end{align}
Here, the absence of node symmetry breaking is evident. Proceeding to subsequent layers, we find a similar pattern of asymptotically identically distributed nodes. This iterative process ultimately extends to the output nodes, which will also be equally distributed.\newline
Determining whether a generic activation function falls into one of the two categories discussed may not be straightforward. To illustrate, we present an example set of functions, all of which fulfill Condition \eqref{eq:null_mean_AF_cond}.

\subsubsection{Anti-Symmetric activation functions}\label{sec:AS_act_func}
The previous discussions have underscored how the choice of activation function significantly influences a network's behavior, notably in the context of IGB (refer to Fig.~\ref{fig:Comp_SL_Pf0}). While the trivial identity function does not exhibit IGB, this section introduces a broader class of activation functions, the anti-symmetric activation functions, which similarly do not exhibit IGB. Interestingly, the identity function is a member of this class.

We begin by reinterpreting the concept of IGB within the context of our analysis:

\begin{defbox}[IGB]
Assume $\pdf{\ROutNode{c}{}}{(\Dataset)}{\OutNode{}{}} $ asymptotically converges to a Gaussian distribution with non-zero variance (see App.~\ref{sec:Cond_PO}). If 
\begin{equation}\label{eq:delta_P_muo}
\pdf{\EinputComp{c}}{}{\Mean{}}  = \Dirac{\Mean{}-a}, \; \forall c \,,
\end{equation}
for some $a \in \mathbb{R}$, we have an absence of IGB. Conversely, if condition \eqref{eq:delta_P_muo} is not satisfied, IGB emerges, leading to disproportionate values of $\{ \ClassFraction{i} \}$ even in the infinite dataset size limit, thus excluding finite size effects.
\label{def:IGB_CS}
\end{defbox}

To illustrate the concept with a practical example, consider an activation function defined as
\begin{equation}\label{eq:tanh_AF_def}
\RCompHidNodeAAF{l}{i} = g\left( \RCompHidNodeBAF{l}{i} \right) = \tanh \left( \RCompHidNodeBAF{l}{i} \right), \; l \in \{ 0, \dots, L \}, \; i \in \{ 0, \dots, \LayerNumNodes{l} \}.
\end{equation}
Recalling Eq.~\eqref{eq:h1_distr}, the asymptotic distribution of $\pdf{\RCompHidNodeBAF{1}{i}}{(\Dataset)}{\ArgHidNodeBAF{}{}}$ is known and remains independent of the activation function choice. The question arises: "How is this distribution transformed after passing through the activation function (Eq.~\eqref{eq:tanh_AF_def})?"

By employing the relation 
\begin{equation}
\pdf{X}{}{ x}dx = \pdf{Y}{}{ y(x)}dy,
\end{equation}
we derive 
\begin{equation}\label{eq:tanh_distr_expr}
\pdf{\RCompHidNodeAAF{1}{i}}{(\Dataset)}{\ArgHidNodeAAF{}{}} = \frac{e^{-\frac{(\atanh(\ArgHidNodeAAF{}{}))^2}{2 \sigma^2_{\RCompHidNodeBAF{1}{} }}}}{\sqrt{2 \pi \sigma^2_{\RCompHidNodeBAF{1}{} } }} \frac{1}{(1-\ArgHidNodeAAF{}{}^2)}.
\end{equation}
We observe from Eq.~\eqref{eq:tanh_distr_expr} that the distribution is symmetric, i.e., $\pdf{\RCompHidNodeAAF{1}{i}}{(\Dataset)}{\ArgHidNodeAAF{}{}} = \pdf{\RCompHidNodeAAF{1}{i}}{(\Dataset)}{-\ArgHidNodeAAF{}{}}$. This symmetry arises directly from the anti-symmetric nature of the activation function combined with the symmetry of $\pdf{\RCompHidNodeBAF{1}{i}}{(\Dataset)}{\ArgHidNodeBAF{}{}}$. This would also hold true for other anti-symmetric activation functions.

As a result of this symmetry, we have:
\begin{equation}
\Einput{\RCompHidNodeAAF{1}{i}} = 0 \Longrightarrow \Einput{\RCompHidNodeBAF{2}{j}} = 0, \;\;\; \forall i,j.
\end{equation}

This is a stark contrast to the scenarios discussed in the previous section, as it means all nodes $\{ \RCompHidNodeBAF{2}{j} \}$ are identically distributed. Specifically,
\begin{equation}
\pdf{\Einput{\RCompHidNodeBAF{2}{i}}}{}{\Mean{\ArgHidNodeBAF{}{}}} = \Dirac{\Mean{\ArgHidNodeBAF{}{}}} \; \forall i.
\end{equation}

As we continue to propagate through the network's remaining hidden layers, this pattern persists. Given the similarity between $\pdf{\RCompHidNodeBAF{2}{i}}{(\Dataset)}{\ArgHidNodeBAF{}{}}$ and $\pdf{\RCompHidNodeBAF{1}{i}}{(\Dataset)}{\ArgHidNodeBAF{}{}}$, the same reasoning can be applied layer by layer, leading to
\begin{equation}
\pdf{\Einput{\RCompHidNodeBAF{3}{i}}}{}{\Mean{\ArgHidNodeBAF{}{}}} = \Dirac{\Mean{\ArgHidNodeBAF{}{}}} \; \forall i,
\end{equation}
and this holds for all $l \in \{ 3, \dots, L+1 \}$.

Specifically, for $l = L+1$, it implies that
\begin{equation}
\pdf{\EinputComp{c}}{}{\Mean{}} = \Dirac{\Mean{}} \; \forall c.
\end{equation}

Therefore, by employing an anti-symmetric activation function such as $\tanh$, IGB is not observed, under the assumption that the chosen initialization maintains a non-zero asymptotic variance for $\pdf{\ROutNode{c}{}}{(\Dataset)}{\OutNode{}{}}$. 

\subsection{Eliminate/Trigger IGB with a generic activation function}\label{sec:subset_shift}

\begin{figure}
\centering
    \includegraphics[width=0.7\textwidth]{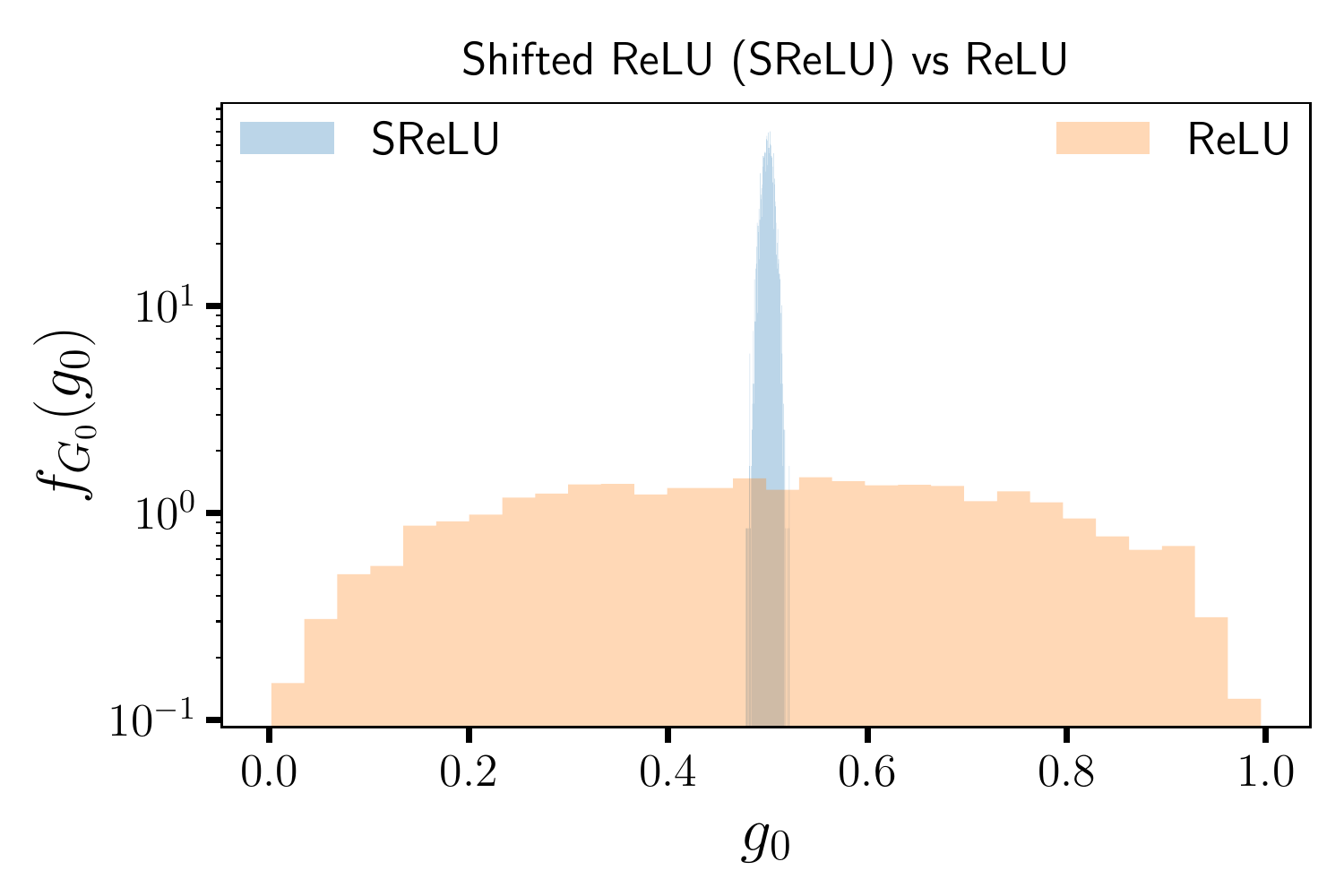}
  \caption{Contrast between SReLU and ReLU. Shifting ReLU to meet Eq.~\eqref{eq:h2_dist_n_af} conditions eliminates IGB. The figure shows the differing regimes through $\pdf{\RClassFraction{0}}{}{\ClassFraction{0}}$. Dataset \GB~ and model \MLPA~ were used for these simulations (see App.~\ref{sec:reprod} for details).}
  \label{fig:Comp_ReLU_SReLU}
\end{figure}

The preceding discussion reveals that activation functions can be divided into two categories based on their influence on IGB. These categories are defined by simple attributes (\emph{e.g.}, Eq.~\eqref{eq:N-N_subset_cond} and Eq.~\eqref{eq:null_mean_AF_cond}). This insight suggests that modifying an existing activation function, or applying certain regularizations, can switch a network from exhibiting IGB to one that doesn't, and vice versa.

This concept aligns with strategies used in machine learning, such as Batch Normalization introduced in \citet{ioffe2015batch}, which is employed to mitigate the \emph{Internal Covariate Shift}. While Batch Normalization addresses shifts during training, our focus is on initialization-induced node differences. Nevertheless, our analysis suggests that a similar scaling approach could effectively manage IGB as well.

To illustrate, consider the ReLU activation function, which induces IGB. We introduce a shifted version of ReLU, named SReLU, defined as follows:
\begin{align}
    \AFOperator{\RCompHidNodeBAF{1}{i}} = \begin{cases}
        & \RCompHidNodeBAF{1}{i} - \frac{\sigma_w}{\sqrt{2 \pi}},  \; \;\; \text{if} \;\; \RCompHidNodeBAF{1}{i} >0\,, \\
        & - \frac{\sigma_w}{\sqrt{2 \pi}} \; \;\;\; \;\; \; \;\; \; \;\;  \text{if} \;\; \RCompHidNodeBAF{1}{i} <0\,,
    \end{cases}
\end{align}
where $\frac{\sigma_w}{\sqrt{2 \pi}}$ is the mean value of ReLU, calculated from Eq.~\eqref{Pg_Relu}. SReLU, satisfying Eq.~\eqref{eq:null_mean_AF_cond}, falls into the subset of functions that do not exhibit IGB (see Fig.~\ref{fig:Comp_ReLU_SReLU}).

While the expectation value is easily computed and subtracted in this example, for a generic activation function, a similar shift could be based on empirical averages computed from the forwarding batch, akin to the approach in \citet{ioffe2015batch}.

\section{Amplification of the IGB level}\label{sec:ampl_IGB}
The previous sections focused on clarifying the conditions that ensure the emergence of IGB.
However, as highlighted by the comparison in Fig.~\ref{fig:Comp_SL_Pf0}, the presence of IGB is not uniform across all settings.
This section aims to clarify the conditions that can lead to an amplification of IGB.
To quantify the level of IGB, we will refer to the measure introduced in Eq.~\eqref{eq:VarRatio_def}, analyzing settings in which $\VarRatio$ diverges.
In the following sections, we will analyze different settings and limits leading to this divergence.

\subsection{Effect of data standardization}\label{sec:Stand_effects}

In this section, we discuss the relationship between dataset standardization and IGB. We begin by showing that this element alone is sufficient to induce the emergence of the phenomenon, irrespective of the architectural design choice (discussed in previous and subsequent sections). Specifically, we focus here on a particular standardization procedure, which involves simply recentering the input data by adding a constant, \(K \in \mathbb{R}\), to each component. To be more precise, the modification from the original setting consists of considering the pre-processed data

\begin{align}\label{eq:prep_input}
    \PreprDataOp{\InputValue_b^{(a)}} = \InputValue_b^{(a)} + K\,,
\end{align}
with \(\InputValue_b^{(a)} \sim \mathcal{N}(0,1)\). We then consider an architectural setting without max pooling and with a linear activation function. As seen in App.~\ref{sec:E_chi_O}, this architectural setting does not lead to the emergence of IGB using independent inputs centered at 0.
Repeating the same calculation using the preprocessed dataset, \(\PreprData\), according to Eq.~\eqref{eq:prep_input}
we can instead show the appearance of IGB. Indeed, starting from Eq.~\eqref{eq:lin_case_cond}
we have
\begin{align}\label{eq:rho_cen_stand}
    \Einput{\RCompHidNodeAMP{1}{1}{i}} = \Einput{\RCompHidNodeBAF{1}{i}{}} = \Einput{\sum_b ( \InputValue_b^{(a)} + K) \Weights{0}{ib}} = K \sum_b \Weights{0}{ib} ,
\end{align}
or 
\begin{align}\label{eq:rho_cent_dist_stand}
    \Einput{\RCompHidNodeAMP{1}{1}{i}} = \Einput{\RCompHidNodeBAF{1}{i}{}} \sim \NormalDistr{0}{K^2 \sigma_w^2}\,.
\end{align}

Proceeding in a manner similar to the derivation of Eq.~\eqref{eq:<h>^l+1}

\begin{align}
\begin{split}\label{eq:stand_cen_lin}
    &\pdf{\EinputComp{c}}{}{\Mean{}}  = \NormalDens{\Mean{}}{\sum_j \Eweights{\Weights{1}{ci}} \Einput{\RCompHidNodeBAF{1}{i}}}{\sum_j \CVar{\WeightSet{}}{\Weights{l}{ij}}  \Einput{\RCompHidNodeBAF{1}{i}}^2}  = \NormalDens{\Mean{}}{0}{\sigma_w^2 \frac{1}{\LayerNumNodes{1}} \sum_{j=1}^{\LayerNumNodes{1}} \Einput{\RCompHidNodeBAF{1}{i}}^2}\,,
\end{split}
\end{align}
which, combined with Eq.~\eqref{eq:rho_cent_dist_stand}, leads us to
\begin{align}\label{eq:dist_muc_lin_stand}
    \lim_{\LayerNumNodes{1} \rightarrow \infty} \pdf{\EinputComp{c}}{}{\Mean{}}  = \NormalDens{\Mean{}}{0}{\sigma_w^4 K^2 } \,\,.
\end{align}

Note that a similar result would have been obtained by using a generic non-zero vector for standardization, i.e., 
\begin{equation*}
    \PreprDataOp{\InputValue_b^{(a)}} = \InputValue_b^{(a)} + K_b.
\end{equation*}
The only difference would have been in Eq.~\eqref{eq:rho_cen_stand}, where the index dependence would have prevented the factorization of the generic term \( K_b \) outside the summation. This would have resulted in a linear combination of Gaussian variables (instead of a simple scaled sum of Gaussian variables), which similarly still follows a Gaussian distribution. Finally, we would have obtained a result analogous to Eq.~\eqref{eq:rho_cent_dist_stand}, with \( K^2 \equiv |\boldsymbol{K}|^2 = \sum_b k_b^2 \).

Therefore, the distribution of the output node centers is non-degenerate, thus indicating the presence of IGB.

After demonstrating the emergence of IGB, we show how an increase in \( |K| \) leads to the amplification of IGB. To measure the level of IGB, we consider the measure \( \VarRatio \) defined in Eq.~\eqref{eq:VarRatio_def}. In particular, proceeding in a manner similar to Eq.~\eqref{eq:Ph^l_W}, it is straightforward to show that:
\begin{equation}\label{eq:Oc_W}
\pdf{\ROutNode{c}{}}{(\Dataset)}{\OutNode{}{}} = \NormalDens{\OutNode{}{}}{\EinputComp{c}}{\sum_j \left( \Weights{1}{ij} \right)^2 \CVar{\Dataset}{\RCompHidNodeBAF{1}{j}}} \,.
\end{equation}
From this, it follows that 
\begin{align}\label{eq:VarO_stand}
    \lim_{d, \LayerNumNodes{1} \rightarrow \infty} \CVar{\Dataset}{\ROutNode{c}{}} = \sigma_w^4\,.
\end{align}

Finally, combining Eq.~\eqref{eq:stand_cen_lin} with Eq.~\eqref{eq:VarO_stand}, we have
\begin{align}
    \lim_{|K| \rightarrow \infty} \VarRatio = \lim_{|K| \rightarrow \infty} \frac{K^2 \sigma_w^4}{\sigma_w^4} = \infty\,.
\end{align}

Thus, the ratio \( \frac{\CVar{\WeightSet{}}{\EinputComp{c}}}{\CVar{\Dataset}{\ROutNode{c}{}}} \) diverges as \( |K| \) increases, indicating a significant amplification of IGB.

Building on this result, we can prove one of the claims of Thm.~\ref{thm:amp_IGB} (Eq.~\eqref{eq:thm_inf_stand_ampl}), which is reformulated more precisely as follows:
\begin{theorembox}
Consider a dataset with Gaussian-distributed, \textit{i.i.d.} components, where $\InputValue_b^{(a)} \sim \mathcal{N}(0,1)$, and $\InputValue_b^{(a)}$ represents the $b$-th component of the $a$-th input vector. This dataset is processed through an MLP with a single hidden layer ($L=1$). The MLP maps the preprocessed input, shifted by a constant vector $\boldsymbol{K}$, i.e., $\PreprDataOp{\InputValue_b^{(a)}} = \InputValue_b^{(a)} + K_b$, to the output as per Eq.~\eqref{eq:h_prop_def}, Eq.~\eqref{eq:g_prop_def}, and Eq.~\eqref{eq:rho_prop_def}. The weights are initialized following the Kaiming normal scheme, specifically $\Weights{l}{ij} \sim \mathcal{N} \left(0,\frac{\sigma_w^2}{\LayerNumNodes{l-1}}\right)$, with zero biases. Focusing on a setting without a pooling layer and employing ReLU as the activation function, an increase in the norm of $\boldsymbol{K}$ leads to an asymptotic amplification of the Initial Guessing Bias (IGB). Formally:
\begin{equation}
    \lim_{|\boldsymbol{K}| \rightarrow \infty} \VarRatio = \infty.
\end{equation}
\end{theorembox}
\begin{proof}
Following a similar approach as outlined in App.~\ref{sec:RescGaussVar}, particularly considering Eq.~\eqref{eq:<h>^l+1} and Eq.~\eqref{eq:Var_hl_cen_conv}, in our specific case (\textit{i.e.}, with $\RCompHidNodeBAF{l+1}{c} = \ROutNode{c}{}$), we obtain:
    \begin{align}\label{eq:Var_muc_ReLU}
      \lim_{\LayerNumNodes{1} \rightarrow \infty}  \CVar{\WeightSet{}}{\EinputComp{c}} = \sigma_w^2 \int_{-\infty}^{\infty} \Einput{\RCompHidNodeAAF{l}{j}}^2  \pdf{\Einput{\RCompHidNodeBAF{1}{j}{}}}{}{\Mean{\ArgHidNodeBAF{}{}}} d\Mean{\ArgHidNodeBAF{}{}}\,,
    \end{align}
where the integral on the r.h.s. represents the expectation of $\Einput{\RCompHidNodeAAF{l}{j}}^2$ over the population $\left\{ \Einput{\RCompHidNodeAAF{1}{j}} \right\}$; specifically, as $\Einput{\RCompHidNodeAAF{l}{j}}$ is a function of $\Einput{\RCompHidNodeBAF{1}{i}{}}$, we can express the mean in terms of the measure of the latter. 
Starting from Eq.~\eqref{eq:Var_muc_ReLU}, we have
    \begin{align}\label{eq:Varmu_mag}
         \lim_{\LayerNumNodes{1} \rightarrow \infty}  \CVar{\WeightSet{}}{\EinputComp{c}} \stackrel{\boldsymbol{a}}{>}  ~~&\sigma_w^2 \int_{0}^{\infty} \Einput{\RCompHidNodeAAF{l}{j}}^2  \pdf{\Einput{\RCompHidNodeBAF{1}{j}{}}}{}{\Mean{\ArgHidNodeBAF{}{}}} d \Mean{\ArgHidNodeBAF{}{}} \stackrel{\boldsymbol{b}}{>} \sigma_w^2 \int_{0}^{\infty} \Einput{\RCompHidNodeBAF{1}{j}{}}^2  \pdf{\Einput{\RCompHidNodeBAF{1}{j}{}}}{}{\Mean{\ArgHidNodeBAF{}{}}} d\Mean{\ArgHidNodeBAF{}{}} ~~\stackrel{\boldsymbol{c}}{=}~~  \\[2ex]
         =~~& \frac{\sigma_w^2}{2}  \int_{- \infty}^{\infty} \Einput{\RCompHidNodeBAF{1}{j}{}}^2  \pdf{\Einput{\RCompHidNodeBAF{1}{j}{}}}{}{\Mean{\ArgHidNodeBAF{}{}}} d \Mean{\ArgHidNodeBAF{}{}} ~~\stackrel{\boldsymbol{d}}{=}~~ \frac{\sigma_w^4}{2} |\boldsymbol{K}|^2\,,
    \end{align}
    where inequality $\boldsymbol{a}$ follows from the positivity of the integral in Eq.~\eqref{eq:Var_muc_ReLU}, while inequality $\boldsymbol{b}$ derives from the fact that the average value of a rectified Gaussian is, by construction, greater than the mean value of the original Gaussian; equality $\boldsymbol{c}$ follows from the parity of the integrand, and finally, equality $\boldsymbol{d}$ is deduced from the analysis on the linear system (particularly Eq.~\eqref{eq:dist_muc_lin_stand}).\newline
    Since, by construction, the variance of a rectified Gaussian is always less than that of the original Gaussian, starting from Eq.~\eqref{eq:Ph^l_W}, applying expectation over the population of activated hidden layer nodes (as just done) and using the result derived for the linear architecture (Eq.~\eqref{eq:VarO_stand}), we have:
    \begin{align}\label{eq:Var_PO_min_stand}
        \lim_{d, \LayerNumNodes{1} \rightarrow \infty} \CVar{\Dataset}{\ROutNode{c}{}} < \sigma_w^4\,.
    \end{align}
    Finally, combining Eq.~\eqref{eq:Varmu_mag} and Eq.~\eqref{eq:Var_PO_min_stand}, it follows that
    \begin{align}
        \lim_{|\boldsymbol{K}| \rightarrow \infty} \VarRatio = \infty\,.
    \end{align}
\end{proof}

\subsection{Effect of max pooling}\label{sec:MP_effects}
\begin{theorembox}\label{thm:mp_app}
Consider a dataset with Gaussian-distributed, \textit{i.i.d.} components, where $\InputValue_b^{(a)} \sim \mathcal{N}(0,1)$, and $\InputValue_b^{(a)}$ represents the $b$-th component of the $a$-th input vector. Inputs are processed through an MLP with a single hidden layer ($L=1$), mapping the input to output according to Eq.~\eqref{eq:h_prop_def}, Eq.~\eqref{eq:g_prop_def}, and Eq.~\eqref{eq:rho_prop_def}. The weights are initialized following the Kaiming normal scheme, i.e., $\Weights{l}{ij} \sim \mathcal{N} \left(0,\frac{\sigma_w^2}{\LayerNumNodes{l-1}}\right)$, with zero bias weights. In a setting with max pooling layer and ReLU activation function, an increase in the kernel size, $k$, leads to an asymptotic amplification of the Initial Guessing Bias (IGB). Formally:
\begin{equation}
    \lim_{m \rightarrow \infty} \VarRatio = \infty.
\end{equation}
\end{theorembox}

\begin{proof}
Consider the set $\{ \RCompHidNodeAAF{1}{i} \}_{i=1}^{k}$ such that
\begin{equation}
   \pdf{\RCompHidNodeAAF{1}{i}}{(\Dataset)}{\ArgHidNodeAAF{}{}}  =  \frac{1}{2} \Dirac{\ArgHidNodeAAF{}{}} + \Heavyside{\ArgHidNodeAAF{}{}} \NormalDens{\ArgHidNodeAAF{}{}}{0}{\sigma_{w}^2},
\end{equation}
for a fixed set $\WeightSet{}$. Now, define the set of random variables $\{  \omega_i \}_{i=1}^{k}$, where $n = \log (k)$, as follows:
\begin{equation} \label{eq:omega_def}
    \omega_i \equiv \frac{\RCompHidNodeAAF{1}{i}}{\sqrt{\log (k)}} = \frac{\RCompHidNodeAAF{1}{i}}{\sqrt{n}}.
\end{equation}
Utilizing the relation 
\begin{equation}
   \pdf{\omega_i}{(\Dataset)}{\omega}  d\omega = \pdf{\RCompHidNodeAAF{1}{}}{(\Dataset)}{\ArgHidNodeAAF{}{}(\omega)} d \ArgHidNodeAAF{}{},
\end{equation}
we can express
\begin{equation}
   \pdf{\omega_i}{(\Dataset)}{\omega}  = \frac{1}{2} \Dirac{\omega} + \Heavyside{\omega} \NormalDens{\omega}{0}{\frac{\sigma_{w}^2}{n}}.
\end{equation}

Now consider the number of elements in the set $\{ \omega_i \}$ that fall in the interval $[\tilde \omega, ( \tilde \omega + d\tilde \omega)], \; \tilde \omega >0$, denoted as $\#(\tilde \omega)$. Focusing on its expected value, we obtain
\begin{equation}\label{eq:occ_w}
    \CE{}{\#(\tilde \omega)} = m \CE{}{\mathbb{I}(\omega \in [\tilde \omega, ( \tilde \omega +d \tilde \omega)])} \sim m \sqrt{\frac{n}{2 \pi \sigma_{w}^2}} e^{-\frac{\omega^2 n}{2 \sigma_{w}^2 }} d \tilde \omega = \sqrt{\frac{n}{2 \pi \sigma_{w}^2}} e^{- n \left( \frac{\omega^2}{2 \sigma_{w}^2 } -1 \right)} d \tilde \omega.
\end{equation}

Furthermore, since the probability of obtaining a value from within the interval $[\tilde \omega, ( \tilde \omega +d \tilde \omega)]$, $\pdf{\omega}{(\Dataset)}{\tilde \omega} d\tilde \omega$, is small, it follows a Poisson law, implying that the variance is equal to the mean. Defining 
\begin{equation}\label{eq:s_def}
    s(\omega) \equiv \frac{\omega^2}{2 \sigma_{w}^2 } -1,
\end{equation}
we observe that:
\begin{itemize}
    \item If $s(\omega) <0$, the average number of draws with value $\omega$ is exponentially close to 0. From \eqref{eq:s_def}, this occurs (considering the positive interval) when $\omega > \sqrt{2 \sigma_{w}^2} $. In this case, applying the Markov inequality, we can conclude that, \emph{w.h.p.} the actual number of draws is effectively 0. Specifically, according to Markov inequality,
    \begin{equation}
        \Prob{}{\#(\tilde \omega)\geq 1} \leq \CE{}{\#(\tilde \omega)}.
    \end{equation}
    Thus, \emph{w.h.p.} as $n \rightarrow \infty$, all draws have values less than $\sqrt{2 \sigma_{w}^2}$.

    \item If $s(\omega) >0$, the expectation value of the number of draws is nonzero. Using the Markov inequality again, we show that, also in this scenario, the actual number of draws concentrates around the non-null expectation value as $n \rightarrow \infty$. Specifically,
    \begin{align}
        \begin{split}
            \Prob{}{\left| \frac{\# (\tilde \omega)}{\CE{}{\#(\tilde \omega)}} -1\right| \geq k} &=  \Prob{}{\left( \frac{\# (\tilde \omega)}{\CE{}{\#(\tilde \omega)}} -1\right)^2 \geq k^2} \leq \frac{\CE{}{\left(\# (\tilde \omega) - \CE{}{\#(\tilde \omega)}\right)^2} }{k^2 \left( \CE{}{\#(\tilde \omega)} \right)^2}\\[2ex]
            & \leq \frac{\CVar{}{\#(\tilde \omega)}}{k^2 \left( \CE{}{\#(\tilde \omega)} \right)^2} \simeq \frac{\cancel{ \CE{}{\#(\tilde \omega)}}}{k^2 \left( \CE{}{\#(\tilde \omega)} \right)^{\cancel{2}}} \propto \frac{e^{-n s(\tilde \omega)}}{k^2},
        \end{split}
    \end{align}
    where we used the property of the Poisson distribution, where the variance equals the mean. As $n$ increases, the probability of deviation from the mean exponentially decreases when $s(\tilde \omega) > 0$, so that \emph{w.h.p.}, $\frac{\# (\tilde \omega)}{\CE{}{\#(\tilde \omega)}}$ is arbitrarily close to 1.
\end{itemize}

More refined analyses are possible to characterize the extreme value statistics. For further information on this topic, refer to \citet{schehr2014exact} and the references therein. However, for the purposes of our demonstration, we have sufficient elements to proceed.

We have shown that, as $k \to \infty$, $\max\{\omega_i\}$ converges to a deterministic value, implying that
\begin{align}
    \lim_{k \to \infty} \frac{\sqrt{\CVar{}{\max\{\omega_i\}}}}{\CE{}{\max\{\omega\}}} = 0.
\end{align}
It is important to note that this condition remains invariant under any scaling factor over the random variable $\omega_i$, as both the standard deviation and the mean have identical dimensions. In particular, from Eq.~\eqref{eq:omega_def}, it follows:
\begin{align}\label{eq:lim_m_VarE_rho_ratio}
    \lim_{k \to \infty} \frac{\sqrt{\CVar{\Dataset}{\RCompHidNodeAMP{1}{k}{}}}}{\Einput{\RCompHidNodeAMP{1}{k}{}}} = \frac{\cancel{\log (k)} \sqrt{\CVar{}{\max\{\omega\}}}}{\cancel{\log (k)} \CE{}{\max\{\omega\}}} = 0.
\end{align}
Combining Eqs.~\eqref{eq:VarO_rho} and \eqref{eq:P<O>_teo_IGB_cond_ext} with Eq.~\eqref{eq:lim_m_VarE_rho_ratio}, we obtain

\begin{align}
     \lim_{k \rightarrow \infty} \VarRatio = \frac{\Einput{\RCompHidNodeAMP{1}{k}{}}^2}{\CVar{\Dataset}{\RCompHidNodeAMP{1}{k}{}}} = \infty.
\end{align}

\end{proof}

In the proof of Thm.~\ref{thm:mp_app}, we focused on the statistical behavior of the maximum value in a growing set of random variables (r.v.s), particularly emphasizing positive event occurrences. This analysis remains applicable if we substitute the ReLU activation function with a linear one, as ReLU does not change the distribution within the positive domain of its support. This implies that the amplification of Initial Guessing Bias (IGB) in models with max pooling is independent of the choice of activation function, manifesting even without any activation function. This finding is not surprising, given that App.~\ref{app:proof_out_dist} demonstrates that the key factor in the emergence of IGB is the statistics of nodes post-pooling layer, in particular $\Einput{\RCompHidNodeAMP{1}{k}{}}$. Hence, when considering models with pooling layers, it's crucial to evaluate the joint impact of the activation function and the pooling layer to accurately assess the presence and intensity of IGB.

\subsection{Deep architectures}\label{sec:deep_arc}
In App.~\ref{sec:SLP_analysis}, we discussed how to derive $\pdf{\RClassFraction{0}}{}{\ClassFraction{0}}$ for a neural network with a single hidden layer. In this section, we will discuss the extension of the computation to a network with an arbitrary number of hidden layers. After that, we will focus on the impact that depth has on IGB. In particular, in App.~\ref{sec:proof_conv_deep_f0} we will show that network depth does not cause IGB, but it amplifies it when the network's depth increases.

\subsubsection{Multi-layer perceptron}\label{sec:MLP}
The strategy that we will follow is similar to the one presented for the single-hidden-layer counterpart (see App.~\ref{sec:SLP_analysis}). We will discuss here the case of MLP with ReLU activation function to focus on the effects of network depth. Extending the results discussed in the single-layer case to deeper networks, including variations such as the inclusion of Max-Pooling, is straightforward.  We will propagate the signal across the network layers, keeping track of the changes in the distributions $\pdf{\RCompHidNodeBAF{l+1}{i}}{(\Dataset)}{\ArgHidNodeBAF{}{}}$ and $ \pdf{ \Einput{\RCompHidNodeBAF{l+1}{i}}}{}{\Mean{}}$. Using the CLT and its extensions considerably simplifies this propagation process because to keep track of changes in the distribution it is sufficient to see how few quantities vary.\footnote{This is because a Gaussian distribution is completely determined by the first two cumulants.}
Compared to the analysis in App.~\ref{sec:SLP_analysis}, however, there is an important complicating element. While the elements in the set $\{ \RCompHidNodeBAF{1}{i} \}$ follow the same distribution (regardless of the choice of the activation function), from the second layer onward the activation function can induce a breaking in symmetry among the layer nodes, in the sense that they will follow, fixed a given configuration for the network weights, different distributions. This symmetry breaking, as we shall see, can cause an accentuation of IGB.
To show this point, we will consider in our analysis the ReLU activation function.\footnote{We do not include the max pooling to isolate the effect of depth from that of max pooling. Note that absence of max pooling is equivalent of a max pooling with minimum kernel size.}
Starting from the same setting described in App.~\ref{sec:SLP_analysis}, we will derive a set of iterative equations to propagate across multiple layers.
\begin{itemize}
\item \textbf{layer-1} \newline
As shown in App.~\ref{sec:SLP_analysis} and in App.~\ref{sec:E_chi_O}, in the limit $d \rightarrow \infty$,
\begin{align}
\begin{split}
& \RCompHidNodeBAF{1}{i} \equiv \sum_j \Weights{0}{ij} \InputValue_j \xRightarrow{\text{CLT}} \pdf{\RCompHidNodeBAF{1}{i}}{(\Dataset)}{\ArgHidNodeBAF{}{}}  \xrightarrow{d \rightarrow \infty}  \mathcal{N}(0,\sigma_w^2)  \xRightarrow{\text{ReLU}} \pdf{\RCompHidNodeAAF{1}{i}}{(\Dataset)}{\ArgHidNodeAAF{}{}}  \xrightarrow{d \rightarrow \infty}  \Heavyside{\ArgHidNodeAAF{}{}} \NormalDens{\ArgHidNodeAAF{}{}}{0}{\sigma_w^2} + \frac{1}{2}\Dirac{\ArgHidNodeAAF{}{}} \,.
\end{split}
\end{align}
All the nodes $\{ \RCompHidNodeBAF{1}{i} \}$ follow the same distribution, and so do the nodes $\{ \RCompHidNodeAAF{1}{i} \}$ . In particular, this means that
\begin{equation}
\lim_{\DatasetSize \to \infty} \Einput{\RCompHidNodeAAF{1}{i}} = \Einput{\RCompHidNodeAAF{1}{}} , \; \forall i\,.
\end{equation}
\item \textbf{layer-2} \newline
We start again from a combination of \textit{r.v.}s
\begin{equation}
\RCompHidNodeBAF{2}{i} \equiv \sum_j \Weights{1}{ij} \RCompHidNodeAAF{1}{j}\,.
\end{equation} 
It is easy to prove that the generic \textit{r.v.} $\left( \Weights{1}{ij} \RCompHidNodeAAF{1}{j} \right)$ involved in the sum satisfies the conditions of Eq.~\eqref{eq:Var_conv_cond} and Eq.~\eqref{eq:fast_tail_dec_cond}. So, in the limit $ \LayerNumNodes{1} \rightarrow \infty$,  we again have a convergence to a normal distribution. In particular,
\begin{align}\label{eq:l2_h_conv}
\begin{split}
 &\pdf{\RCompHidNodeBAF{2}{i}}{(\Dataset)}{\ArgHidNodeBAF{}{}}
  \xrightarrow{\LayerNumNodes{1} \rightarrow \infty} \NormalDens{\ArgHidNodeBAF{}{}}{\Einput{\RCompHidNodeAAF{1}{}} S_{\boldsymbol{w}_i}}{\CVar{\Dataset}{\RCompHidNodeAAF{1}{}} S_{\boldsymbol{w}_i^2}}  = \NormalDens{\ArgHidNodeBAF{}{}}{\Einput{\RCompHidNodeAAF{1}{}} S_{\boldsymbol{w}_i}}{\CVar{\Dataset}{\RCompHidNodeAAF{1}{}} \Eweights{S_{\boldsymbol{w}_i^2}}} \, .
 \end{split}
\end{align}
In the last step of Eq.~\eqref{eq:l2_h_conv}, as done for $ \pdf{\ROutNode{c}{}}{(\Dataset)}{\OutNode{}{}}$ in Sec.~\ref{sec:SLP_analysis}, we used the concentration result derived in App.~\ref{sec:RescGaussVar}. In particular that the distribution of the \textit{r.v.} $S_{\boldsymbol{w}_i} \equiv \sum_{j=1}^{\LayerNumNodes{1}} \Weights{1}{ij}$ stays stable in the limit $ \LayerNumNodes{1} \rightarrow \infty$,\footnote{Fluctuations and mean stay both $\mathcal{O} \left( 1 \right)$.} while the distribution of $S_{\boldsymbol{w}_i^2} \equiv \sum_{j=1}^{\LayerNumNodes{1}} \left( \Weights{1}{ij} \right)^2 $ asymptotically narrows around the mean value.
Note that, as the distribution of $S_{\boldsymbol{w}_i}$ does not asymptotically concentrate, and since  $\Einput{\RCompHidNodeAAF{1}{}} \neq 0$, each node in the set $\{ \RCompHidNodeBAF{2}{i} \}$ will follow a different distribution. In particular we will have a set of normally distributed \textit{r.v.}s, centred on different random points, $\{ \Einput{\RCompHidNodeAAF{1}{}} S_{\boldsymbol{w}_i} \}_i$. 
After passing through the ReLU activation function, we will have
\begin{align}
\begin{split}
 & \pdf{\RCompHidNodeAAF{2}{i}}{(\Dataset)}{\ArgHidNodeAAF{}{}} \xrightarrow{\LayerNumNodes{1} \rightarrow \infty}
 \Heavyside{\ArgHidNodeAAF{}{}}   \NormalDens{\ArgHidNodeAAF{}{}}{\Einput{\RCompHidNodeAAF{1}{}} S_{\boldsymbol{w}_i}}{\CVar{\Dataset}{\RCompHidNodeAAF{1}{}} \Eweights{S_{\boldsymbol{w}_i^2}} }   + \Bigg( \frac{1}{2} + \frac{1}{2}  \erf\Bigg( \frac{\Einput{\RCompHidNodeAAF{1}{}} S_{\boldsymbol{w}_i}}{\sqrt{2 \CVar{\Dataset}{\RCompHidNodeAAF{1}{}} \Eweights{S_{\boldsymbol{w}_i^2}}}} \Bigg) \Bigg) \Dirac{\ArgHidNodeAAF{}{}}\,.
\end{split}
\end{align}

\item \textbf{layer-3} \newline
We can repeat the approach of the previous layer for the new set of variables, 
\begin{equation}
\RCompHidNodeBAF{3}{i} \equiv \sum_j \Weights{2}{ij} \RCompHidNodeAAF{2}{j}\,,
\end{equation}
getting 
\begin{equation}
\pdf{\RCompHidNodeBAF{3}{i}}{(\Dataset)}{\ArgHidNodeBAF{}{}}\xrightarrow{\LayerNumNodes{2} \rightarrow \infty} \NormalDens{\ArgHidNodeBAF{}{}}{\sum_{j=1}^{\LayerNumNodes{2}} \Weights{2}{ij} \Einput{\RCompHidNodeAAF{2}{j}}}{\sum_{j=1}^{\LayerNumNodes{2}}  \left( \Weights{2}{ij} \right)^2  \CVar{\Dataset}{\RCompHidNodeAAF{2}{j}}} \,.
\end{equation}
Note that in this case we cannot take out from the sums $\Einput{\RCompHidNodeAAF{2}{j}}$ and $\CVar{\Dataset}{\RCompHidNodeAAF{2}{j}}$
since the nodes $\{ \RCompHidNodeAAF{2}{j} \}$ are not identically distributed. In App.~\ref{sec:LayerConcExample} we analyze the differences in $ \pdf{\RCompHidNodeBAF{3}{i}}{(\Dataset)}{\ArgHidNodeBAF{}{}}$ between the different nodes $\{ \RCompHidNodeBAF{3}{i} \}$. In particular, we show that

\begin{align}
& \pdf{\RCompHidNodeBAF{3}{i}}{(\Dataset)}{\ArgHidNodeBAF{}{}} \xrightarrow{\LayerNumNodes{2} \rightarrow \infty} \NormalDens{\ArgHidNodeBAF{}{}}{\Einput{\RCompHidNodeBAF{3}{i}}}{\CVar{\Dataset}{\RCompHidNodeBAF{3}{i}}}
\,,\\
& \pdf{\Einput{\RCompHidNodeBAF{3}{i}}}{}{\Mean{\ArgHidNodeBAF{}{}}} \xrightarrow{\LayerNumNodes{2} \rightarrow \infty} \NormalDens{\Mean{\ArgHidNodeBAF{}{}}}{0}{\sigma^2_w  \Eweights{\Einput{\RCompHidNodeAAF{2}{j}}^2}} \,,\\
& \pdf{\CVar{\Dataset}{\RCompHidNodeBAF{3}{i}}}{}{v}  \xrightarrow{\LayerNumNodes{2} \rightarrow \infty} \Dirac{v -\sigma_w^2 \Eweights{\CVar{\Dataset}{\RCompHidNodeAAF{2}{j}}}}\,.
\end{align}

\item \textbf{layer-}$\boldsymbol{l}$ \newline
The steps described for the propagation of the \emph{layer-3} provide us with an iteration scheme that we can follow for a generic  layer $l\geq3$. In fact, knowing the statistics of the previous layers, we can compute
\begin{align}
&\Eweights{\CVar{\Dataset}{\RCompHidNodeAAF{l-1}{j}}} = \int_{\mathbb{R}} \CVar{\Dataset}{\RCompHidNodeAAF{l-1}{j}} (\Mean{\ArgHidNodeBAF{}{}}) \; \; \NormalDens{\Mean{\ArgHidNodeBAF{}{}}}{0}{\sigma_w^2 \Eweights{ \Einput{\RCompHidNodeAAF{l-2}{j}}^2}}
   d \Mean{\ArgHidNodeBAF{}{}} \label{eq:mean_sigma_l-1}\,,\\
& \Eweights{ \Einput{\RCompHidNodeAAF{l-1}{j}}^2} = \int_{\mathbb{R}} \Einput{\RCompHidNodeAAF{l-1}{j}}^2 (\Mean{\ArgHidNodeBAF{}{}}) \; \; \NormalDens{\Mean{\ArgHidNodeBAF{}{}}}{0}{\sigma_w^2 \Eweights{ \Einput{\RCompHidNodeAAF{l-2}{j}}^2}} d \Mean{\ArgHidNodeBAF{}{}}. \label{eq:mean_mu2_l-1}
\end{align}

From Eq.~\eqref{eq:mean_sigma_l-1} and Eq.~\eqref{eq:mean_mu2_l-1} we can compute then 
\begin{align}
& \pdf{\RCompHidNodeBAF{l}{i}}{(\Dataset)}{\ArgHidNodeBAF{}{}}  \xrightarrow{\LayerNumNodes{l-1} \rightarrow \infty} \NormalDens{\ArgHidNodeBAF{}{}}{\Einput{\RCompHidNodeBAF{l}{i}}}{\sigma_w^2 \Eweights{\CVar{\Dataset}{\RCompHidNodeAAF{l-1}{j}}}} \,,
\label{eq:P_h_l}\\
& \pdf{\Einput{\RCompHidNodeBAF{l}{i}}}{}{\ArgHidNodeBAF{}{}}  \xrightarrow{\LayerNumNodes{l-1} \rightarrow \infty} \NormalDens{\ArgHidNodeBAF{}{}}{0}{\sigma^2_w  \Eweights{ \Einput{\RCompHidNodeAAF{l-1}{j}}^2}} \,.\label{eq:P_mu_l}
\end{align}
Finally we can use \eqref{eq:P_mu_l} to iterate the set of \eqref{eq:mean_sigma_l-1} and \eqref{eq:mean_mu2_l-1} for the layer $l$, \emph{i.e.}
\begin{align}
&\Eweights{\CVar{\Dataset}{\RCompHidNodeAAF{l}{j}}} =  \int_{\mathbb{R}} \CVar{\Dataset}{\RCompHidNodeAAF{l}{j}} (\Mean{\ArgHidNodeBAF{}{}}) \;\;  \pdf{\Einput{\RCompHidNodeBAF{l}{j}}}{}{\Mean{\ArgHidNodeBAF{}{}}} d \Mean{\ArgHidNodeBAF{}{}} \label{eq:mean_sigma_l}\,,\\
& \Eweights{ \Einput{\RCompHidNodeAAF{l}{j}}^2} =
 \int_{\mathbb{R}} \Einput{\RCompHidNodeAAF{l}{j}}^2 (\Mean{\ArgHidNodeBAF{}{}})  \;\; \pdf{\Einput{\RCompHidNodeBAF{l}{j}}}{}{\Mean{\ArgHidNodeBAF{}{}}}  d \Mean{\ArgHidNodeBAF{}{}} \,. \label{eq:mean_mu2_l}
\end{align}
\item \textbf{layer-}$\boldsymbol{(L+1)}$ \newline
Arriving at the output layer, following the same iterative scheme we will have 

\begin{align}
\begin{cases}
& \pdf{\ROutNode{c}{}}{(\Dataset)}{\OutNode{}{}} \xrightarrow{\LayerNumNodes{L} \rightarrow \infty} \NormalDens{\OutNode{}{}}{\Einput{\ROutNode{c}{}}}{\sigma_w^2 \Eweights{\CVar{\Dataset}{\RCompHidNodeAAF{L}{j}}}}
\\
& \pdf{\EinputComp{c}}{(\Dataset)}{\Mean{}}  \xrightarrow{\LayerNumNodes{L} \rightarrow \infty} \NormalDens{\Mean{}}{0}{\sigma^2_w  \Eweights{ \Einput{\RCompHidNodeAAF{L}{j}}^2}} \label{eq:P_O}
\end{cases}\,.
\end{align}
These two distributions are the only ingredient that we need to replicate the steps described at the end of App.~\ref{sec:SLP_analysis} to get $\pdf{\RClassFraction{0}}{}{\ClassFraction{0}}$.
\end{itemize}

\begin{figure}[h]
    \centering
    \includegraphics[width=.7\textwidth]{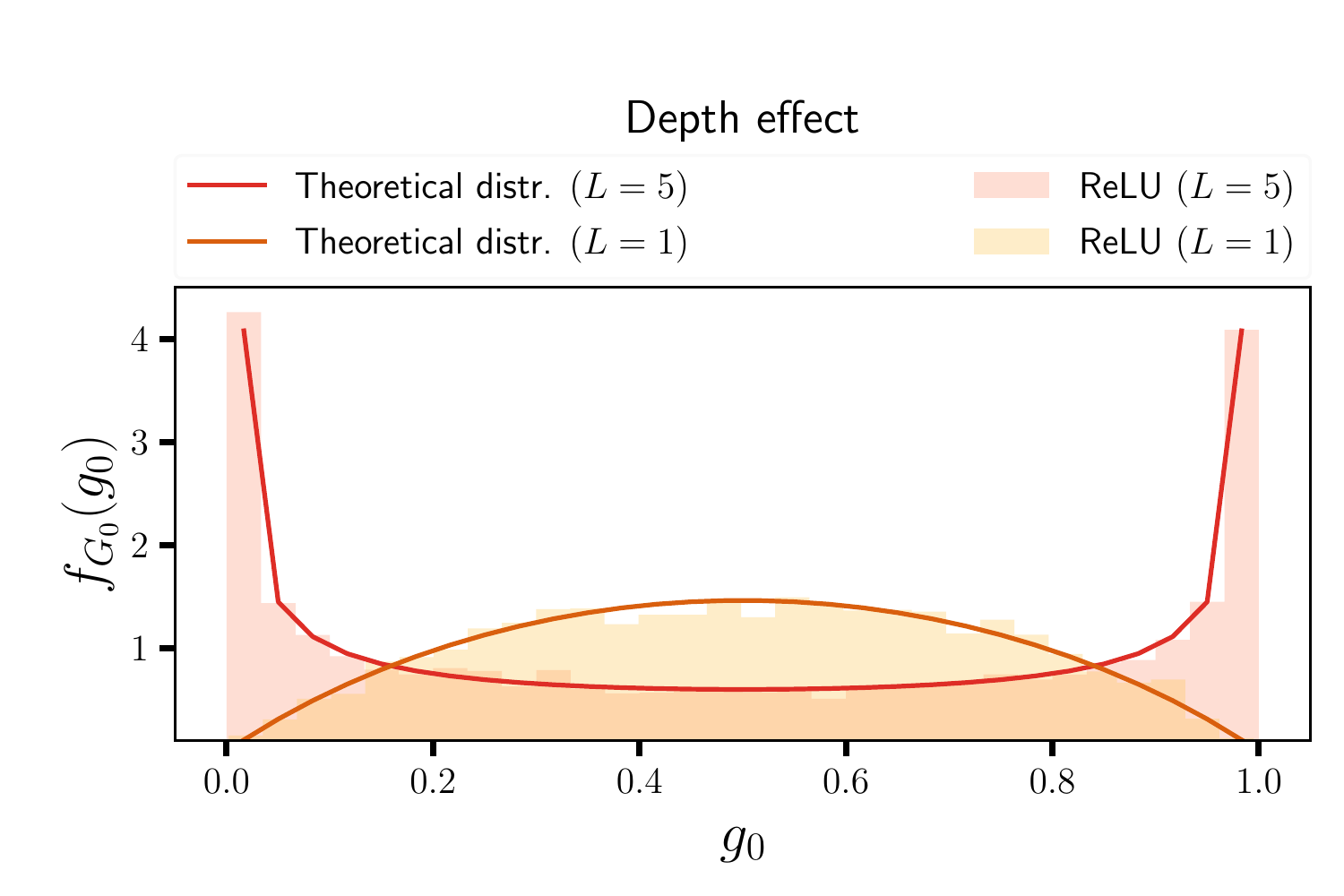}
    \caption{Effects induced by the depth of the network. Simulations on an MLP with $L$ hidden layers and ReLU activation function (without any pooling layers) are compared for two different depth values. The theoretical curves are superimposed on the empirical distributions, demonstrating a good agreement between the predictions of our analysis and empirical observations. For a single hidden layer network $(L=1)$ we have a stable distribution (\emph{i.e.} that does not narrow around $\Eweights{\RMultiClassFraction{0}{}} = 0.5$ for $\DatasetSize \rightarrow \infty$) but still peaked on $\Eweights{\RMultiClassFraction{0}{}} = 0.5$. Increasing the number of hidden layers the probability mass moves from the center to the extremes of the support. For these simulations we used the \GB~dataset on model \MLPB~(see App.~\ref{sec:reprod} for more details).} 
    \label{fig:ReLU_deepness_effect}
\end{figure}

\subsubsection{Inclusion of Non-Null Biases}\label{App:Bias_ext}
In the analysis presented in App.~\ref{sec:MLP}, we considered a setting with null biases for the sake of clarity. However, the analysis can be extended to initializations where the biases are set to non-null values. For instance, considering an MLP with Gaussian-initialized bias parameters, the pre-activations of a generic hidden layer can be described as:
\begin{equation}
H'_i = H_i + B_i\,.
\end{equation}
Here, $H_i$ represents node $i$'s pre-activation in the absence of biases, which, as per our analysis, follows a normal distribution.\footnote{We omitted the layer index $l$ on the variables $H_i, \, B_i , \, H'_i $ to lighten the notation.} The term $B_i$ denotes the bias parameter. Given that $H'_i$ is a sum of two independent Gaussian random variables, it too will exhibit a normal distribution. This allows for a seamless extension of our analysis. Similarly, if the bias parameters are initialized with a constant value ($B_i = c$), the analysis can be adapted accordingly. Adding a constant to $H_i$ merely shifts the distribution's center, preserving the Gaussian profile for $H'_i$ and reaffirming the robustness of our findings regarding IGB across different initialization strategies.

\subsubsection{Amplification of IGB with depth}\label{sec:proof_conv_deep_f0}
Here we prove that the distribution of $\pdf{\RClassFraction{0}}{}{\ClassFraction{0}}$ converges to a delta-peaked distribution in a multi-layer perceptron with ReLU activation when the number of layers goes to infinity. More precisely:

\begin{theorembox}[]
Consider a dataset with Gaussian-distributed, \textit{i.i.d.} components, where $\InputValue_b^{(a)} \sim \mathcal{N}(0,1)$, and $\InputValue_b^{(a)}$ represents the $b$-th component of the $a$-th input vector. Inputs are processed through an MLP, mapping the input to output according to Eq.~\eqref{eq:h_prop_def}, Eq.~\eqref{eq:g_prop_def}, and Eq.~\eqref{eq:rho_prop_def}. The weights are initialized following the Kaiming normal scheme, i.e., $\Weights{l}{ij} \sim \mathcal{N} \left(0,\frac{\sigma_w^2}{\LayerNumNodes{l-1}}\right)$, with zero bias weights.
Let us consider a multi-layer perceptron with $L$ hidden layers, ReLU activation function, and without pooling layers. Let us assume that, in the $\lim_{\LayerNumNodes{l} \rightarrow \infty}\forall l \in [0, \dots, L]$, \emph{w.h.p.}:

\begin{align}\label{eq:non_deg_var_l}
 \CVar{\Dataset}{\RCompHidNodeBAF{l}{}} > 0, \; \forall l\,,
\end{align}
and in particular
\begin{align}\label{eq:non_deg_var_asym}
\lim_{l \rightarrow \infty} \CVar{\Dataset}{\RCompHidNodeBAF{l}{}} > 0,
\end{align}
where $\CVar{\Dataset}{\RCompHidNodeBAF{l}{}}$ indicate the variance of $\pdf{\RCompHidNodeBAF{l}{}}{(\Dataset)}{\ArgHidNodeBAF{}{}}$.
Then 
\begin{equation}\label{eq:Var_ratio_asymp}
  \lim_{L \rightarrow \infty}   \VarRatio = \infty \, .
\end{equation}

\label{thm:P_f0_conv}
\end{theorembox}

To prove Eq.~\eqref{eq:Var_ratio_asymp}, we will show that 
\begin{align}
& \CVar{\WeightSet{}}{\Einput{\RCompHidNodeBAF{l+1}{}}} > (K + \epsilon) \CVar{\WeightSet{}}{\Einput{\RCompHidNodeBAF{l}{}}}\,,
\label{eq:sigma_mh_LB}\\
 & \CVar{\Dataset}{\RCompHidNodeBAF{l+1}{}} < (K- \epsilon) \CVar{\Dataset}{\RCompHidNodeBAF{l}{}}\label{eq:sigma_h_LB}\,,
\end{align}
where $K$ is a constant.
Once proved the above relations we can, indeed, write 
\begin{align}
&\frac{\CVar{\WeightSet{}}{\Einput{\RCompHidNodeBAF{l+2}{}}}}{\CVar{\Dataset}{\RCompHidNodeBAF{l+2}{}} } > \left(\frac{1}{K}\right)^l \frac{\CVar{\WeightSet{}}{\Einput{\RCompHidNodeBAF{2}{}}}}{\CVar{\Dataset}{\RCompHidNodeBAF{2}{}}} \left( K +\epsilon \right)^l > l \underbrace{\left(
\frac{1}{K} \frac{\CVar{\WeightSet{}}{\Einput{\RCompHidNodeBAF{2}{}}}}{\CVar{\Dataset}{\RCompHidNodeBAF{2}{}}} \epsilon \right)}_{\equiv \frac{1}{C}}\,, \notag \\
&\Longrightarrow \frac{\CVar{\Dataset}{\RCompHidNodeBAF{l+2}{}}}{\CVar{\WeightSet{}}{\Einput{\RCompHidNodeBAF{l+2}{}}}} < C \frac{1}{l}.
\end{align}
Therefore if we consider an infinite depth network, \emph{i.e.} $L \rightarrow \infty$, we have:
\begin{equation}
\frac{\CVar{\Dataset}{\RCompHidNodeBAF{L}{}}}{\CVar{\WeightSet{}}{\Einput{\RCompHidNodeBAF{L}{}}}} = \mathcal{O} \left( \frac{1}{L} \right) \Longrightarrow \frac{\CVar{\WeightSet{}}{\EinputComp{c}}}{\CVar{\Dataset}{\ROutNode{c}{}}} = \mathcal{O} \left( L \right) \Longrightarrow \lim_{L \rightarrow \infty} \frac{\CVar{\WeightSet{}}{\EinputComp{c}}}{\CVar{\Dataset}{\ROutNode{c}{}}} = \infty \,.
\end{equation}

\begin{proof}
Let us start from Eq.~\eqref{eq:sigma_mh_LB}. As a first step we prove that, for $\LayerNumNodes{l-1} ~\rightarrow~ \infty$,
\begin{align}\label{eq:pre_ineq_deep_proof}
    \frac{1}{2} \CVar{\Dataset}{\RCompHidNodeBAF{l}{}} >  \Eweights{\CVar{\Dataset}{\RCompHidNodeAAF{l}{}}} &  \equiv \Eweights{\Einput{\left( \RCompHidNodeAAF{l}{} \right)^2}} - \Eweights{\Einput{\RCompHidNodeAAF{l}{}}^2}.
\end{align}
To prove this inequality we will now focus separately on the two terms of the \textit{r.h.s.}.
First we analyze $\Eweights{\Einput{\RCompHidNodeAAF{l}{}}^2}$.
From Eq.~\eqref{eq:P_h_l}
\begin{equation}
\CVar{\WeightSet{}}{\Einput{\RCompHidNodeBAF{l+1}{}}}= \sigma^2_w   \Eweights{ \Einput{\RCompHidNodeAAF{l}{j}}^2}.
\end{equation}
 
\begin{align}\label{eq:espr_sigma_mu_h}
\begin{split}
&\Eweights{ \Einput{\RCompHidNodeBAF{l}{j}}^2} =  \int_{-\infty}^{\infty} \pdf{\Einput{\RCompHidNodeBAF{l}{j}}}{}{\Mean{z}} \Mean{z}^2 \, d\Mean{z} =\int_{-\infty}^{0} \pdf{\Einput{\RCompHidNodeBAF{l}{j}}}{}{\Mean{z}} \Mean{z}^2 \, d\Mean{z} + \int_{0}^{\infty} \pdf{\Einput{\RCompHidNodeBAF{l}{j}}}{}{\Mean{z}} \Mean{z}^2 \, d\Mean{z} \,.
\end{split}
\end{align}
We now use the symmetry of $\pdf{\Einput{\RCompHidNodeBAF{l}{j}}}{}{\Mean{z}}$ (which is a Gaussian centered in 0), \emph{i.e.}
\begin{equation}\label{eq:dist_sym_cond}
\pdf{\Einput{\RCompHidNodeBAF{l}{j}}}{}{\Mean{z}} = \pdf{\Einput{\RCompHidNodeBAF{l}{j}}}{}{-\Mean{z}}\,,
\end{equation}

  to rewrite Eq.~\eqref{eq:espr_sigma_mu_h} as 

\begin{align}
\begin{split}
&\frac{1}{2} \Eweights{ \Einput{\RCompHidNodeBAF{l}{j}}^2} = \int_{0}^{\infty} \pdf{\Einput{\RCompHidNodeBAF{l}{j}}}{}{\Mean{z}} \Mean{z}^2 d\Mean{z}\equiv \int_{0}^{\infty} \pdf{\Einput{\RCompHidNodeBAF{l}{j}}}{}{\Mean{z}} \left( \mu_+(\Mean{z}) + \mu_-(\Mean{z})  \right)^2 d\Mean{z}\,,
\end{split}
\end{align}
where we defined 
\begin{align}\label{eq:mu_pm_def}
\begin{split}
& \mu_+ \left( \Mean{\ArgHidNodeBAF{}{}} \right) = \int_{0}^{\infty} \NormalDens{\ArgHidNodeBAF{}{}}{\Mean{\ArgHidNodeBAF{}{}}}{\CVar{\Dataset}{\RCompHidNodeBAF{l+1}{}}}  \ArgHidNodeBAF{}{} d\ArgHidNodeBAF{}{} >0\,,\\
& \mu_- \left(  \Mean{\ArgHidNodeBAF{}{}}\right) = \int_{- \infty}^{0} \NormalDens{\ArgHidNodeBAF{}{}}{\Mean{\ArgHidNodeBAF{}{}}}{\CVar{\Dataset}{\RCompHidNodeBAF{l+1}{}}} \ArgHidNodeBAF{}{} d\ArgHidNodeBAF{}{} <0\,.
\end{split}
\end{align}

We can thus rewrite 
\begin{align}
\frac{1}{2} \Eweights{ \Einput{\RCompHidNodeBAF{l}{j}}^2} =
\int_{0}^{\infty} \pdf{\Einput{\RCompHidNodeBAF{l}{j}}}{}{x} \big( \mu_+^2 + \mu_-^2 +2 \underbrace{\mu_+ \mu_-}_{\leq 0}  \big) dx.
\end{align}
Note that $\mu_+ \mu_- = 0$ implies either $\mu_+=0$ or $\mu_- = 0$ \emph{i.e.} a distribution $\pdf{\RCompHidNodeBAF{l}{j}}{(\Dataset)}{\ArgHidNodeBAF{}{}}$ with  non-positive or non-negative support. Since $\pdf{\RCompHidNodeBAF{l}{j}}{(\Dataset)}{\ArgHidNodeBAF{}{}}$ is a Gaussian distribution this may only happen in the limit of its variance going to 0 \emph{i.e.} if the Gaussian shrink into a Dirac delta distribution. By hypothesis we excluded this possibility (Eq.~\eqref{eq:non_deg_var_l} and Eq.~\eqref{eq:non_deg_var_asym}); therefore $\mu_+ \mu_- < 0$ and we can rewrite
\begin{align}\label{eq:mu2_lb}
\frac{1}{2} \Eweights{ \Einput{\RCompHidNodeBAF{l}{j}}^2} <
\underbrace{\int_{0}^{\infty} \pdf{\Einput{\RCompHidNodeBAF{l}{j}}}{}{\Mean{\ArgHidNodeBAF{}{}}} \left( \mu_+(\Mean{\ArgHidNodeBAF{}{}})^2 + \mu_-(\Mean{\ArgHidNodeBAF{}{}})^2  \right) d\Mean{\ArgHidNodeBAF{}{}}}_{\equiv I} \; .
\end{align}
Now let us focus on the integral $I$ of Eq.~\eqref{eq:mu2_lb}. We note that 
\begin{align}
-\mu_-(\Mean{\ArgHidNodeBAF{}{}}) &= -\int_{- \infty}^{0} \NormalDens{\ArgHidNodeBAF{}{}}{\Mean{\ArgHidNodeBAF{}{}}}{\CVar{\Dataset}{\RCompHidNodeBAF{l+1}{}}} \ArgHidNodeBAF{}{} d \ArgHidNodeBAF{}{} ==\int_{0}^{\infty} \NormalDens{y}{-\Mean{\ArgHidNodeBAF{}{}}}{\CVar{\Dataset}{\RCompHidNodeBAF{l+1}{}}} y dy = \mu_+(-\Mean{\ArgHidNodeBAF{}{}})\,,
\end{align}
where in the second step we just changed the integration variable $\ArgHidNodeBAF{}{} \rightarrow y\equiv- \ArgHidNodeBAF{}{}$.
We can therefore rewrite
\begin{align}
\begin{split}
    &I = \int_{0}^{\infty} \pdf{\Einput{\RCompHidNodeBAF{l}{j}}}{}{\Mean{\ArgHidNodeBAF{}{}}} \left( \mu_+(\Mean{\ArgHidNodeBAF{}{}})^2 + \mu_+(-\Mean{\ArgHidNodeBAF{}{}})^2  \right) d\Mean{\ArgHidNodeBAF{}{}} = \int_{-\infty}^{\infty} \pdf{\Einput{\RCompHidNodeBAF{l}{j}}}{}{\Mean{\ArgHidNodeBAF{}{}}} \mu_+(\Mean{\ArgHidNodeBAF{}{}})^2 d\Mean{\ArgHidNodeBAF{}{}} = \Eweights{\Einput{\RCompHidNodeAAF{l}{j}}^2}\,,
    \end{split}
\end{align}
where for the second step we used again Eq.~\eqref{eq:dist_sym_cond}. In summary, we showed that
\begin{equation}\label{eq:E_mu_ineq}
\Eweights{\Einput{\RCompHidNodeAAF{l}{j}}^2} > \left(\frac{1}{2} \right) \Eweights{\Einput{\RCompHidNodeBAF{l}{j}}^2}\,.
\end{equation}

We now turn our attention to the second term appearing in the \textit{r.h.s.} of Eq.~\eqref{eq:pre_ineq_deep_proof}, \textit{i.e.} $\Eweights{\Einput{\left( \RCompHidNodeAAF{l}{} \right)^2}}$:
\begin{align}\label{eq:E_x2_g}
\begin{split}
\Eweights{\Einput{\left( \RCompHidNodeAAF{l}{} \right)^2}} &= \int_{-\infty}^{\infty} \pdf{\Einput{\RCompHidNodeBAF{l}{j}}}{}{\Mean{\ArgHidNodeBAF{}{}}} \Einput{\left( \RCompHidNodeAAF{l}{} \right)^2}(\Mean{\ArgHidNodeBAF{}{}}) \; d\Mean{\ArgHidNodeBAF{}{}}  \\ &= \int_{0}^{\infty} \pdf{\Einput{\RCompHidNodeBAF{l}{j}}}{}{\Mean{\ArgHidNodeBAF{}{}}} \left( \Einput{\left( \RCompHidNodeAAF{l}{} \right)^2}(\Mean{\ArgHidNodeBAF{}{}}) + \Einput{\left( \RCompHidNodeAAF{l}{} \right)^2}(-\Mean{\ArgHidNodeBAF{}{}}) \right) \; d\Mean{\ArgHidNodeBAF{}{}}  = \\ & = \int_{0}^{\infty} \pdf{\Einput{\RCompHidNodeBAF{l}{j}}}{}{\Mean{\ArgHidNodeBAF{}{}}} \left( \Einput{\left( \RCompHidNodeAAF{l}{} \right)^2}_+(\Mean{\ArgHidNodeBAF{}{}}) + \Einput{\left( \RCompHidNodeAAF{l}{} \right)^2}_-(\Mean{\ArgHidNodeBAF{}{}}) \right) \; d\Mean{\ArgHidNodeBAF{}{}} ,
\end{split}
\end{align}
where, analogously to Eq.~\eqref{eq:mu_pm_def}, we defined
\begin{align}
\begin{split}
& \Einput{\left( \RCompHidNodeAAF{l}{} \right)^2}_+(\Mean{\ArgHidNodeBAF{}{}}) = \int_{0}^{\infty}  \NormalDens{\ArgHidNodeBAF{}{}}{\Mean{\ArgHidNodeBAF{}{}}}{\CVar{\Dataset}{\RCompHidNodeBAF{l+1}{}}} \ArgHidNodeBAF{}{}^2 d\ArgHidNodeBAF{}{}\,, \\
& \Einput{\left( \RCompHidNodeAAF{l}{} \right)^2}_-(\Mean{\ArgHidNodeBAF{}{}}) = \int_{-\infty}^{0}  \NormalDens{\ArgHidNodeBAF{}{}}{\Mean{\ArgHidNodeBAF{}{}}}{\CVar{\Dataset}{\RCompHidNodeBAF{l+1}{}}} \ArgHidNodeBAF{}{}^2 d\ArgHidNodeBAF{}{}.
\end{split}
\end{align}
From the above definitions, we see that 
\begin{equation}
\Einput{\left( \RCompHidNodeAAF{l}{} \right)^2}_+(\Mean{\ArgHidNodeBAF{}{}}) + \Einput{\left( \RCompHidNodeAAF{l}{} \right)^2}_-(\Mean{\ArgHidNodeBAF{}{}}) = \Einput{\left( \RCompHidNodeBAF{l}{} \right)^2}(\Mean{\ArgHidNodeBAF{}{}}) ;
\end{equation}
therefore substituting into Eq.~\eqref{eq:E_x2_g} we get
\begin{align}
\begin{split}
&\Eweights{\Einput{\left( \RCompHidNodeAAF{l}{j} \right)^2}} = \int_{0}^{\infty} \pdf{\Einput{\RCompHidNodeBAF{l}{j}}}{}{\Mean{\ArgHidNodeBAF{}{}}}  \Einput{\left( \RCompHidNodeBAF{l}{j} \right)^2}(\Mean{\ArgHidNodeBAF{}{}})\; d\Mean{\ArgHidNodeBAF{}{}} = \frac{1}{2} \int_{-\infty}^{\infty} \pdf{\Einput{\RCompHidNodeBAF{l}{j}}}{}{\Mean{\ArgHidNodeBAF{}{}}}  \Einput{\left( \RCompHidNodeBAF{l}{j} \right)^2}(\Mean{\ArgHidNodeBAF{}{}}) \; d\Mean{\ArgHidNodeBAF{}{}} = \frac{1}{2} \Eweights{\Einput{\left( \RCompHidNodeBAF{l}{j} \right)^2}}.
\end{split}
\end{align}

Finally, let us consider 
\begin{align}\label{eq:E_sg_ineq}
\begin{split}
\Eweights{\CVar{\Dataset}{\RCompHidNodeAAF{l}{}}} &  \equiv \Eweights{\Einput{\left( \RCompHidNodeAAF{l}{} \right)^2}} - \Eweights{\Einput{\RCompHidNodeAAF{l}{}}^2} = \frac{1}{2} \Eweights{\Einput{\left( \RCompHidNodeBAF{l}{} \right)^2}} - \Eweights{\Einput{\RCompHidNodeAAF{l}{}}^2} < \\
&< \frac{1}{2} \Eweights{\Einput{\left( \RCompHidNodeBAF{l}{} \right)^2}} - \frac{1}{2}  \Eweights{\Einput{\RCompHidNodeBAF{l}{}}^2} = \frac{1}{2} \Eweights{\CVar{\Dataset}{\RCompHidNodeBAF{l}{}}} \xrightarrow{\LayerNumNodes{l-1} ~\rightarrow~ \infty}  \frac{1}{2} \CVar{\Dataset}{\RCompHidNodeBAF{l}{}}.
\end{split}
\end{align}
where in the last step we used the concentration result discussed in App.~\ref{sec:RescGaussVar}, \emph{i.e.} that the distribution of $\CVar{\Dataset}{\RCompHidNodeBAF{l}{}}$ asymptotically narrows around its mean value, becoming, \emph{w.h.p.} independent of the realization $\WeightSet{}$. \newline
Now we have all the ingredient we need; in fact, from \eqref{eq:P_h_l} we know that
\begin{equation}
\CVar{\Dataset}{\RCompHidNodeBAF{l}{}} = \sigma_w^2 \Eweights{\CVar{\Dataset}{\RCompHidNodeAAF{l-1}{j}}}.
\end{equation}
Therefore, from \eqref{eq:E_sg_ineq}, we can conclude
\begin{equation}
\CVar{\Dataset}{\RCompHidNodeBAF{l+1}{}} < \left( \frac{\sigma_w^2}{2} -\epsilon\sp{\prime}_l \right) \CVar{\Dataset}{\RCompHidNodeBAF{l}{}}\,,
\end{equation}
with $\epsilon\sp{\prime}_l>0$.
Similarly, combining \eqref{eq:P_mu_l} with \eqref{eq:E_mu_ineq} we get
\begin{equation}
\CVar{\WeightSet{}}{\Einput{\RCompHidNodeBAF{l+1}{}}} > \left( \frac{\sigma_w^2}{2} +\epsilon_l \right) \CVar{\WeightSet{}}{\Einput{\RCompHidNodeBAF{l}{}}}\,,
\end{equation}
with $\epsilon_l > 0$.
Calling 
\begin{equation}
\epsilon \equiv \inf_l \epsilon_l\,,
\end{equation}
we can finally write 
\begin{align}
\begin{split}
 & \frac{\CVar{\WeightSet{}}{\Einput{\RCompHidNodeBAF{l+2}{}}}}{\CVar{\Dataset}{\RCompHidNodeBAF{l+2}{}}} > \left(\frac{2}{\sigma_w^{2} }\right)^l \frac{\CVar{\WeightSet{}}{\Einput{\RCompHidNodeBAF{2}{}}}}{\CVar{\Dataset}{\RCompHidNodeBAF{2}{}}} \left( \frac{\sigma_w^2}{2} +\epsilon \right)^l > l \underbrace{\left(
 \frac{2}{\sigma_w^{2} } \frac{\CVar{\WeightSet{}}{\Einput{\RCompHidNodeBAF{2}{}}}}{\CVar{\Dataset}{\RCompHidNodeBAF{2}{}}} \epsilon \right)}_{\equiv \frac{1}{C}} \\
 & \Longrightarrow \frac{\CVar{\Dataset}{\RCompHidNodeBAF{l+2}{}}}{\CVar{\WeightSet{}}{\Einput{\RCompHidNodeBAF{l+2}{}}}} < C \frac{1}{l}.
 \end{split}
\end{align}
Therefore if we consider an infinite depth network, \emph{i.e.} $L \rightarrow \infty$, we have:
\begin{equation}
\frac{\CVar{\Dataset}{\RCompHidNodeBAF{L}{}} }{\CVar{\WeightSet{}}{\Einput{\RCompHidNodeBAF{L}{}}}} = \mathcal{O} \left( \frac{1}{L} \right) \Longrightarrow \frac{\CVar{\WeightSet{}}{\EinputComp{c}}}{\CVar{\Dataset}{\ROutNode{c}{}}} = \mathcal{O} \left( L \right) \Longrightarrow \lim_{L \rightarrow \infty} \frac{\CVar{\WeightSet{}}{\EinputComp{c}}}{\CVar{\Dataset}{\ROutNode{c}{}}} = \infty \,.
\end{equation}
\end{proof}

We observe a fundamental distinction compared to the previous cases. Network depth can amplify the level of IGB in systems where it is already present. However, depth alone does not induce IGB. Considering a model with a linear activation function and no max pooling after the first hidden layer, and using Gaussian data centered at zero, we have $\Einput{\RCompHidNodeAMP{1}{1}{}} = \Einput{\RCompHidNodeBAF{1}{}{}} = 0$. Thus, there is no Node Symmetry Breaking (NSB) in the second layer; in other words, the nodes in the second layer are Gaussian \textit{r.v.}s centered at zero. It can be easily shown that, by iterating this process through each layer, the same scenario will be repeated in each subsequent layer up to the output layer. Therefore, NSB does not occur even in the output nodes, resulting in the absence of IGB in the system.

\section{Extension to multi-class problems}\label{sec:MC_extension}

We want now to extend the analysis to the multi-class case, \emph{i.e.} to problems with $\NumberClasses > 2$.
Following the framework we introduced earlier,  we can easily extend the computation. 
We will again consider the distribution among the $\NumberClasses$ classes of the dataset elements after initialization. In particular, we can define $\NumberClasses$ values associated with the fraction of dataset elements classified as belonging to each of the $\NumberClasses$ classes, $\left\{ \RMultiClassFraction{i}{} \right\}_{i=0}^{\NumberClasses-1}$. 
We could also consider the distribution of the sorted frequencies; in other words, in each experiment, we order the classes according to the corresponding frequencies and not by their label. To distinguish these frequencies (and their statistics) from the set $\left\{ \RMultiClassFraction{i}{} \right\}_{i=0}^{\NumberClasses-1}$ , we use indices $\tilde i$, and call this new set $\left\{ \RMultiClassRankedFraction{i}{} \right\}_{\tilde i=0}^{\NumberClasses-1}$. In particular $\RMultiClassRankedFraction{0}{}$ indicates the biggest frequency among the $\NumberClasses$, $\RMultiClassRankedFraction{1}{}$ the second one and so on.
Finally we add the set (or subset) cardinality, $M$,  on the  output nodes set $\left\{ \ROutNode{c,}{M} \right\}_{c=0}^{M-1}$.\footnote{Here, the cardinality of the set indicates the number of classes over which the statistics are computed. The statistics of the \textit{r.v.} change with the number of classes (or nodes), i.e., with the cardinality.}
 Analogously to the set of fractions, we define the set of ranked output nodes $\left\{ \ROutNode{\tilde c,}{M} \right\}_{\tilde c=0}^{M-1}$ 
In the following, similarly to Sec.~\ref{sec:SLP_analysis} we will derive the distribution of $\RMultiClassFraction{0}{}$.

\begin{figure}[h]
    \centering
    \includegraphics[width=.78\textwidth]{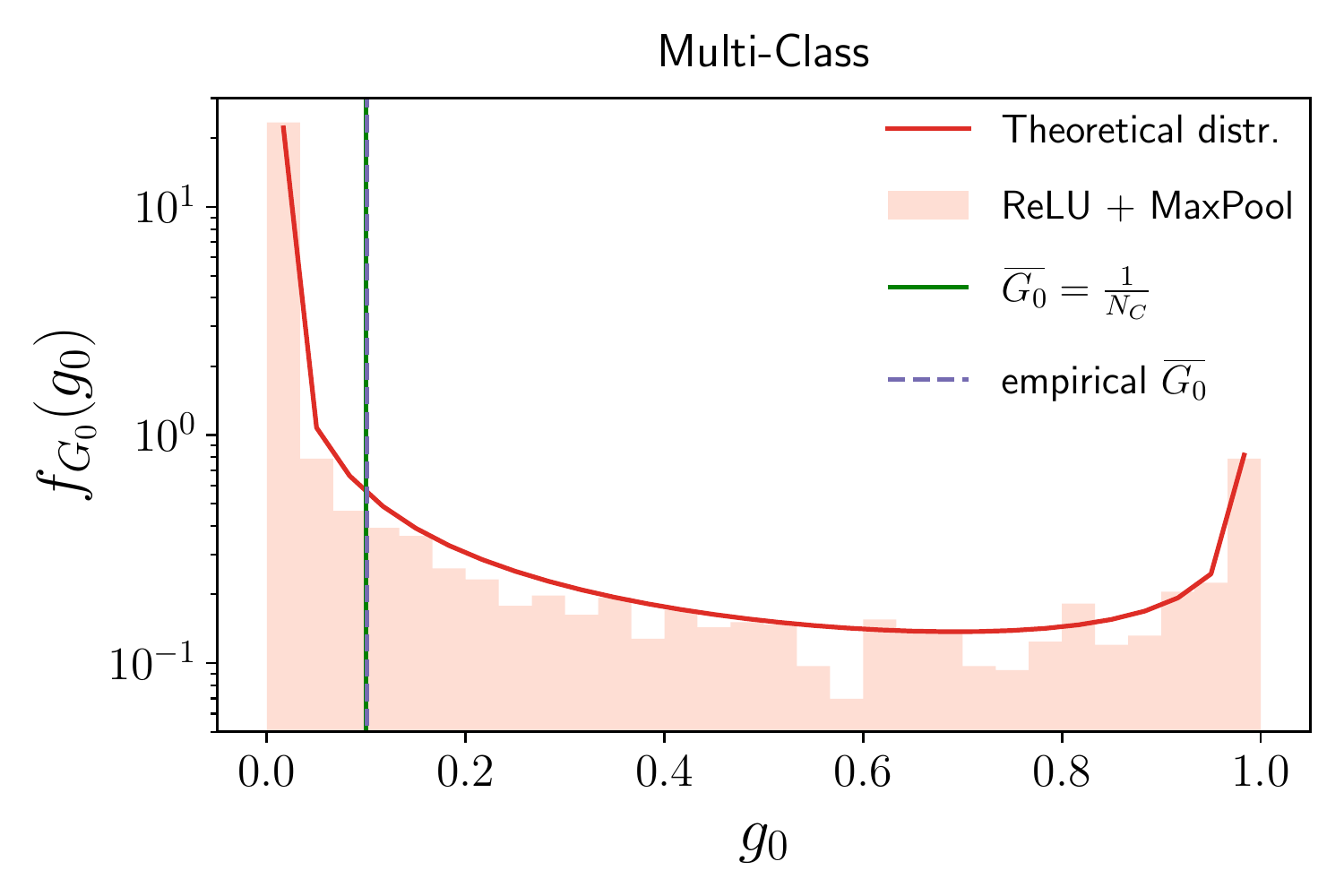}
    \caption{The distribution $\pdf{\RClassFraction{0}}{}{\ClassFraction{0}}$  in a multi-class problem with $\NumberClasses=10$. The empirical mean value is reported; note that despite the difference in the distribution the mean value is the same that we would observe using a linear activation function $ \Eweights{\RMultiClassFraction{0}{}} = \frac{1}{\NumberClasses}$. For these simulations we used the dataset \GB~ on the model \MLPA~ with max pooling ($k=20$). (see App.~\ref{sec:reprod} for more details).}
\label{MaxPool_Pf0_distr_multi}
\end{figure}
We will make in the derivation of $\RMultiClassFraction{0}{}$ the following approximation: \textit{We will repeat the same approach of the binary setting comparing the generic class "$0$" with the class with bigger mean output value among the $\NumberClasses -1$ remaining classes}. \newline
The first ingredient that we need is therefore the statistics of $\ROutNode{\tilde 0,}{\NumberClasses -1}$. From the analysis of previous sections, we know that $\pdf{\ROutNode{\tilde 0,}{\NumberClasses -1}}{(\Dataset)}{\OutNode{}{}}$ is asymptotically Gaussian. The mean value of this Gaussian variable will follow the distribution of the maximum among $\NumberClasses - 1$ drawn from the Gaussian distribution $\pdf{\RMean{c}}{}{\Mean{}}$ 
as, by definition,
\begin{equation}
  \Einput{\ROutNode{\tilde 0,}{\NumberClasses -1}}  \equiv \max_{c \in \{1, \dots,  \NumberClasses  \}} \Einput{\ROutNode{c}{}}\,.
\end{equation}
\paragraph{Maximum over $\NumberClasses-1$ Gaussian draws}
We will follow the same approach used in App.~\ref{sec:MaxPool_effect} . We will again start from Eq.~\eqref{P_Y_relu}. In this case,
\begin{equation}\label{eq:P_X_MULTI}
\pdf{X}{}{x} \equiv \pdf{\EinputComp{c}}{}{x} =   \NormalDens{x}{0}{\hat{\sigma^2}}\,,
\end{equation} 
where we used a generic $\hat{\sigma^2}$ to indicate the variance. We will perform the computation in this general settings; then to analyze the specific cases we can just substitute the right variance for the specific case (see App.~\ref{sec:E_chi_O}).

\begin{equation}
F_X(y) \equiv \int_{-\infty}^{y} \pdf{X}{}{x}dx = \int_{-\infty}^{y} \NormalDens{x}{0}{\hat{\sigma^2}} dx = \frac{1}{2} \left( 1 + \erf \left( \frac{y}{\sqrt{2} \hat{\sigma}}\right) \right)\,.
\end{equation}

Now, putting all pieces together in Eq.~\eqref{P_Y_relu}, we get our target distribution,
\begin{equation}\label{eq:P_Y_MULTI}
 \pdf{Y}{}{y} \equiv   \pdf{\Einput{\ROutNode{\tilde 0,}{\NumberClasses -1}}}{}{y}  = \left( \NumberClasses -1 \right)    \NormalDens{y}{0}{\tilde \sigma^2} \left(\frac{1}{2} \left( 1 + \erf \left( \frac{y}{\sqrt{2} \tilde \sigma}\right) \right) \right)^{\NumberClasses -2}\,.
\end{equation} 

Now we can proceed following again the steps presented at the end of App.~\ref{sec:SLP_analysis}. The only difference we will have is in Eq.~\eqref{eq:P_f0_MF}; now we will have
\begin{equation}\label{eq:P_f0_MULTI}
 \pdf{ \RMultiClassFraction{0}{} }{}{\MultiClassFraction{0}{} } = \int_{-\infty}^{\infty} \pdf{\Einput{\ROutNode{ 0,}{\NumberClasses}}}{}{\tilde x}  \pdf{\Einput{\ROutNode{\tilde 0,}{\NumberClasses -1}}}{}{\tilde x - \hat{\Delta} \left( \MultiClassFraction{0}{}  \right) } d \tilde{x}\,,
\end{equation}
where $ \Einput{\ROutNode{\tilde 0,}{\NumberClasses -1}}$ the biggest mean output among the $\NumberClasses -1$. Note that $\hat{\Delta} \left(  \MultiClassFraction{0}{}  \right)$ is a function of $\MultiClassFraction{0}{}$. Substituting Eq.~\eqref{eq:P_X_MULTI} and Eq.~\eqref{eq:P_Y_MULTI} we have:
\begin{equation}
\pdf{ \RMultiClassFraction{0}{}}{}{\MultiClassFraction{0}{} } = \int_{-\infty}^{\infty} \NormalDens{\tilde x}{0}{\hat{\sigma^2}} (\NumberClasses-1) \NormalDens{\tilde x}{\hat \Delta\left( \MultiClassFraction{0}{}  \right)}{\hat{\sigma^2}} \left(\frac{1}{2} \left( 1 + \erf \left( \frac{\tilde x - \hat \Delta\left(  \MultiClassFraction{0}{}  \right)}{\sqrt{2} \hat \sigma}\right) \right) \right)^{\NumberClasses -2}  d\tilde x\,.
\end{equation}

\subsection{Increasing number of classes exacerbates IGB}
Having a large number of classes increases the probability that the center of the distribution related to one of these classes is an outlier of the distribution \( \pdf{\RMean{c}}{}{\Mean{}} (x) \). However, since \( \pdf{\RMean{c}}{}{\Mean{}} \) is Gaussian (a fast-decaying distribution), the typical value of the maximum increases slowly. Therefore, an extremely large number of classes, \( N_c \), will be needed to observe significant changes.

We can analyze this quantitatively.
The typical value of the maximum out of \( N_c \) Gaussian random variables of variance \( \epsilon \) grows as \( \sqrt{(2 \epsilon \log(N_c))} \)~\cite{gumbel1958statistics,hartarsky2019maximum}.
In the absence of IGB, \( \epsilon \) is the variance of \( \pdf{\RMean{c}}{}{\Mean{}} \) and scales inversely with the dataset size, i.e., \( \sim 1/D \) (see e.g., App.~\ref{sec:E_chi_O} and Fig.~\ref{D_Comparison}). This implies that outliers will only appear if the number of classes far exceeds the number of datapoints, a situation that does not occur in real life.

In summary, if there is no IGB, a large enough dataset ensures that increasing the number of classes does not affect the presence of IGB. 

Conversely, if IGB is present, a sufficiently large number of classes may result in high imbalance even with a relatively small value of \( \VarRatio \).

\section{Experiments}\label{app:exp}
The analysis presented is able to identify IGB and analytically describe the phenomena in a variety of settings (\textit{e.g.}, MLPs with different activation functions, arbitrary depth, and the presence or absence of pooling layers). However, IGB is significantly broader in scope, as it may be observed in a wide range of dataset and architectural combinations. Although our theoretical analysis does not go into these scenarios (a further extension of the analytical conclusions to more sophisticated designs will be the focus of future research), we provide some representative instances to demonstrate the breadth of scenarios where IGB is significant in real-world circumstances.
The experiments presented, while not exhaustive, aim to provide insights and emphasize three fundamental points that will be developed in a separate publication:
\begin{itemize}
    \item IGB is amplified in structured data, particularly in cases of strongly correlated data (App.~\ref{app:exp_data}).
    \item IGB is a ubiquitous phenomenon, that happens with most combinations of datasets and architectures (App.~\ref{app:exp_data}, \ref{app:dyn}).
    \item The differences induced by IGB on the initial state of the network have an impact on its dynamics, rendering it qualitatively distinct, especially in the initial phase, compared to the case without IGB (App.~\ref{app:dyn}).
\end{itemize}

\subsection{Experiments on real data}\label{app:exp_data}
In this section, we show some empirical evidence of IGB on combinations of models and data which are not covered by our theory, such as CNN architectures and the MNIST, CIFAR10 and CIFAR100 datasets.\footnote{MNIST is a particularly insightful dataset for IGB, because the correlations between pixels are high. This is opposite to what we cover with our theory, which is based on i.i.d. inputs.}
In App.~\ref{app:dyn}, we will also include experiments on additional architectures (ResNet, MLP-mixer and Vision Transformers).
The experiments we will discuss do not presume to show in a complete and exhaustive way the range of models and cases in which IGB can occur; this question is outside the scope of our work. Rather, they are some didactic examples intended to show how, beyond the assumptions used in our analysis to quantitatively treat the phenomenon, IGB also emerges in a broader context.  We first present and discuss the experiments. For technical details for the reproducibility we refer the reader to App.~\ref{sec:reprod}.

\subsubsection{A prototype for highly correlated data: MNIST}\label{sec:mnist}
\begin{figure}
    \includegraphics[width=0.48\textwidth]{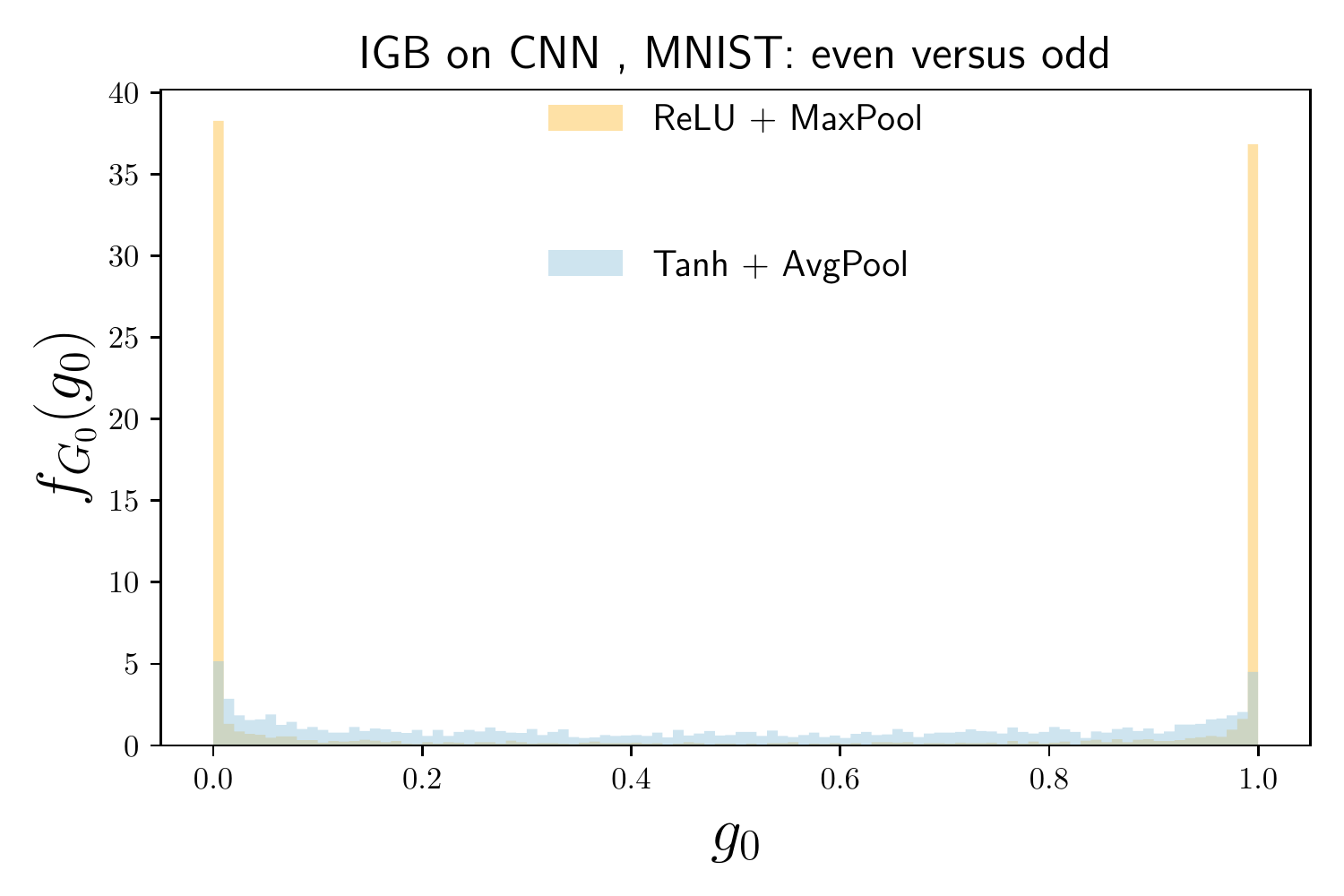}
    \includegraphics[width=0.48\textwidth]{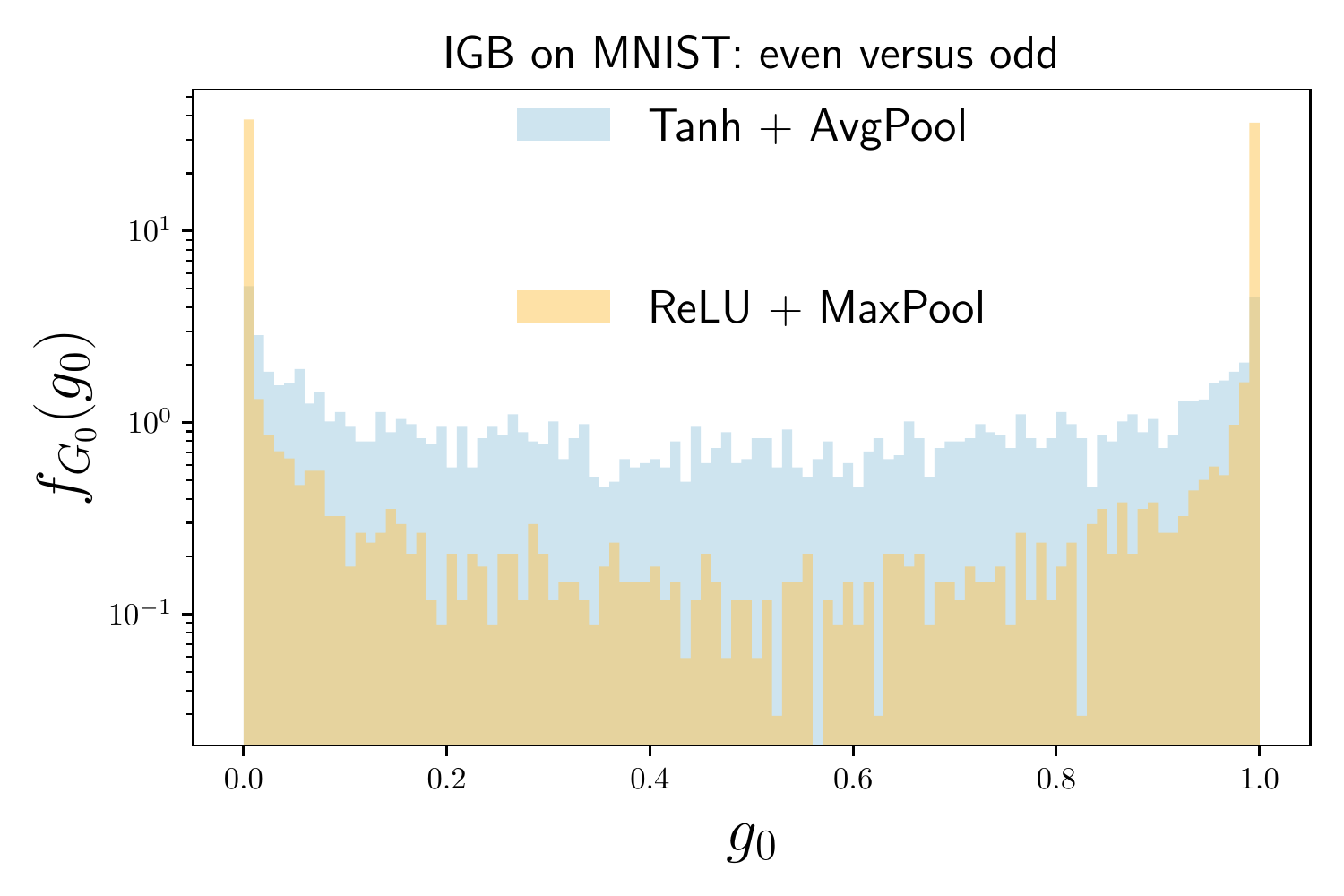}
\caption{Comparison of the distribution $\pdf{\RClassFraction{0}}{}{\ClassFraction{0}}$ between two untrained neural networks fed with MNIST dataset; the ten classes of the dataset were merged in two macro groups: even and odd. The two architectures employed in the comparison differ in the choice of activation function and pooling. For these simulations we used the \MNIST~dataset on the model \CNNB. (see App.~\ref{sec:reprod} for more details). On the right, the same plot in a logarithmic scale to better visualize the differences in the low-density regions.}
\label{fig:Binary_MNIST_CNN}
\end{figure}

In the analysis we presented in the main part of the paper, we focused on a simple data structure, to underline effects induced by architecture design: a remarkable conclusion from this setting is the following:\newline
\textit{In a classification problem, an untrained network with biases set to zero and fed with i.i.d. data-points (i.e. different classes are identically distributed) often starts the training process with a strong bias toward one of the classes.}\newline
For our theoretical analysis, we placed ourselves in a setting where the effect of IGB is minimal and unaffected by the data structure. This ensures that the sources of IGB, which our analysis links to architectural design elements, do not stem from dataset characteristics. 
Below, we present experiments in which these hypotheses on data are violated, to demonstrate how the phenomenon manifests itself (more prominently) in real datasets, where:
\begin{itemize}
    \item Data-points belonging to different classes will not be identically distributed.
    \item Components of a single data-point will not be independent; for example pixels of an image will clearly be correlated.
    \item Similarly, correlations between different data-points (belonging to the same class) is possible.
\end{itemize}

MNIST constitutes a good candidate to represent a scenario of strong correlations in the data. In fact, in this dataset the images are characterized by a small percentage of nonzero pixels; this leads to the formation of areas, within the same image, characterized by similar values and therefore strongly correlated. In addition, image areas are strongly correlated between different elements. For example, we are unlikely to observe the writing of a number on the edges of an image; therefore, different images will have similar values of pixels along the edges. When considering images of the same class, this correlation is even more pronounced. 
From the MNIST dataset, we defined two macro-classes, even and odd, by merging the 10 starting classes according to their parity. We then propagated the binary dataset, obtained in this way, through two untrained convolutional neural networks. The difference between the two networks lies in the choice of activation and pooling function employed.
In particular the dataset is propagated through \CNNB~(network details in App.~\ref{sec:reprod}). 
The \CNNB~model is a CNN where the last convolutional layer is directly connected to the outputs. We choose it this way, because, as earlier shown, IGB cannot arise without hidden layers. This indicates that the observation of IGB is caused by the convolutional layers or, to put it more precisely, it emerges from the propagation of structured data through the convolutional layers.

We show this in Fig.~\ref{fig:Binary_MNIST_CNN}, which contains two important messages:
\begin{itemize}
    \item[$\diamondsuit$] Unlike the cases presented in our theoretical analysis, neither of the two choices has total absence of IGB. This is consistent with the intuition that correlations in the data cause or amplify IGB.
    \item[$\diamondsuit$] While we observe presence of IGB in both cases, this is more pronounced for the architecture with ReLU and max pooling, consistent with what we saw in our study.
\end{itemize}
These observations show how IGB is a rather universal phenomenon, related to the combined effect of data structure and the design of the neural network.

\subsubsection{A further example: CIFAR10}\label{sec:cifar}
\begin{figure}
    \includegraphics[width=0.48\textwidth]{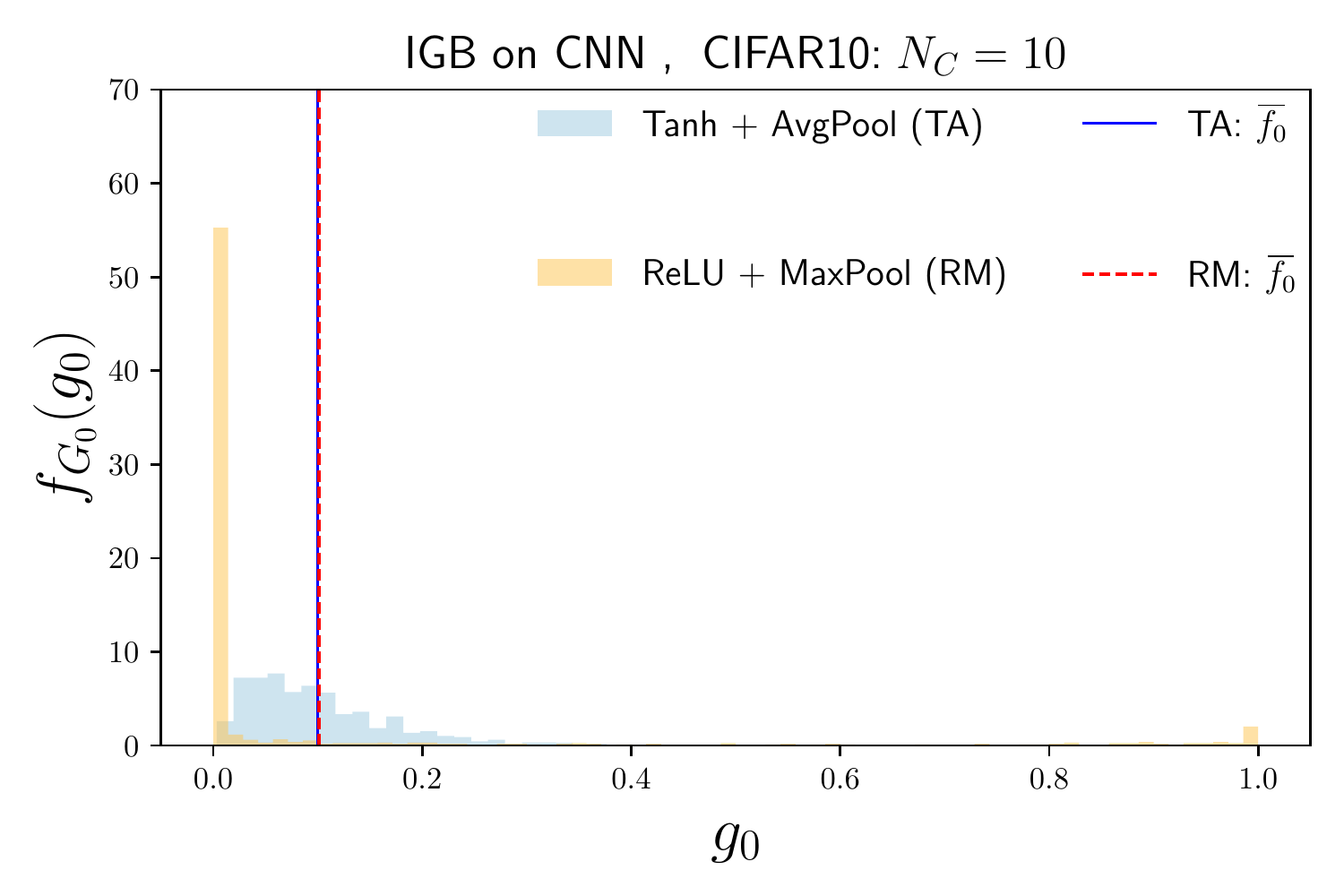}
    \includegraphics[width=0.48\textwidth]{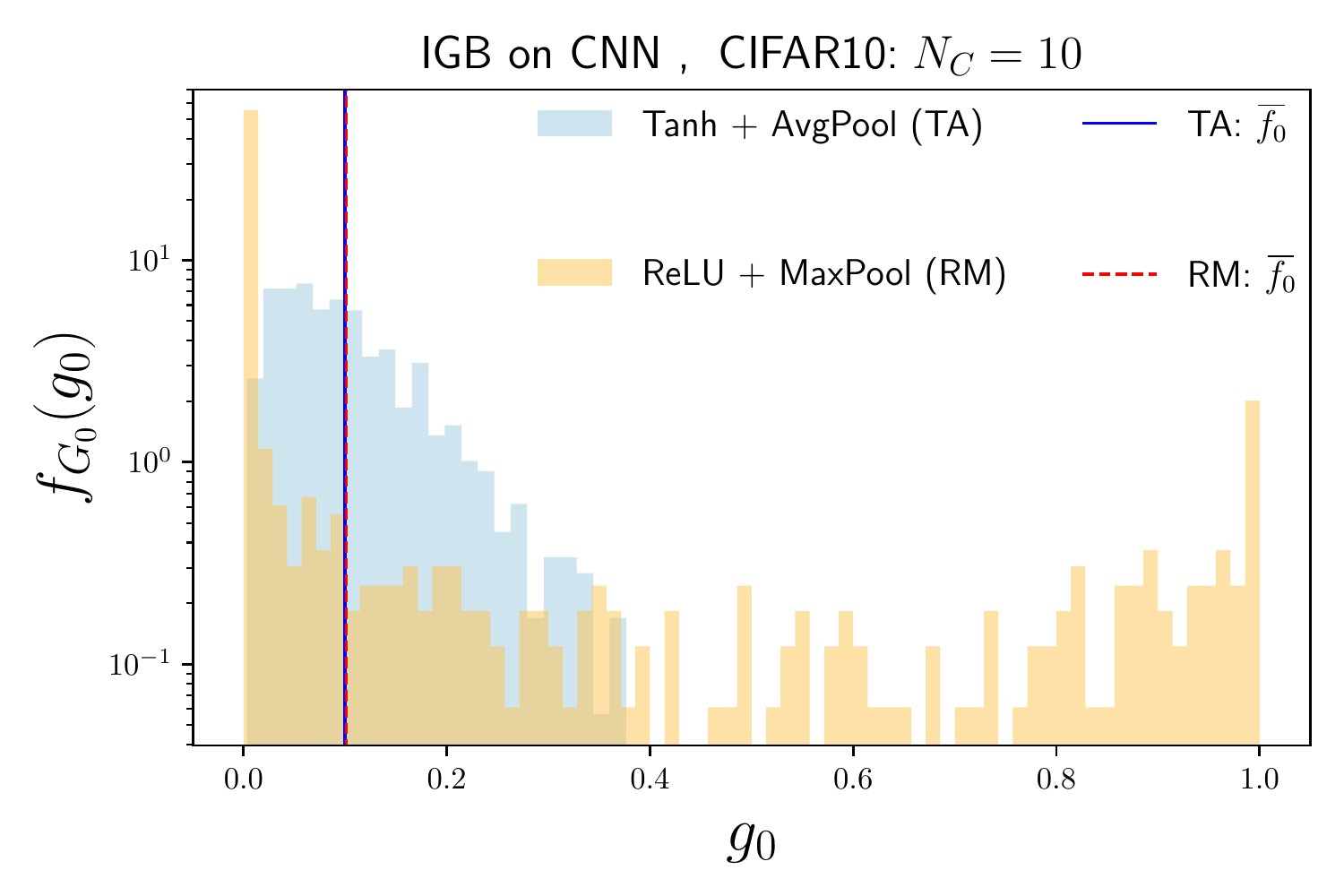}
\caption{Comparison of the distribution $\pdf{\RClassFraction{0}}{}{\ClassFraction{0}}$ between two untrained neural networks fed with CIFAR10 dataset (with 10 classes). The two architectures employed in the comparison differ in the choice of activation function and pooling. In the absence of IGB, the distributions would concentrate around the vertical line.
For these simulations we used the dataset \CIFAR~on the model \CNNB~described below (see App.~\ref{sec:reprod}). On the right, the same plot in a logarithmic scale to better visualize the differences in the low-density regions.}
\label{fig:Multi_CIFAR10_CNN}
\end{figure}
We now show a multi-class example, on the same network as in App.~\ref{sec:mnist} (\CNNB) we propagated CIFAR10. The results are reported in Fig.~\ref{fig:Multi_CIFAR10_CNN}. As with MNIST, the results are qualitatively consistent with the conclusions of our theory: the network with ReLU and max pooling displays a stronger IGB than its counterpart with $\tanh$ and AvgPooling. 
Note, also, how in both cases, symmetry between classes is preserved at the ensemble level. In both cases, in fact, we get $\Eweights{\RMultiClassFraction{c}{}} = 1/ \NumberClasses$ .
To show this we show in Fig.~\ref{fig:Multi_CIFAR10_CNN} two vertical lines at the mean values calculated on the empirical distributions (histograms), each surrounded by its own uncertainty, estimated through the standard error. 
\subsubsection{High cardinality dataset: Cifar100}

We also provide another example of a multi-class dataset. In this case, we choose Cifar-100 as a prototype of a high-cardinality dataset (with a high number of classes) to demonstrate the presence of IGB in this scenario as well. In the absence of IGB, the distribution should be peaked at $ \RClassFraction{0} = 1/ \NumberClasses = 1/100$. However, the histogram shows a peak at 0 (most of the time, the generic class does not receive any assigned elements). The gap between the peak at 0 and the rest of the distribution, particularly evident in the log-log scale plot, indicates a resolution limit due to the finite number of dataset elements ($\DatasetSize = 10^4$).

\begin{figure}
    \includegraphics[width=.48\textwidth]{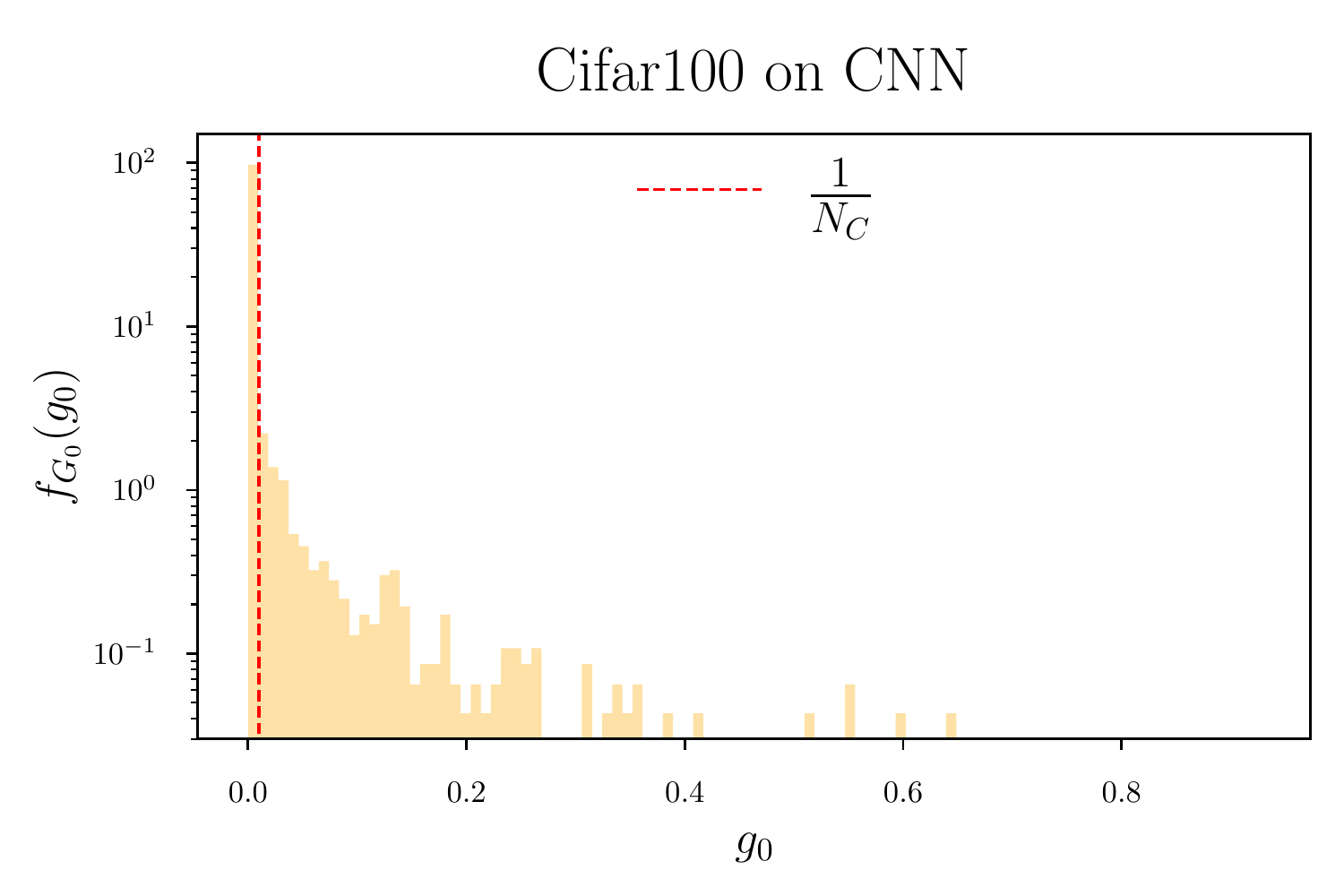}
    \includegraphics[width=.48\textwidth]{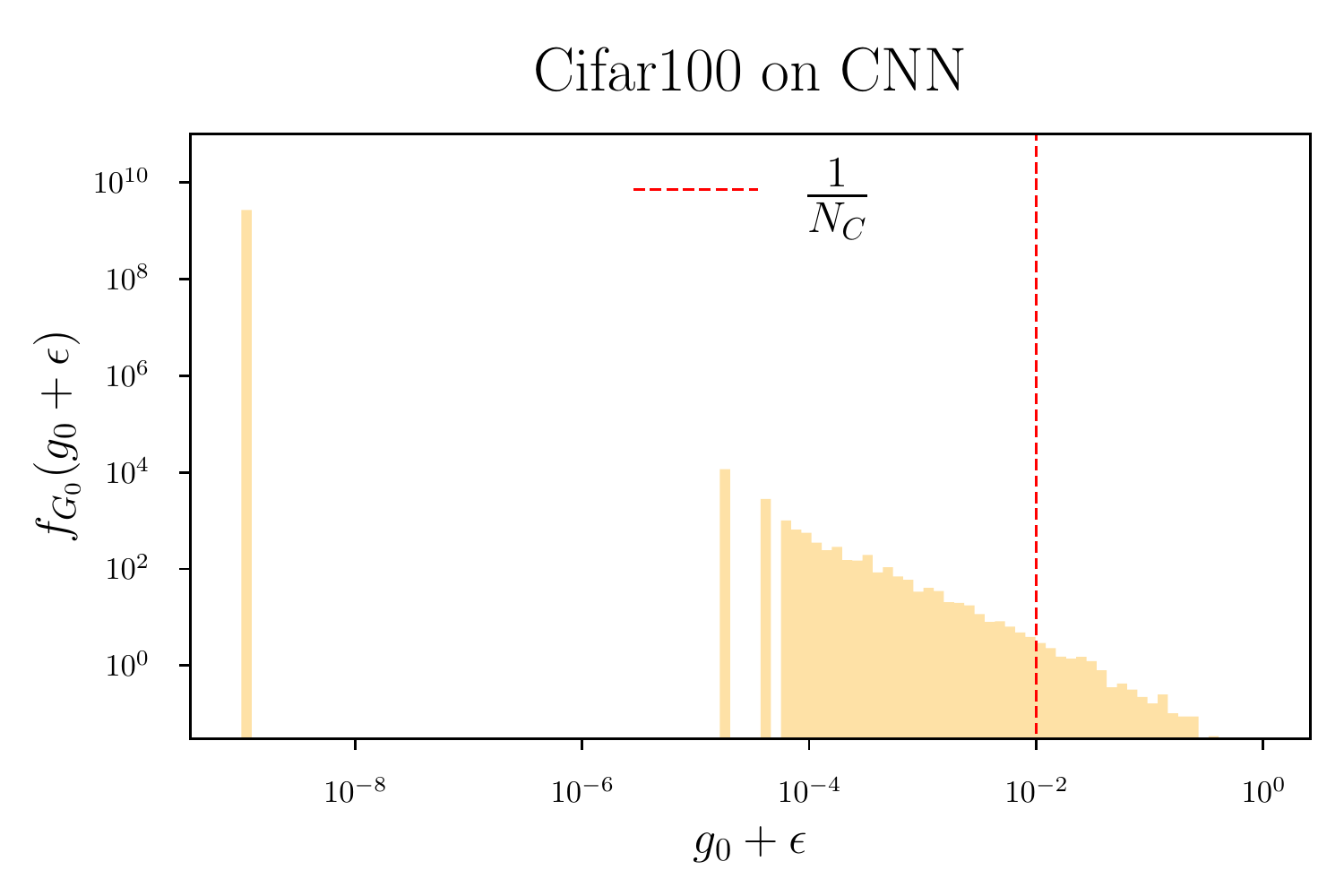}
    \caption{\CNNA~(ReLU + max pool case) with \CIFARHC~ in this case ($N_C=100$). The peak at $\RClassFraction{0} = \frac{1}{\NumberClasses}$, corresponding to the No IGB case (in the absence of IGB, the distribution should concentrate around this value), is reported as a reference. The plot on the right contains the same data as that on the left, but has a logarithmic $x$ axis. We added a small numer, $\epsilon = 1e-9$, to the value of $\RClassFraction{0}$ in order to show the peak at $\RClassFraction{0}=0$.} 
\end{figure}

\begin{figure*}[h]
    \centering    
    \vspace{-3mm}
    \includegraphics[width=.9\textwidth]{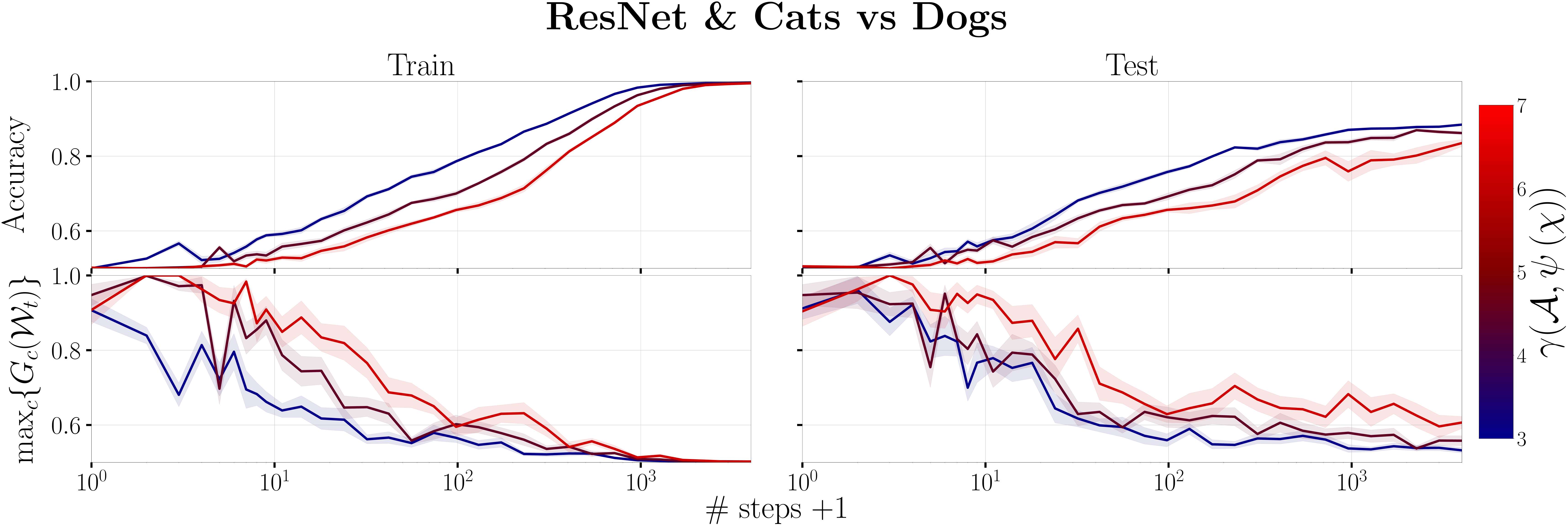}
        \vspace{4mm}

        \includegraphics[width=.9\textwidth]{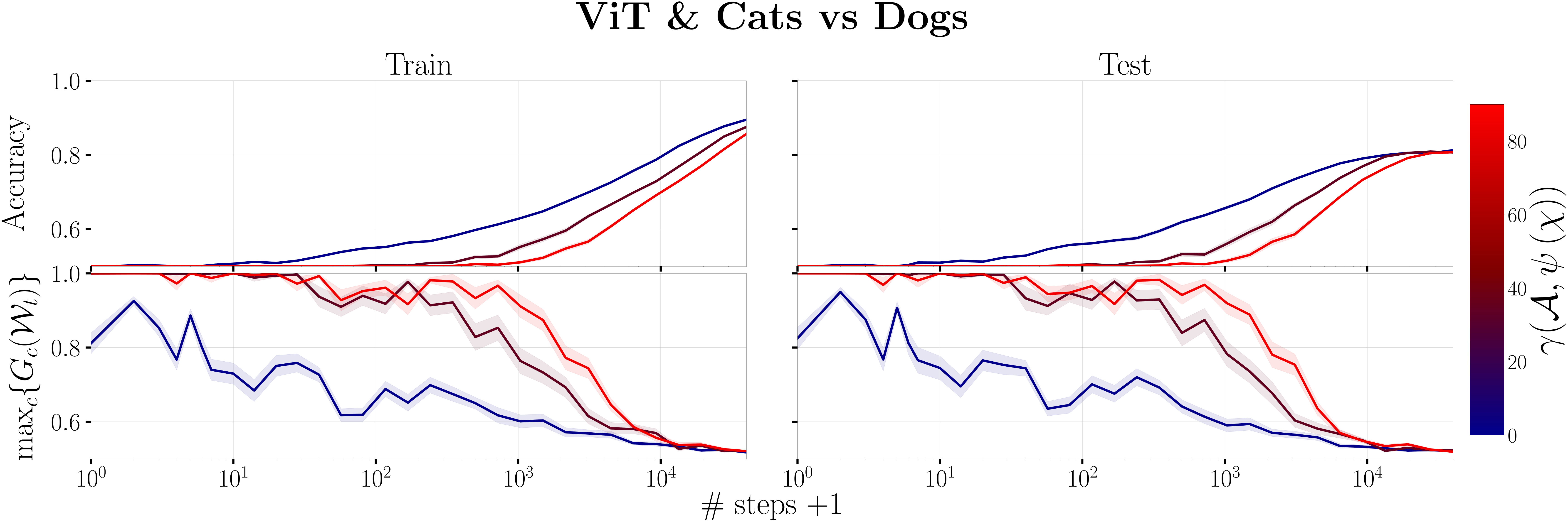}
    \caption{
    \textbf{Right}: Comparison of the trend of $\max_i \{ \RClassFraction{i} \}$ with that of accuracy during the learning dynamics, varying with the level of guessing bias \underline{at initialization} (IGB). The curves show a consistent pattern with the diagram on the shown in Fig.~\ref{fig:Acc_bound}. The simulations were conducted on a ResNet (top) and Vision Transformer (bottom) using a binary dataset (dogs \textit{vs} cats from CIFAR) as input; more details on the setting  are provided in App.~\ref{sec:reprod}.
    } \label{fig:dyn_exp_CD}
\end{figure*}

\begin{figure*}[!th]
    \centering    
    \vspace{-3mm}
    \includegraphics[width=.9\textwidth]{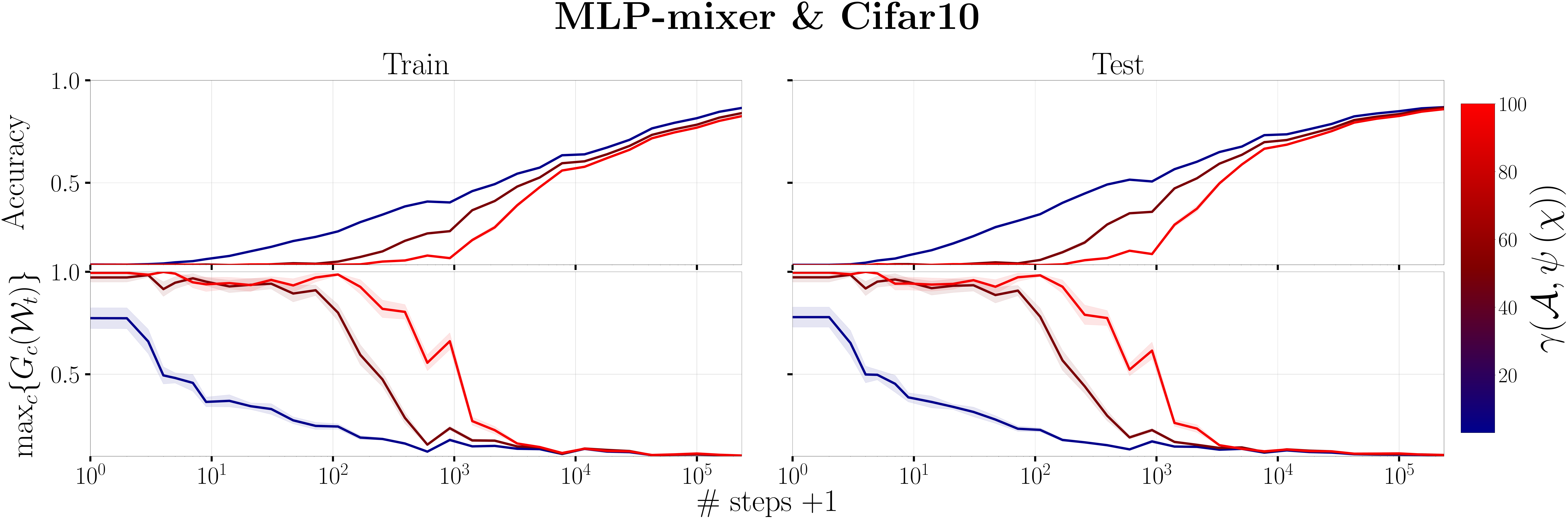}
        \vspace{4mm}

        \includegraphics[width=.9\textwidth]{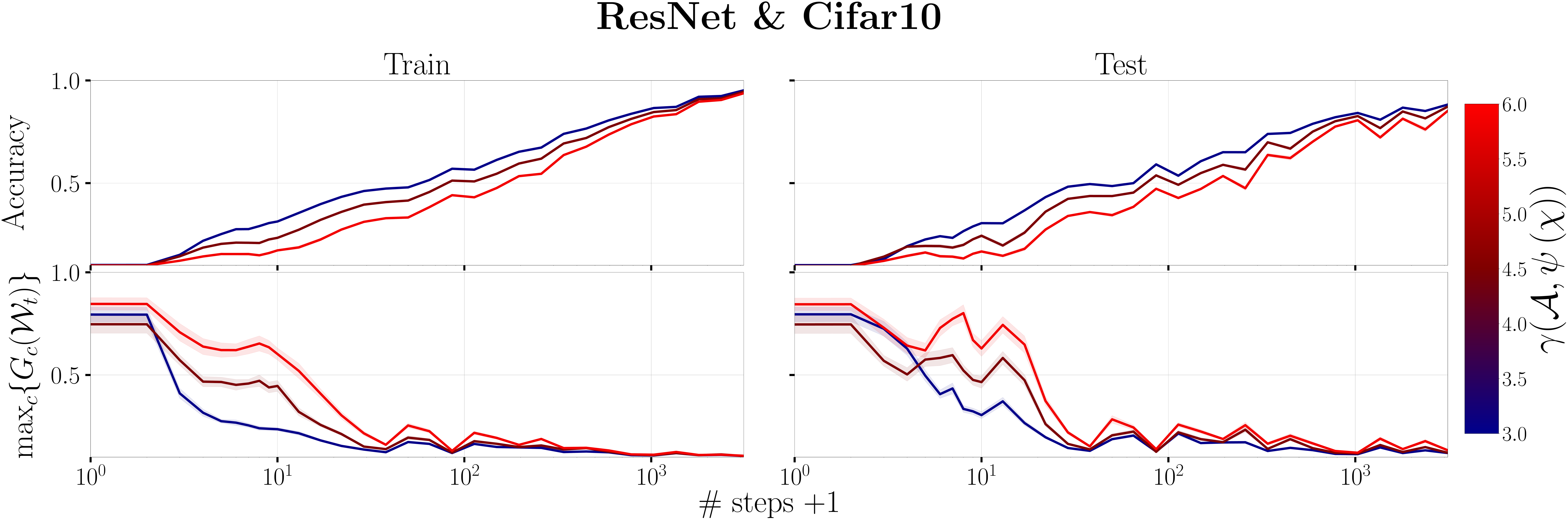}
                \vspace{4mm}

        \includegraphics[width=.9\textwidth]{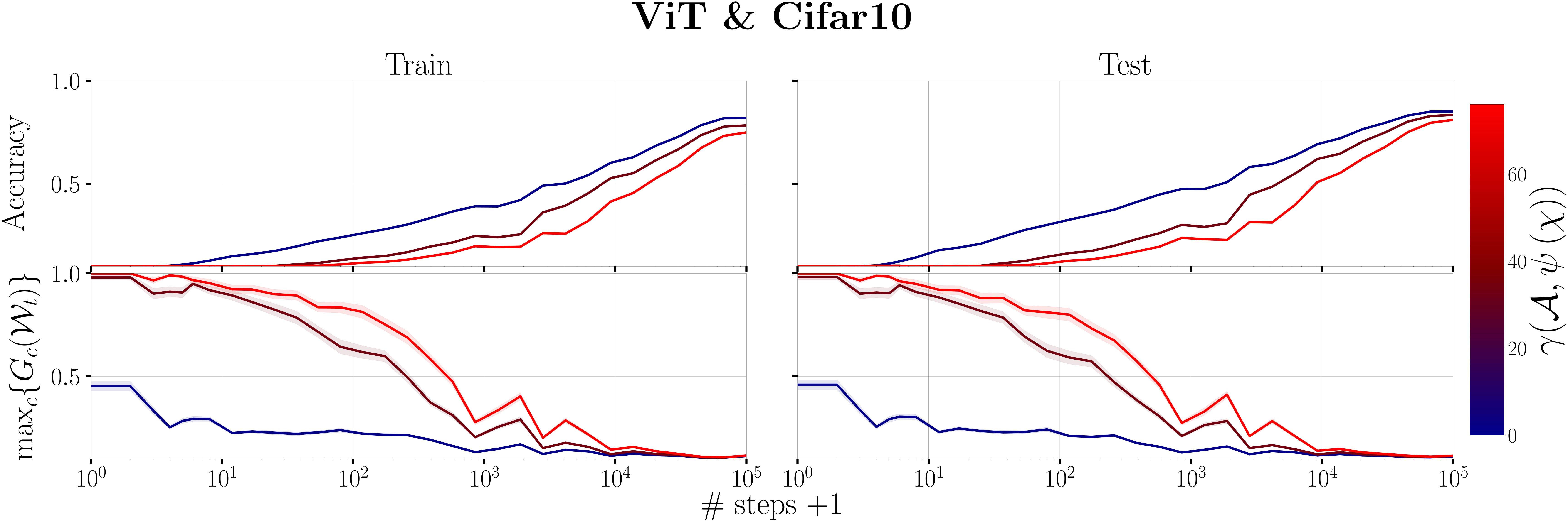}
    \caption{
    \textbf{Right}: Comparison of the trend of $\max_i \{ \RClassFraction{i} \}$ with that of accuracy during the learning dynamics, varying with the level of guessing bias \underline{at initialization} (IGB). The curves show a consistent pattern with the diagram on the shown in Fig.~\ref{fig:Acc_bound}. The simulations were conducted on different architectures (MLP- mixer (top), ResNet (middle) and Vision Transformer (bottom)) using a multi-class dataset (\CIFAR) as input; more details on the setting  are provided in App.~\ref{sec:reprod}.
    } \label{fig:dyn_exp_Cifar}
\end{figure*}

\subsubsection{Additional architectures}
We also provide empirical evidence of IGB in different architectures.
To improve the legibility of the paper, we show results for \ResNet, \ViT~ and \MLPmix~in Sec.~\ref{app:dyn}. 
While App.~\ref{app:dyn} is devoted to the dynamics und IGB, the reader can recognize the presence of IGB in Figs.~\ref{fig:dyn_exp_CD} and \ref{fig:dyn_exp_Cifar}.
Specifically, IGB is identified by checking the quantity $\max_c\{\RClassFraction{c}(\mathcal{W}_t)\}$, for $t=0$.; we can observe that at time $t=0$, \textit{i.e.}, at initialization, each of these curves shows $\max_c\{ \RClassFraction{c} \} \neq 1/\NumberClasses$.\footnote{The reader might have particular interest in the $K=0$ case, where the data is centered and IGB cannot be attributed to data preprocessing.}

\subsection{Experiments on other architectures \& effects on the training dynamics}\label{app:dyn}

The experiments presented in App.~\ref{app:exp_data} illustrate how Initial Guessing Bias (IGB) can emerge in CNNs. This section provides examples demonstrating the breadth of settings and architectures where this phenomenon is observed. In particular, beyond illustrating the emergence of IGB in additional architectures (ResNet, Vision Transformer, and MLP-mixer), we show the impact of the phenomenon on the dynamics.

While an exhaustive review is beyond the scope of this study, we present key experimental results that emphasize the practical significance of IGB, suggesting its relevance for further investigation. Given the observation of IGB in various settings, including advanced architectures that achieve state-of-the-art performance, it seems that IGB does not profoundly affect the final convergence. However, qualitative differences from non-IGB cases and similarities to class imbalance situations indicate potential challenges during training associated with IGB.

Experiments in Sec.~\ref{sec:Imp_IGB} show that the time required to absorb IGB increases with the level of IGB itself. Since the absorption of IGB is necessary for performance improvement in training balanced datasets, a higher level of IGB translates to slower convergence. Similar results are observed across various settings: in Fig.~\ref{fig:dyn_exp_CD}, for binary classification extended to other architectures (ResNet and Vision Transformers), and in Fig.~\ref{fig:dyn_exp_Cifar}, the experiments are repeated for multi-class cases with analogous outcomes.

As discussed in App.~\ref{sec:ampl_IGB}, there are various ways to increase the level of IGB. In the experiments shown (Fig.~\ref{fig:Acc_bound}, Fig.~\ref{fig:dyn_exp_CD}, Fig.~\ref{fig:dyn_exp_Cifar}), IGB is tuned through data standardization, where inputs are pre-processed as:
\begin{align}
    \PreprDataOp{\InputValue_b^{(a)}} = \InputValue_b^{(a)} + K\,.
\end{align}
The experiments are repeated for different values of $K$ (specifically $K \in \{ 0, 2, 4 \}$). Increasing $|K|$, the level of IGB also rises, measured using $\VarRatio$. This choice of IGB amplification allows for comparison without altering the architectural design or data structure (inputs are simply shifted by a constant value).

Even in the case of $K=0$, the system exhibits IGB, as indicated by $\max_i \{ \RClassFraction{i} \} \neq 1/\NumberClasses$ at initialization. Also, it is worth noticing that the susceptibility of networks to IGB varies, as seen from the comparison of the $\VarRatio$ range in the colormaps of different experiments. Despite using the same set of $K$ values, the resulting IGB level varies significantly across networks. The IGB values, dependent on the combined effect of $\PreprData$ and $\Arch$, show substantial variation among the considered architectures, suggesting the presence of regulatory or amplifying elements for IGB within these architectures.

\subsection{IGB in Pre-Trained Models}\label{app:pt-m}

Thus far, the analysis and experiments presented have focused on randomly initialized DNNs. While the results in App.~\ref{app:dyn} provide valuable insights into the effects of IGB on the learning dynamics of untrained networks, we now demonstrate that even pre-trained models exhibit the presence of IGB.

Transfer learning is becoming an increasingly prevalent paradigm \citep{han2021pre}, where large models are pre-trained on extensive datasets and then fine-tuned for specific, different tasks. In practice, before initiating the fine-tuning phase, the final layer, referred to as the head, is replaced by untrained layer(s). This replacement ensures compatibility with the new task, such as modifying the number of output nodes in the classifier (i.e., the last fully connected layer) to match the number of classes in the new task.

While the head could, in principle, consist solely of the classifier, other configurations with more complex structures for the head are possible \citep{ren2023prepare}. Using a more articulated head is a common practice, especially in situations where the head is the only part of the network trained during fine-tuning. Approaches that avoid full-model fine-tuning have recently received more attention, as they offer reduced computational costs and address privacy concerns \citep{xiao2023offsite}, while also ensuring better performance in the presence of out-of-distribution data with large distribution shifts \citep{kumar2022fine}.

Fig.~\ref{fig:fc_PTM} shows the presence of IGB on pre-trained models at the beginning of the fine-tuning phase. The experiments are conducted on different architectures; in each of them the last fully connected layer (classifier) is replaced to match the new number of classes and reinitialized. 
\begin{figure}
    \includegraphics[width=.32\textwidth]{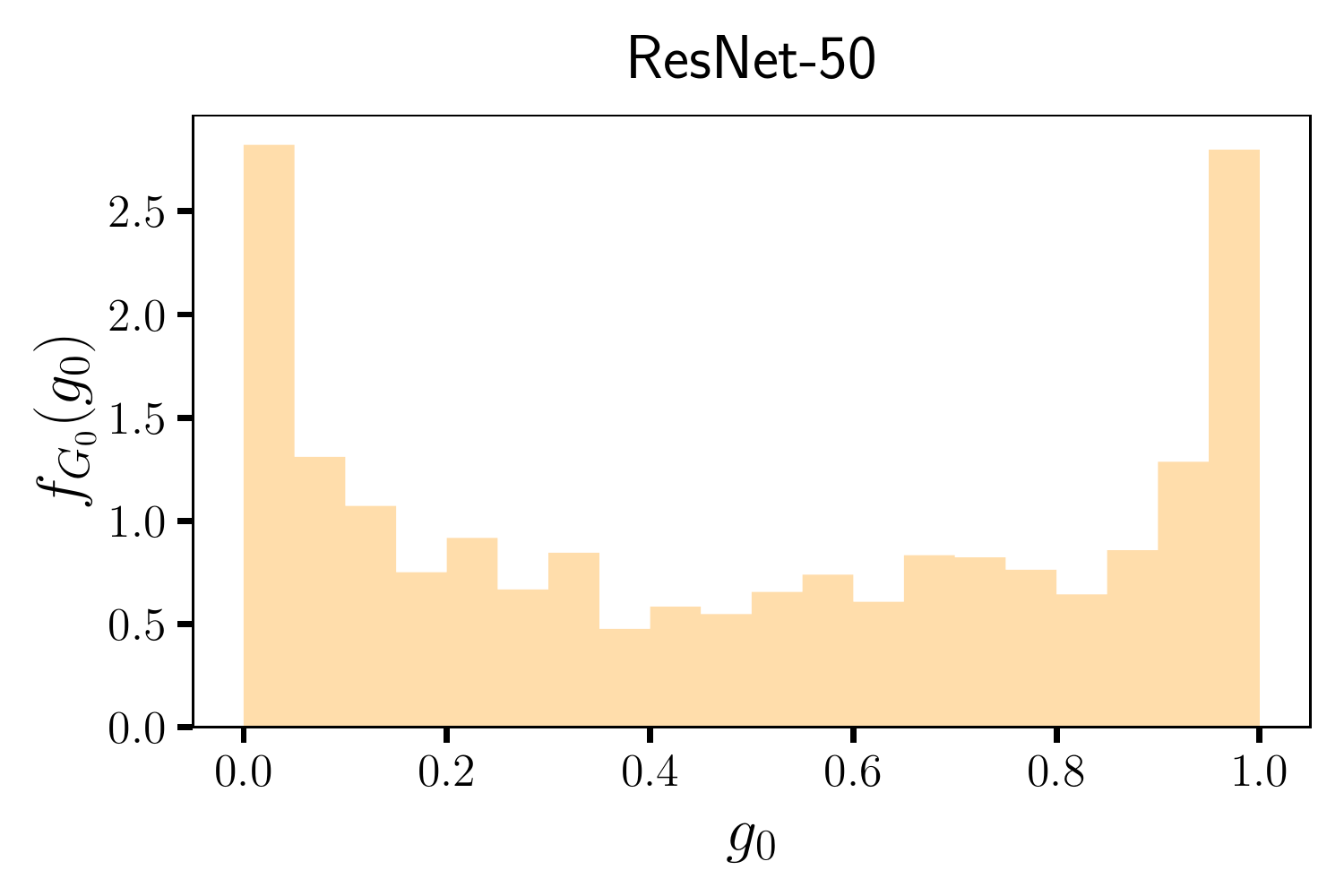}
    \includegraphics[width=.32\textwidth]{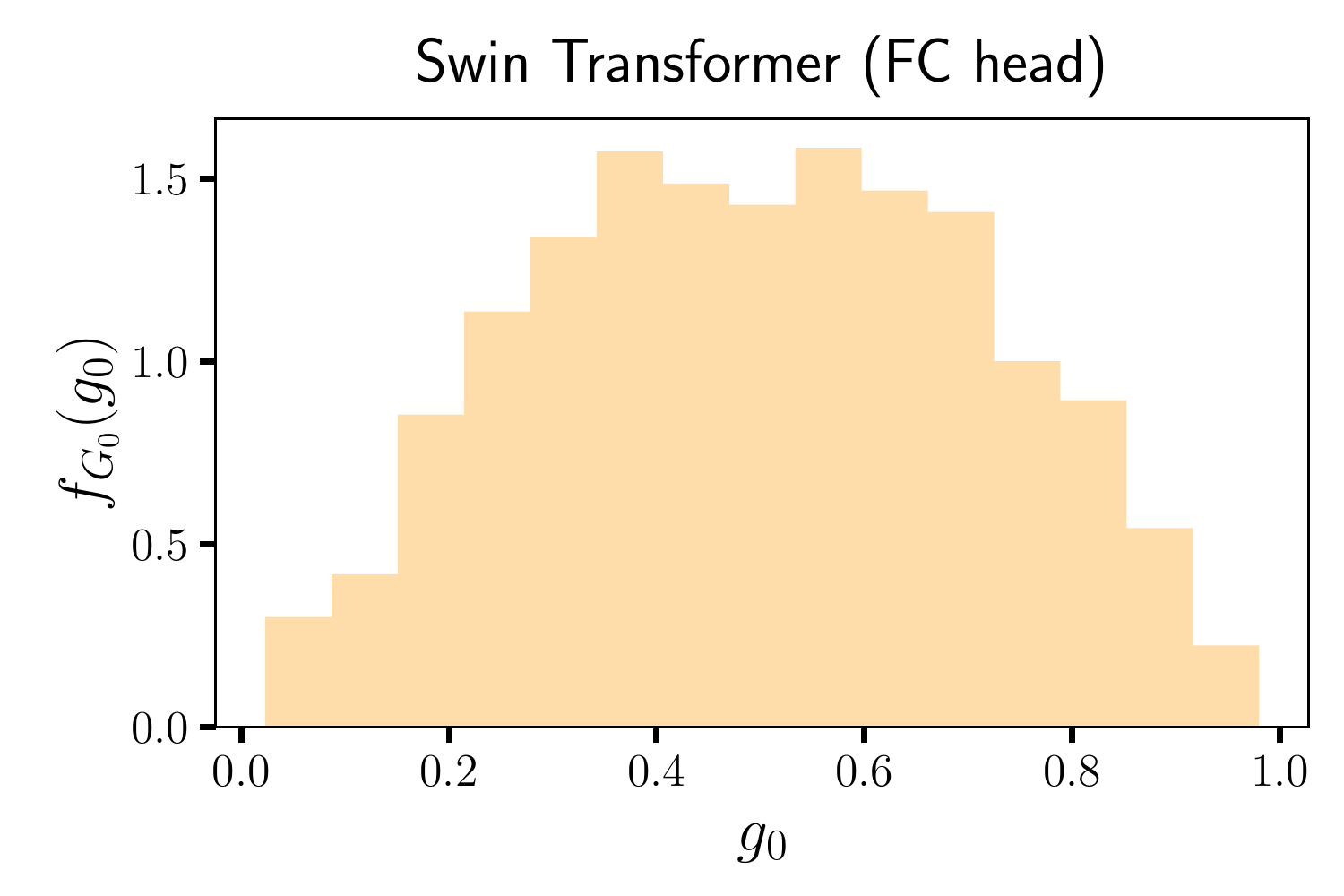}
    \includegraphics[width=.32\textwidth]{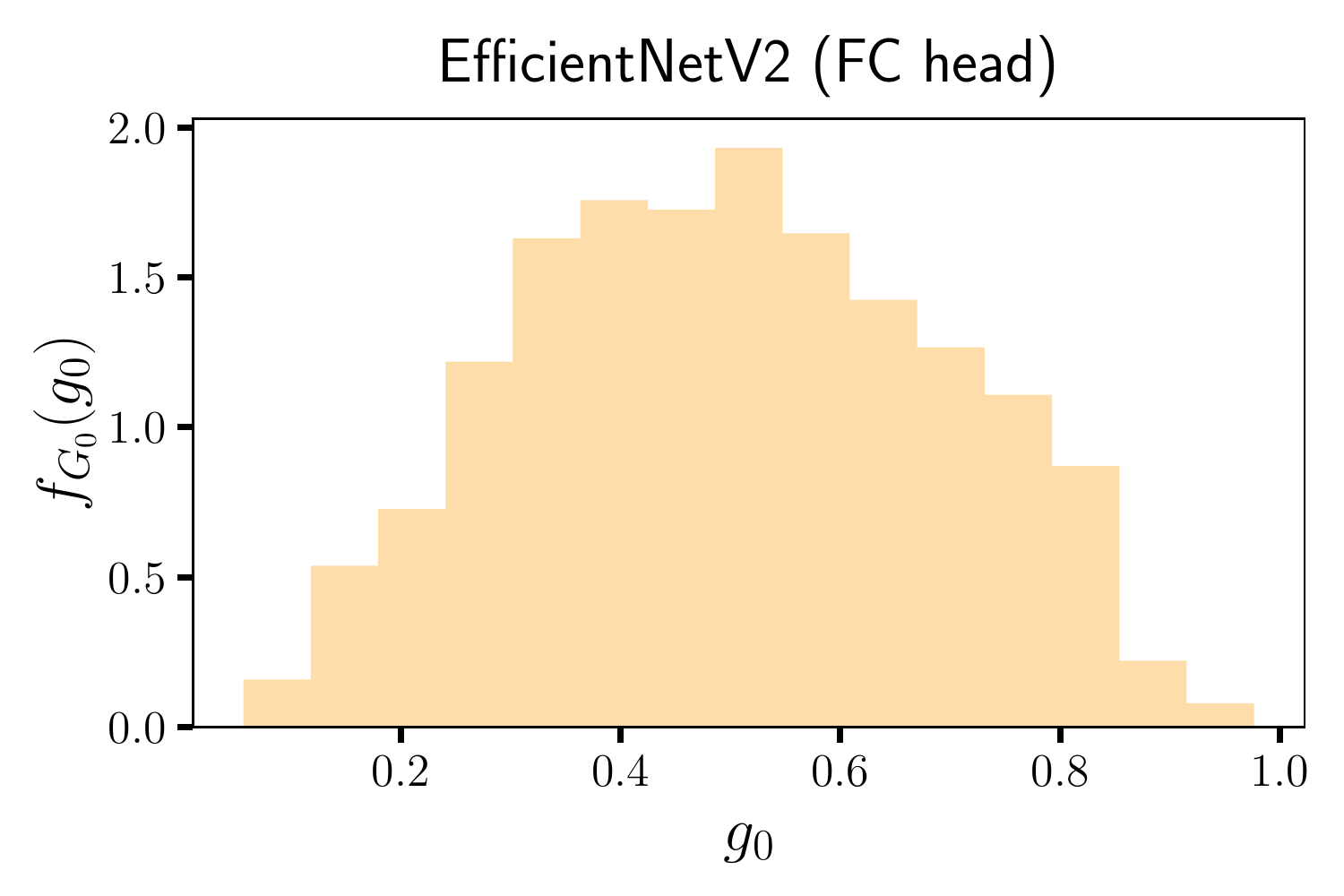}
    \caption{The distribution $\pdf{\RClassFraction{0}}{}{\ClassFraction{0}}$ on pre-trained models. From left to right are shown \PTResNet, ~ \PTSwinTFC, ~ \PTENFC~ on a binary dataset (Cats and Dogs from \CIFAR).} 
    \label{fig:fc_PTM}
\end{figure}
Interestingly, using a deeper untrained structure as head leads to an amplification of IGB as shown in Fig.\ref{fig:mlp_PTM}; the same architectures from experiments in Fig.~\ref{fig:fc_PTM} are considered with the only difference that the fully connected head is now replaced by an untrained MLP.

\begin{figure}
    \includegraphics[width=.48\textwidth]{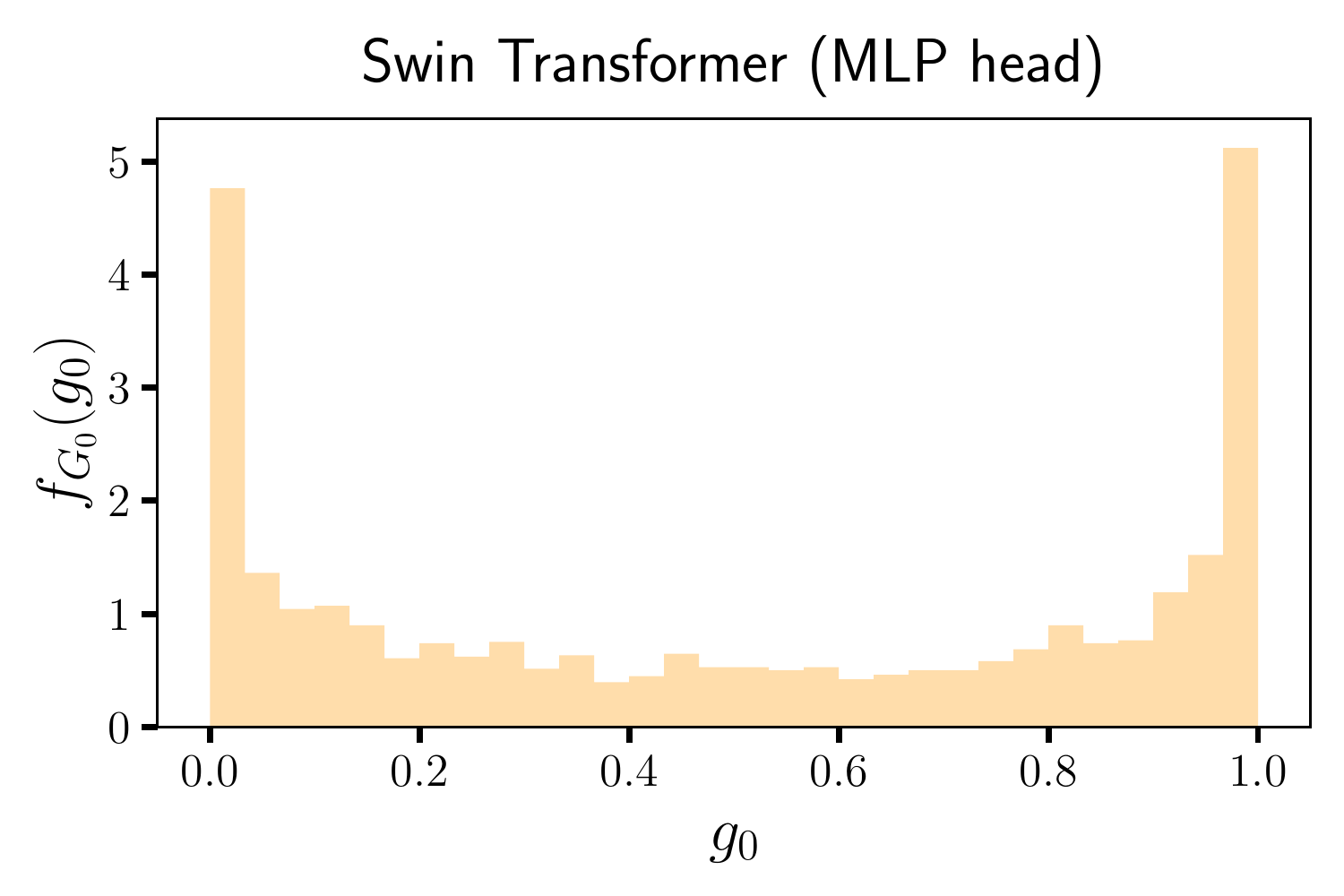}
    \includegraphics[width=.48\textwidth]{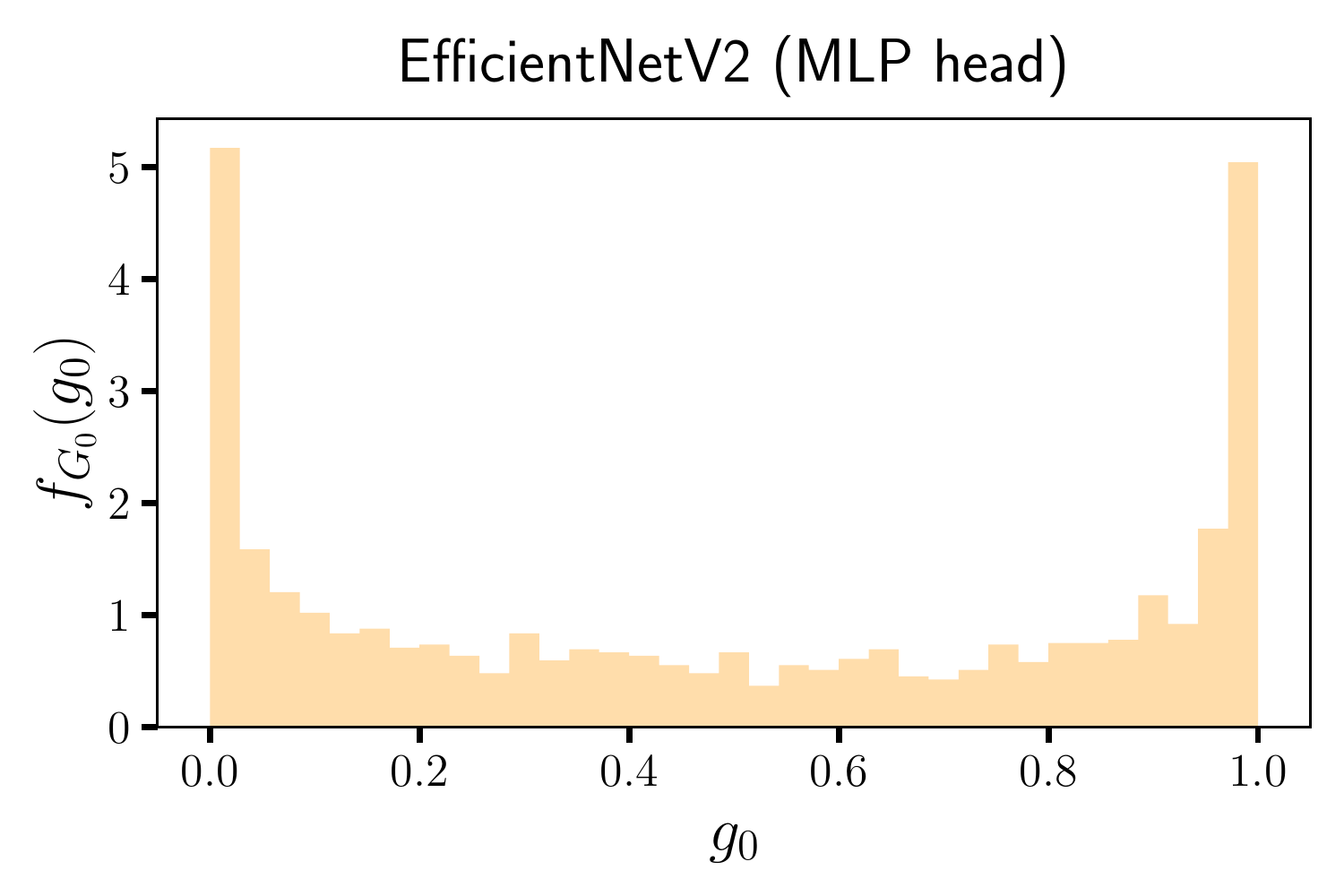}
    \caption{The distribution $\pdf{\RClassFraction{0}}{}{\ClassFraction{0}}$ on pre-trained models. From left to right are shown \PTSwinTMLP, ~ \PTENMLP~ on a binary dataset (Cats and Dogs from \CIFAR)} 
    \label{fig:mlp_PTM}
\end{figure}

\subsection{Reproducibility}\label{sec:reprod}
Here we provide technical details about the experiments, to allow for reproducibility. The code used for the experiments presented in this work are available at \url{https://github.com/EmanueleFrancazi/IGB-Algorithms}.

\paragraph{Datasets}
\begin{itemize}
    \item \textbf{Gaussian Blob} (\GB): Consistent with the analysis presented, for most of the experiments shown we used a Gaussian blob as input. Specifically, all elements of the dataset are \textit{i.i.d.} and each individual element consists of a random vector of $d = 3072$ \textit{i.i.d.} normally distributed components, \textit{i.e.}:
    \begin{equation}\notag
            \InputValue^{(a)}_{b} \sim \NormalDistr{0}{1}\,.
    \end{equation}
    Note that the value of $d$ is chosen so that the random vectors we generate have the same dimension of \texttt{CIFAR10}.

    \item \textbf{CIFAR10} (\CIFAR): We use \texttt{CIFAR10} (\url{https://www.cs.toronto.edu/~kriz/cifar.html})~\citep{cifar} as an example of a real multi-class dataset. Before the start of the simulation we perform the standardization of the dataset: the pixel values are rescaled in the interval $[0, 1]$ and then shifted by the mean value and rescaled by the standard deviations (calculated on each channel).
    \item \textbf{CIFAR100} (\CIFARHC): We use \texttt{CIFAR100} (\url{https://www.cs.toronto.edu/~kriz/cifar.html})~\citep{cifar} as an example of high cardinality dataset, \textit{i.e.} a dataset with a big number of classes.
    \item \textbf{MNIST} (\MNIST): We use \texttt{MNIST} (\url{http://yann.lecun.com/exdb/mnist/})~\citep{deng2012mnist} to reproduce binary experiments on real data. The binary dataset is defined by merging the starting classes into two macro groups according to the parity of digits; thus, we will have the \texttt{even} number class $\{ 0,2,4,6,8 \}$ and the \texttt{odd} number class $\{ 1,3,5,7,9 \}$.
\end{itemize}

\paragraph{Models}
We here provide details on the architectures we used for our experiments. Our description of the models does not include the loss functions because the loss function is irrelevant in untrained networks. 
\begin{itemize}
    \item \textbf{MLP}:  Our analysis provides theoretical predictions for MLP. In order to support the results of the study, we considered two different MLP networks in the proposed experiments:
    \begin{itemize}
        \item \textbf{MLP with a single hidden layer} (\MLPA):  The number of nodes nodes in the hidden layer varies between $\LayerNumNodes{1}=100$ for networks without Pooling and $\LayerNumNodes{1}=500$ for networks equipped with max pooling layer. Different activation functions have been coupled to the networks to show the differences; details regarding the choice of these elements are given case by case.
        \item \textbf{MLP with multiple hidden layers} (\MLPB): In this case we have $L$ hidden layers, each one composed by $N=100$ nodes. As in the previous case, the activation function is an element we varied to set a comparison and underline the differences coming from this choice.
    \end{itemize}
    \item \textbf{CNN}: To show how IGB manifests outside the setting employed in the quantitative treatment presented we propose, some experiments on convolutional neural networks (CNNs) as an alternative to the MLPs discussed above. In particular, we used:
    \begin{itemize}
        \item \CNNA: This architecture was used for simulations related to the histogram in Fig.~\ref{IGB_diagram}; two classes from CIFAR10 (three-channel images) were selected for these experiments. Starting from the input layer we have:
        \begin{itemize}
            \item  a first convolutional layer with: out channels=16 (Number of channels produced by the convolution), $k=5$ ( Size of the convolving kernel), stride=1, padding=2. The output of the layer is then passed through an activation function and a pooling layer. The choice of these elements varies to compare two different scenarios (as explained in Fig.~\ref{IGB_diagram}). In both cases, however,  we set common parameters for the pooling layer, \textit{i.e.} $k=2$ (kernel size), stride=2.
            \item Next comes a second convolutional layer with the same parameters as the previous one, except for the number of output chanels; in this case we have out channels=64. Again the convolutional layer is followed by an activation function and a pooling layer (same as the first layer). The parameters of the pooling layer in this case are \textit{i.e.} $k=4$ (kernel size), stride=4. The processed signal is then connected with a weights layer to the output layer.
        \end{itemize}
        \item \CNNB: We also consider a second CNN architecture deeper than \CNNA. Specifically starting from the input layer:
        \begin{itemize}
            \item we start with a sequence of five convolutional layers (each followed by an activation function and pooling layer). Except for the number of output channels the rest of the parameters are fixed the same for each of these layers, in particular we have for the convolutional layer $=5$, stride=1, padding=4. For the pooling layer, however, $k=5$, stride=1. Finally, the number of output channels for the various layers is $[16, 32, 32, 64, 32]$.
            \item This is followed by an additional convolutional layer defined by the following parameters: out channels=16, $k=5$, stride=1, padding=2. This is accompanied by the activation function and a Pooling layer whose parameters are: $k=2$, stride=2.
            \item Finally, a last sequence of convolutional layers, activation function and pooling layer precedes the output layer. The only parameter that differs from the sequence that precedes it is the kernel size of the pooling layer; specifically in this case $k=4$. The processed signal is then connected with a weights layer to the output layer.         
        \end{itemize}
    \end{itemize}
    \item \textbf{MLP-mixer}: We propose the MLP-mixer (\MLPmix) introduced in \citep{tolstikhin2021mlp} as an example of a more advanced MLP architecture. We use the architectural design documented in \url{https://github.com/omihub777/MLP-Mixer-CIFAR/blob/main/README.md}.

            \item \textbf{ResNet}: We propose ResNet34 (\ResNet), introduced in \citet{he2016deep} as an example of an architecture equipped with skip connections. We use the architectural design documented in \url{https://www.kaggle.com/code/kmldas/cifar10-resnet-90-accuracy-less-than-5-min}.
        \item \textbf{Vision Transformer}: We propose ViT (\ViT), introduced in \citet{dosovitskiy2020image}  as an example of an architecture equipped with multi-head attention mechanism. We use the architectural design documented in \url{https://github.com/tintn/vision-transformer-from-scratch/blob/main/vit.py}.
        \item \textbf{Pre-Trained Resnet}: ResNet50 (\PTResNet) pretrained on Imagenet \citep{deng2009imagenet} (for more details see \url{https://pytorch.org/vision/stable/models/generated/torchvision.models.resnet50.html#torchvision.models.resnet50}). The head is substituted by an untrained classifier (fully connected layer) randomly initialized with Kaiming normal weights and null bias.
        \item \textbf{Pre-Trained Swin Transformer}: swin tiny architecture  pretrained on Imagenet \citep{deng2009imagenet} (for more details see \url{https://pytorch.org/vision/main/models/generated/torchvision.models.swin_t.html#torchvision.models.swin_t}\begin{itemize}
            \item \PTSwinTFC: The classifier (last fully connected layer) is randomly re-initialized with Kaiming normal weights and null bias.
            \item \PTSwinTMLP: The classifier is substituted by a multi-layer perceptron (5 layers) with ReLU activation function randomly initialized with Kaiming normal weights and null bias.
        \end{itemize}
        \item \textbf{Pre-Trained EfficientNet}: EfficientNetV2-S architecture pretrained on Imagenet \citep{deng2009imagenet} (for more details see
        \url{https://pytorch.org/vision/main/models/generated/torchvision.models.efficientnet_v2_s.html#torchvision.models.efficientnet_v2_s}
        )
        \begin{itemize}
            \item \PTENFC: The classifier (last fully connected layer) is randomly re-initialized with Kaiming normal weights and null bias.
            \item \PTENMLP: The classifier is substituted by a multi-layer perceptron (5 layers) with ReLU activation function randomly initialized with Kaiming normal weights and null bias.
        \end{itemize}
\end{itemize}
Note that in both the CNN architectures we use, the final convolutional layer is directly connected to the output, without any additional fully-connected hidden layer (as the ones described by our theory). This indicates that the observed IGB is also also a feature of CNNs independently of the presence of fully-connected hidden layers at the end of the network (which as we proved would also cause IGB).

\paragraph{Dynamics}
The simulations concerning the dynamics (Sec.~\ref{sec:Imp_IGB} App.~\ref{app:dyn}) use three different architectures (\ResNet,~\MLPmix,~\ViT) following the settings proposed in their respective repositories. For the \ViT simulations, the dataset was augmented following \url{https://github.com/omihub777/ViT-CIFAR/blob/main/autoaugment.py} in order to slightly improve performance compared to the baseline documented in the repo.
We highlight that the proposed architectures were not selected as representations of the state of the art, but rather as examples of well-documented architectures capable of achieving good performance quickly (\textit{e.g.} \url{https://www.kaggle.com/code/kmldas/cifar10-resnet-90-accuracy-less-than-5-min}). In this way, we not only show the prevalence of IGB, observed in each of these architectures introduced before our work. But, at the same time, the proposed examples do not require excessive resources, facilitating reproducibility and making the results of the work more accessible.

\section{Limitations and ethics}\label{sec:limitations}
\paragraph{Limitations}
Our work focuses on systems simple enough to clearly show the main aspects of the phenomenon and at the same time complex enough to investigate non-trivial effects induced, for example, by network depth. A comprehensive picture emerges that clarifies the effect of some particular elements of the architecture and their connection with IGB. On the other hand, the observation of IGB is not restricted to the subset of networks/datasets considered in the study. Although the treatment of more articulated setups is outside the scope of the study, whose main goal is to present the phenomenon (which to the best of our knowledge has never been reported in the literature) in a clear manner, the characterization of IGB for more realistic systems remains an interesting question.

\paragraph{Ethics}
By informing model selection, data preparation and initial conditions, our results can improve the training of machine learning models. Better-performing machine learning models allow to better address wide ranges of problems, but can also be adapted for potentially harmful applications~\citep{hutson2021should,qadeer21}.\\

The \href{https://www.cs.toronto.edu/~kriz/cifar.html}{CIFAR} datasets, are subsets of the 80 million tiny
images, which are formally withdrawn since it contains some derogatory terms as
categories and offensive images (\url{ http://groups.csail.mit.edu/vision/TinyImages/}). However, note that none of the experiments described in the paper was performed on the overall tiny images dataset: the said derogatory images are not present in CIFAR10 nor CIFAR100.

}

\end{document}